\documentclass{article}

\PassOptionsToPackage{numbers, sort&compress}{natbib}

\usepackage[main, final]{neurips_2026}

\usepackage[utf8]{inputenc} 
\usepackage[T1]{fontenc}    
\usepackage{hyperref}       
\usepackage{url}            
\usepackage{booktabs}       
\usepackage{amssymb}       
\usepackage{nicefrac}       
\usepackage{microtype}      
\usepackage{xcolor}         
\usepackage[pdftex]{graphicx}
\usepackage{wrapfig}
\usepackage{amsmath}
\usepackage{amsthm}
\usepackage{multirow}
\usepackage[para]{threeparttable}
\usepackage{upgreek}
\usepackage{comment}

\newtheorem{theorem}{Theorem}[section]
\newtheorem{corollary}[theorem]{Corollary}
\newtheorem{lemma}{Lemma}[section]
\newtheorem{example}{Example}[section]
\theoremstyle{remark}
\newtheorem{remark}{Remark}

\DeclareMathOperator*{\argmin}{arg\,min}

\title{Projected Neural Additive Models as Universal Approximators}

\author{Nhon N.~Phan\thanks{Work partially done at Columbia University.} \\
Terminal Effects Division \\
DEVCOM Army Research Laboratory \\
Aberdeen Proving Ground, MD 21005, USA \\
\texttt{nhon.n.phan.civ@army.mil} \\
\And
Qiang Du \\
Department of Applied Physics \\
\& Applied Mathematics, \\
and Data Science Institute \\
Columbia University \\
New York, NY 10027, USA \\
\texttt{qd2125@columbia.edu} \\
\And
John D.~Clayton \\
Terminal Effects Division \\
DEVCOM Army Research Laboratory \\
Aberdeen Proving Ground, MD 21005, USA \\
\texttt{john.d.clayton1.civ@army.mil} \\
\And
WaiChing Sun \\
Department of Civil Engineering \\
\& Engineering Mechanics, \\
and Data Science Institute \\
Columbia University \\
New York, NY 10027, USA \\
\texttt{wsun@columbia.edu} \\
}

\begin{document}

\maketitle

\begin{abstract}
Neural additive models (NAMs) offer interpretability but lack expressivity due to their rigid additive structure, whereas multi-layer perceptrons (MLPs) achieve universal approximation at the cost of transparency.
We establish the first theoretical foundation for projected neural additive models (PNAMs), which augment NAMs with a learnable linear transformation $\boldsymbol{T}$ to capture cross-variable interactions.
While standard NAMs cannot approximate even the simplest interaction term $\chi_1 \chi_2$ because their cross-derivatives are identically zero, we prove that PNAMs are universal approximators and characterize the minimum number of projections required for exact polynomial representation.
Furthermore, we derive an approximation rate of $\mathcal{O}(M^{-s / (N - 1)})$ for $\mathcal{C}^s$ functions (i.e., functions with $s$ continuous derivatives), where $M$ denotes the number of projections and $N$ the input dimension; this rate is known to be optimal for ridge function approximation.
To recover interpretability from the projected inputs, we introduce structured sparsity regularizers and post-hoc symbolic regression techniques that enable global input ranking, parameter pruning, and the conversion of learned bases into compact analytical expressions.
Experiments on knot theory invariants, MNIST, and phase field fracture mechanics demonstrate that PNAMs match or surpass similar architectures in accuracy while discovering salient features and yielding symbolic models with orders of magnitude fewer parameters.\footnote{\label{fn:code}Code is available at \url{https://github.com/nhonphan7/pnam}.}
\end{abstract}

\section{Introduction}
While deep neural networks have been widely adopted across diverse tasks due to their expressivity, they inherently lack interpretability \citep{murdoch2019definitions}.
Among the myriad methods introduced to address this issue, \citet{agarwal2021neural} propose the neural additive model (NAM), an architecture constrained to a linear combination of single-variable functions parameterized by independent multi-layer perceptrons (MLPs).
Although this decoupled structure provides feature-wise interpretability, the strictly additive nature of NAMs limits their expressivity; in particular, their cross-derivatives are identically zero, rendering them incapable of representing functions with nontrivial input interactions.

\begin{wrapfigure}[35]{r}{0.4\textwidth}
\centering
\includegraphics[width=0.4\textwidth,trim=2cm 1cm 4cm 1cm,clip]{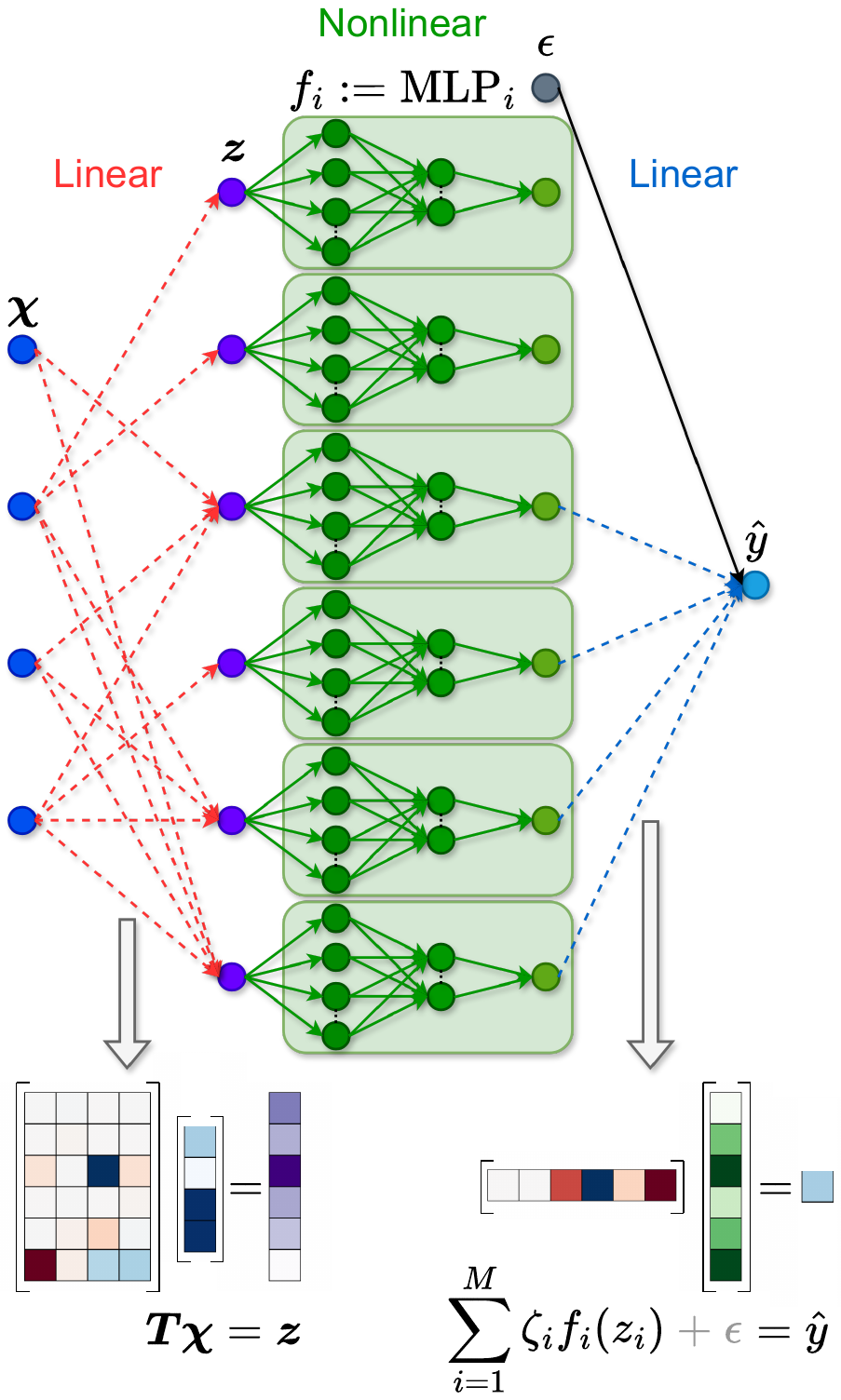}
\caption{Architecture of the PNAM. The output $\widehat{y}$ is predicted via a linear combination of single-variable functions, parameterized by independent MLPs, which are evaluated on linear combinations of the inputs $\boldsymbol{\chi}$. The transformation $\boldsymbol{T}$ and weighting coefficients $\boldsymbol{\zeta}$ (both denoted with dashed lines) can be optimized to yield sparser (and more interpretable) solutions.}
\label{fig:pnam}
\end{wrapfigure}

To overcome this limitation, \citet{phan2025hydra} introduce a learnable linear transformation $\boldsymbol{T}$ preceding the NAM (see Fig.~\ref{fig:pnam})---an architecture we refer to as the projected neural additive model (PNAM).
A complementary line of work---GA$^2$M \citep{lou2013accurate}, NODE-GA$^2$M \citep{chang2021node}, GAMI-Net \citep{yang2021gami}, SPAM \citep{dubey2022scalable}, and SIAN \citep{enouen2022sparse}---augments generalized additive models (GAMs) \citep{hastie1986generalized} and NAMs with explicit interaction terms such as $\mathcal{F}(\chi_i, \chi_j)$ acting on pairs of inputs.
Interpretability is then pairwise, but the number of interaction terms grows combinatorially with the number of inputs $N$, and the interaction order is capped by construction: a model limited to pairwise terms cannot represent, e.g., $\chi_1 \chi_2 \chi_3$ and is therefore not a universal approximator.
If the ground truth is dominated by a few low-order interactions of raw features, such explicit models are more directly interpretable.
The PNAM is instead preferable when interactions are higher-order, dense in a few directions, or unknown a priori, as it captures them along $M$ learned directions, with $M$ far below the $N (N - 1) / 2$ pairs that a pairwise model enumerates.
While the transformation $\boldsymbol{T}$ empirically enhances expressivity by capturing cross-variable couplings, the PNAM architecture was introduced without a rigorous theoretical foundation or a formal proof of universal approximation.

\paragraph{Contributions.}
Our core contributions establish this missing foundation alongside novel mechanisms for interpretability and sparsity.
First, we prove that PNAMs are universal approximators (Theorem \ref{thm:approx})---a fundamental guarantee previously missing for this architecture class.
Second, we characterize the projection complexity of $\boldsymbol{T}$ (Appendix \ref{subsec:reprod}) by establishing constructive upper bounds and a parameter-counting lower bound on the minimum number of projections $M$ required to represent any degree-$P$ polynomial in $N$ variables exactly.
Third, building on these results, we derive an approximation rate of $\mathcal{O}(M^{-s / (N - 1)})$ for $\mathcal{C}^s$ functions (Theorem \ref{thm:rate}), a rate known to be optimal for ridge function approximation \citep{maiorov1999best, pinkus1999approximation}.

To mitigate the reduction in interpretability induced by $\boldsymbol{T}$, we introduce structured regularization techniques that enable global feature ranking, penalize unnecessary input interactions, and promote sparse solutions.
While standard NAMs provide feature-level interpretability via their shape functions, our regularized PNAMs offer global input ranking across the full dataset \citep{molnar2019interpretable, rudin2022interpretable}.
Moreover, we leverage the modularity of PNAMs to prune nonessential parameters and extract compact symbolic expressions from the learned single-variable bases.
Empirically, experiments on knot theory invariants confirm that PNAMs recover accurate analytical expressions comparable to Kolmogorov--Arnold networks (KANs) \citep{liu2025kan}, entirely without prior knowledge of the relevant invariants.

\paragraph{Related work.}
The PNAM belongs to a growing class of architecture-based models designed to enhance the interpretability of deep neural networks without compromising expressivity.
Prominent alternatives include KANs \citep{liu2025kan}, deep polynomial neural networks \citep{chrysos2021deep}, and graph neural networks with inductive biases \citep{cranmer2020discovering}, alongside related interpretability methods such as SINDy \citep{brunton2016discovering}, structural pruning \citep{fang2023depgraph}, and gradient-based attribution \citep{sundararajan2017axiomatic}.
Compared to KANs---perhaps their closest architectural relative\footnote{In fact, PNAMs can be viewed as a special case of KANs in which the inner functions are constrained to be linear.}---PNAMs offer a distinct and complementary theoretical profile.
While KANs leverage the Kolmogorov--Arnold representation theorem (which necessitates two layers of nonlinearities), PNAMs utilize a single linear transformation followed by 1D nonlinearities.
This simpler structure enables the approximation rate of Theorem \ref{thm:rate}.

In addition, we emphasize that although we have equipped PNAMs with post-hoc symbolic regression (SR) capabilities, this technique is primarily employed for pruning.
Like most additive architectures, PNAMs are not designed to recover exact physical laws with precise functional forms; their additive nature prevents them from compactly approximating operations such as division,\footnote{This limitation could potentially be addressed using the EML operator recently introduced in \citep{odrzywolek2026all}.} and the number of projections required to represent complex interactions can be prohibitively large.
Consequently, we reserve the recovery of exact physical laws for dedicated SR algorithms leveraging reinforcement learning \citep{petersen2019deep}, physics-inspired strategies \citep{udrescu2020ai}, and genetic programming \citep{cranmer2023interpretable}.
Further discussion on interpretability and optimization constraints is provided in Section \ref{sec:limit}.

\section{Projected neural additive models} \label{sec:pnams}
Given a training dataset $\{\boldsymbol{\mathcal{X}}, \boldsymbol{y}\} = \{(\boldsymbol{\chi}_\nu, y_\nu)\}_{\nu = 1}^D$ containing $D$ input-output pairs, where $\boldsymbol{\chi} \in \mathbb{R}^N$ is an input vector of $N$ independent variables and $y \in \mathbb{R}$ is the corresponding scalar label, the goal of supervised learning is to construct an approximating function $\mathcal{F}: \mathbb{R}^N \rightarrow \mathbb{R}$.
We hypothesize that this multi-variable function can be approximated by a weighted sum of $M$ single-variable functions $\{f_i\}_{i = 1}^M$ acting on a projected space:
\begin{equation}
\label{eq:output_param}
\widehat{y} = \sum_{i = 1}^M \zeta_i f_i\left(\sum_{j = 1}^N T_{ij} \chi_j\right) + \epsilon = \sum_{i = 1}^M g_i(z_i) + \epsilon,
\end{equation}
where $\widehat{y}$ is the predicted value and $\epsilon$ represents an additive noise term.
The projected variables $\boldsymbol{z} \in \mathbb{R}^M$ in Eq.~\eqref{eq:output_param} result from a linear transformation $\boldsymbol{T}: \mathbb{R}^N \rightarrow \mathbb{R}^M$ of the inputs, such that $\boldsymbol{z} = \boldsymbol{T} \boldsymbol{\chi}$.
For compactness, each feature function $f_i: \mathbb{R} \rightarrow \mathbb{R}$ and its corresponding weighting coefficient $\zeta_i$ are combined into a composite function $g_i: \mathbb{R} \rightarrow \mathbb{R}$.

The single-variable functions $\{f_i\}$ can be implemented using various bases (e.g., polynomials or neural networks); here, we parameterize the output as a linear combination of MLPs:
\begin{equation}
\label{eq:output_mlp}
\widehat{y} = \sum_{i = 1}^M \zeta_i \mathrm{MLP}_i\left(\sum_{j = 1}^N T_{ij} \chi_j; \boldsymbol{\theta}_i\right) + \epsilon,
\end{equation}
where each $f_i$ is an $L$-layer MLP with a set of learnable weights and biases $\boldsymbol{\theta}_i = \{\boldsymbol{W}_i^{(1)}, \dots, \boldsymbol{W}_i^{(L)}, \boldsymbol{s}_i^{(1)}, \dots, \boldsymbol{s}_i^{(L - 1)}\}$ and an element-wise activation function $A$:
\begin{equation}
\label{eq:mlp}
\mathrm{MLP}_i(z_i; \boldsymbol{\theta}_i) = \boldsymbol{W}_i^{(L)} A\left(\boldsymbol{W}_i^{(L - 1)} A\left(\dots A\left(\boldsymbol{W}_i^{(1)} z_i + \boldsymbol{s}_i^{(1)}\right) \dots\right) + \boldsymbol{s}_i^{(L - 1)}\right).
\end{equation}
To effectively capture high-frequency components, we leverage a Fourier feature mapping \citep{tancik2020fourier, bahmani2024discovering} that transforms each projected variable $z_i$ into
\begin{equation}
\label{eq:fourier}
\gamma_i(z_i) = [\cos(2 \pi \boldsymbol{B}_i z_i)^\mathrm{T}, \sin(2 \pi \boldsymbol{B}_i z_i)^\mathrm{T}]^\mathrm{T},
\end{equation}
where entries in $\boldsymbol{B}_i$ are sampled from a Gaussian distribution $\mathcal{N}(0, \sigma^2)$, with standard deviation $\sigma$, and fixed after initialization.
When this mapping is used, the MLP in Eq.~\eqref{eq:mlp} is evaluated on $\gamma_i(z_i)$ rather than the scalar $z_i$.
As illustrated in Fig.~\ref{fig:pnam}, the transformation $\boldsymbol{T}$, weighting coefficients $\boldsymbol{\zeta} = \{\zeta_i\}_{i = 1}^M$, and error term $\epsilon$ in Eq.~\eqref{eq:output_mlp} are encoded as two weight matrices and a bias term, respectively.
This construction formalizes the PNAM architecture, extending the foundational and empirical frameworks established by \citet{agarwal2021neural} and \citet{phan2025hydra}.

\begin{remark}[Comparison with NAMs and MLPs]
In a PNAM, each basis $g_i$ in Eq.~\eqref{eq:output_param} is a function of a transformed variable $z_i$, rather than a raw input $\chi_j$ as in NAMs or other GAMs \citep{hastie1986generalized, radenovic2022neural}.
The linear transformation $\boldsymbol{T}$ enables the model to capture cross-variable couplings (e.g., $\chi_1 \chi_2$ in Example \ref{ex:two_dim}) that standard NAMs cannot.
Although this projection precludes the direct visualization of input contributions typical of NAMs, we demonstrate in Section \ref{subsec:sr} how the static matrix $\boldsymbol{T}$ can be examined to perform global input ranking and identify critical interactions.
This structured interpretability---encompassing feature pruning and dimensionality reduction---is generally absent in black-box, fully connected MLPs.
\end{remark}

\subsection{Universal approximation and approximation rate} \label{subsec:approx}
By the Stone--Weierstrass theorem \citep{stone1937applications, stone1948generalized, cotter1990stone}, polynomials are dense in the space of continuous functions $\mathcal{C}(\mathfrak{D})$ for a compact domain $\mathfrak{D} \subset \mathbb{R}^N$, meaning they can approximate any continuous function to arbitrary precision.
Consequently, we can establish universal approximation for PNAMs by demonstrating that they can reproduce arbitrary multi-variable polynomials.
The key insight is that the learnable linear transformation $\boldsymbol{T}$ enables PNAMs to express these polynomials via a superposition of ridge functions---functions of the form $g_i(\sum_j T_{ij} \chi_j)$---whereas NAMs, restricted to $z_i = \chi_i$ (the identity projection), cannot represent even the degree-2 cross-term $\chi_1 \chi_2$ and therefore fail to reproduce arbitrary polynomials.
We show in Theorem \ref{thm:induction} that single-variable polynomials of projected inputs can serve as the bases to reproduce any multi-variable polynomial, providing the constructive foundation for the general approximation results that follow.

Alternatively, instead of single-variable polynomials, one can employ a sum of 1D neural networks acting on the projected inputs to achieve universal approximation.
Since these networks satisfy the 1D universal approximation property \citep{hornik1989multilayer, leshno1993multilayer, lu2017expressive}, they can readily approximate the required single-variable polynomials used to construct the multi-variable approximation.
This, by extension, guarantees the universal approximation property of the resulting multi-dimensional PNAM architecture.
This capability is formally stated in the following theorem.
\begin{theorem}[Universal approximation]
\label{thm:approx}
Let $\mathfrak{G}$ be a set of single-variable polynomials or 1D neural networks.
Given any function $\mathcal{F}_1$ in the set of continuous real-valued functions $\mathcal{C}(\mathfrak{D})$ for a compact domain $\mathfrak{D} \subset \mathbb{R}^N$ and any $\varepsilon > 0$, there exist a linear transformation $\boldsymbol{T}: \mathbb{R}^N \rightarrow \mathbb{R}^M$, where $M \geq 1$, and a set of functions $\{g_i: \mathbb{R} \rightarrow \mathbb{R}\}_{i = 1}^M$ in $\mathfrak{G}$ such that
\begin{equation}
\label{eq:approx1}
\left|\mathcal{F}_1(\boldsymbol{\chi}) - \sum_{i = 1}^M g_i\left(\sum_{j = 1}^N T_{ij} \chi_j\right)\right| < \varepsilon,
\end{equation}
for all $\boldsymbol{\chi} \in \mathfrak{D}$.
\end{theorem}

\begin{proof}
Since $\mathfrak{D}$ is compact, the Stone--Weierstrass theorem applies \citep{stone1937applications, stone1948generalized, cotter1990stone}, that is, polynomials are dense in $\mathcal{C}(\mathfrak{D})$.
As a result, there exists a (multi-variable) polynomial $\mathcal{F}_2$ such that
\begin{equation}
\label{eq:approx2}
|\mathcal{F}_1(\boldsymbol{\chi}) - \mathcal{F}_2(\boldsymbol{\chi})| < \varepsilon / 2.
\end{equation}
Then for $\mathcal{F}_2$, by Theorem \ref{thm:induction}, we have
\begin{equation}
\label{eq:approx3}
\mathcal{F}_2(\boldsymbol{\chi}) = \sum_{i = 1}^M \widetilde{g}_i\left(\sum_{j = 1}^N T_{ij} \chi_j\right),
\end{equation}
with single-variable polynomials $\{\widetilde{g}_i\}$.
Note that Theorem \ref{thm:induction} covers any polynomial $\mathcal{F}_2$, establishing the existence of a transformation $\boldsymbol{T}$ and functions $\{\widetilde{g}_i\}$ that reproduce it exactly.
If $\mathfrak{G}$ is a set of polynomials, we are done; if $\mathfrak{G}$ represents neural networks, we treat the polynomials $\{\widetilde{g}_i\}$ as intermediate target functions.
By the universal approximation theorem for neural networks \citep{hornik1989multilayer, leshno1993multilayer, lu2017expressive}, for each $i$, there exists a network $g_i$ with
\begin{equation}
\label{eq:approx4}
\|\widetilde{g}_i - g_i\|_{\mathcal{C}(\mathfrak{D}_i)} < \frac{\varepsilon}{2 M},
\end{equation}
where $\|\cdot\|_{\mathcal{C}(\mathfrak{D}_i)}$ denotes the supremum norm over $\mathfrak{D}_i$, the image of $\mathfrak{D}$ under the $i^\mathrm{th}$ projection.
Summing over all $i = 1, \dots, M$ and applying the triangle inequality yield
\[
\left|\mathcal{F}_2(\boldsymbol{\chi}) - \sum_{i = 1}^M g_i\left(\sum_{j = 1}^N T_{ij} \chi_j\right)\right| \leq \sum_{i = 1}^M \|\widetilde{g}_i - g_i\|_{\mathcal{C}(\mathfrak{D}_i)} < M \cdot \frac{\varepsilon}{2 M} = \frac{\varepsilon}{2}.
\]
Combining this result with Eq.~\eqref{eq:approx2} via the triangle inequality gives Eq.~\eqref{eq:approx1}, completing the proof.
\end{proof}

Theorem \ref{thm:approx} establishes that PNAMs can approximate any continuous function to arbitrary precision, but does not quantify the rate at which the approximation error decreases with the number of projections $M$.
Leveraging the polynomial reproducing property of Theorem \ref{thm:induction} and the classical Jackson theorem \citep{devore1993constructive}, the following theorem derives an explicit approximation rate.
\begin{theorem}[Approximation rate]
\label{thm:rate}
Let $\mathcal{F}_1 \in \mathcal{C}^s(\mathfrak{D})$ for a compact domain $\mathfrak{D} \subset \mathbb{R}^N$ with $s \geq 1$ and $\|\mathcal{F}_1\|_{\mathcal{C}^s} \leq B$.
For any $M \geq 1$, there exists a PNAM $\widehat{\mathcal{F}}_M$ with $M$ projections satisfying
\begin{equation}
\label{eq:rate1}
\|\mathcal{F}_1 - \widehat{\mathcal{F}}_M\|_\infty \leq C_1 M^{-s / (N - 1)},
\end{equation}
where $C_1$ depends on $\mathfrak{D}$, $s$, $N$, and $B$.
\end{theorem}

\begin{proof}
By Jackson's theorem \citep{devore1993constructive}, there exists a polynomial $\mathcal{F}_2$ of degree $P$ such that
\begin{equation}
\label{eq:rate2}
\|\mathcal{F}_1 - \mathcal{F}_2\|_\infty \leq C_2 P^{-s},
\end{equation}
where $C_2$ depends on $\mathfrak{D}$, $s$, $N$, and $B$.
By Theorem \ref{thm:induction}, $\mathcal{F}_2$ can be represented exactly by a PNAM with $M \leq (P + 1)^{N - 1}$ projections.
Inverting this relation gives $P \geq M^{1 / (N - 1)} - 1$.
Thus, for large $M$, the asymptotic scaling satisfies $P \asymp M^{1 / (N - 1)}$.
Substituting this result into the Jackson bound from Eq.~\eqref{eq:rate2} yields
\[
\|\mathcal{F}_1 - \mathcal{F}_2\|_\infty \leq C_2 P^{-s} \leq C_1 M^{-s / (N - 1)},
\]
where $C_2$ and factors from the inversion are absorbed into $C_1$.
Because $\mathcal{F}_2$ is represented exactly by the PNAM $\widehat{\mathcal{F}}_M$, the approximation error $\|\mathcal{F}_2 - \widehat{\mathcal{F}}_M\|_\infty$ vanishes.
Applying the triangle inequality:
\[
\|\mathcal{F}_1 - \widehat{\mathcal{F}}_M\|_\infty \leq \|\mathcal{F}_1 - \mathcal{F}_2\|_\infty + \|\mathcal{F}_2 - \widehat{\mathcal{F}}_M\|_\infty \leq C_1 M^{-s / (N - 1)},
\]
which completes the proof.\footnote{While this result is established for single-variable polynomials $\{\widetilde{g}_i\}$, using neural network bases $\{g_i\}$ does not affect the approximation rate of $\mathcal{O}(M^{-s / (N - 1)})$. This rate is governed by the polynomial approximation step via Jackson's theorem; the subsequent 1D network approximation error (see Eq.~\eqref{eq:approx4}) can be made arbitrarily small independently, contributing only to the leading constant $C_1$.}
\end{proof}

\begin{remark}[Choice of projection bound]
The approximation rate of $\mathcal{O}(M^{-s / (N - 1)})$ in Theorem \ref{thm:rate} follows directly from the bound on $M$ established in Theorem \ref{thm:induction}.
Using the alternative bounds from Theorems \ref{thm:polar} or \ref{thm:waring} would produce weaker results: Theorem \ref{thm:waring} gives $M \asymp P^N / N!$, which recovers the classical polynomial approximation rate of $\mathcal{O}(M^{-s / N})$ \citep{devore1993constructive, pinkus1999approximation}, while the exponential scaling of Theorem \ref{thm:polar} yields only a logarithmic rate of $\mathcal{O}((\log M)^{-s})$.
Theorem \ref{thm:induction} therefore provides the best approximation rate among the three bounds, since it requires the fewest projections for a given polynomial degree $P$ in the large-$P$ regime.
Moreover, \citet{maiorov1999best} proves that $\mathcal{O}(M^{-s / (N - 1)})$ is the optimal rate for ridge function approximation of Sobolev-class functions, confirming that the bound in Theorem \ref{thm:induction} is tight in this asymptotic sense.
\end{remark}

\begin{remark}[Parameter complexity]
To achieve approximation error $\varepsilon$, Theorem \ref{thm:rate} requires $M \asymp \varepsilon^{-(N - 1) / s}$ projections.
To achieve a matching error, each $g_i$ approximating a single-variable function in $\mathcal{C}^s$ is a network with width $W \asymp \varepsilon^{-1 / (2 s)}$.
Thus, the parameter complexity of PNAMs is
\[
\mathcal{O}(M N + M L W^2) \asymp \mathcal{O}(\varepsilon^{-(N - 1) / s} \cdot \varepsilon^{-1 / s}) = \mathcal{O}(\varepsilon^{-N / s})
\]
for fixed depth $L$, where the $M N$ term is dominated by $M L W^2$ as $\varepsilon \rightarrow 0$.
This result matches the optimal complexity established for standard MLPs \citep{yarotsky2017error}.
\end{remark}

\subsection{Extension to multiple outputs} \label{subsec:outputs}
To accommodate multiple outputs, let $\boldsymbol{y} = \{\{y_{i\nu}\}_{\nu = 1}^D\}_{i = 1}^K$, $\boldsymbol{\zeta} = \{\{\zeta_{ij}\}_{j = 1}^M\}_{i = 1}^K$, and $\boldsymbol{\epsilon} = \{\epsilon_i\}_{i = 1}^K$, with $K$ denoting the number of dependent variables.
We extend Eq.~\eqref{eq:output_param} to represent each output $\widehat{y}_i$ as
\begin{equation}
\label{eq:outputs_param}
\widehat{y}_i = \sum_{j = 1}^M \zeta_{ij} f_{ij}\left(\sum_{k = 1}^N T_{jk} \chi_k\right) + \epsilon_i = \sum_{j = 1}^M g_{ij}(z_j) + \epsilon_i,
\end{equation}
where $f_{ij}: \mathbb{R} \rightarrow \mathbb{R}$ is the single-variable function associated with the $j^\mathrm{th}$ projection for the $i^\mathrm{th}$ output \citep{xu2023sparse}.
To see that Eq.~\eqref{eq:outputs_param} is a natural extension of Eq.~\eqref{eq:output_param}, consider the case where the bases $\{g_{ij}\}$ share no common inputs $\{z_j\}$ across different outputs $\{\widehat{y}_i\}$.
In this scenario, the total number of projections is $M = \sum_{i = 1}^K m_i$, with $m_i$ denoting the number of projections required to approximate each $\widehat{y}_i$ independently.
Equivalently, for each index $i$, both the coefficients $\{\zeta_{ij}\}$ and functions $\{f_{ij}\}$ are nonzero only for a disjoint subset of indices $\{j\}$ of size $m_i$, yielding $K$ independent basis sets.
By extension of Theorem \ref{thm:approx}, $M$ can be chosen sufficiently large to approximate any $K$-tuple of continuous functions on a compact domain.

\subsection{Learning problem and constraints for sparsity} \label{subsec:constraints}
Let $\boldsymbol{\Theta} = \{\boldsymbol{\theta}_j\}_{j = 1}^M$ denote all learnable parameters of the 1D MLPs.
We optimize the parameters of PNAMs by minimizing the following loss function (in parentheses) for $D$ training samples:
\begin{equation}
\label{eq:nn_loss}
\begin{aligned}
\boldsymbol{T}^\ast, \boldsymbol{\Theta}^\ast, \boldsymbol{\zeta}^\ast, \boldsymbol{\epsilon}^\ast &= \argmin_{\boldsymbol{T}, \boldsymbol{\Theta}, \boldsymbol{\zeta}, \boldsymbol{\epsilon}} \bigg(\mathcal{L}(\boldsymbol{y}, \widehat{\boldsymbol{y}}(\boldsymbol{\mathcal{X}}; \boldsymbol{T}, \boldsymbol{\Theta}, \boldsymbol{\zeta}, \boldsymbol{\epsilon})) + \mathcal{L}_\mathrm{P} \\
&\quad + \frac{1}{D} (w_1 \ell_1 + w_2 \ell_2) + \frac{1}{M} (w_3 \ell_3 + w_4 \ell_4 + w_5 \ell_5)\bigg),
\end{aligned}
\end{equation}
where the empirical loss $\mathcal{L}$ is defined as
\[
\mathcal{L} = -\frac{1}{D} \sum_{i = 1}^K \sum_{\nu = 1}^D y_{i\nu} \log \frac{\exp(\widehat{y}_i(\boldsymbol{\chi}_\nu; \boldsymbol{T}, \boldsymbol{\Theta}, \boldsymbol{\zeta}, \boldsymbol{\epsilon}))}{\sum_{l = 1}^K \exp(\widehat{y}_l(\boldsymbol{\chi}_\nu; \boldsymbol{T}, \boldsymbol{\Theta}, \boldsymbol{\zeta}, \boldsymbol{\epsilon}))}
\]
for classification (cross-entropy), or
\[
\mathcal{L} = \frac{1}{K D} \sum_{i = 1}^K \sum_{\nu = 1}^D (y_{i\nu} - \widehat{y}_i(\boldsymbol{\chi}_\nu; \boldsymbol{T}, \boldsymbol{\Theta}, \boldsymbol{\zeta}, \boldsymbol{\epsilon}))^2
\]
for regression (mean squared error).
The term $\mathcal{L}_\mathrm{P}$ optionally imposes physical constraints that restrict the space of admissible solutions \citep{czarnecki2017sobolev, raissi2019physics, bastek2025physics}.

To prevent overfitting and produce interpretable solutions, we employ two structured regularizers:
\begin{equation}
\label{eq:l1_l2}
\ell_1 = \|\boldsymbol{\Theta}\|_2, \qquad
\ell_2 = \left\|\{\{\{g_{ij}(\boldsymbol{\chi}_\nu; \boldsymbol{T}, \boldsymbol{\Theta}, \boldsymbol{\zeta})\}_{\nu = 1}^D\}_{j = 1}^M\}_{i = 1}^K\right\|_2,
\end{equation}
where $\ell_1$ is the standard $L_2$ regularization of the network weights and biases \citep{krogh1991simple},\footnote{We opt for the unsquared norm $\|\cdot\|_2$, as opposed to $\|\cdot\|_2^2$, so that the penalty terms $\ell_1, \dots, \ell_5$ in Eq.~\eqref{eq:nn_loss} are similar in magnitude. This allows the regularization weights $w_1, \dots, w_5$ to be optimized jointly as a single hyperparameter $w$, drastically reducing the search space.} and $\ell_2$ discourages the bases $\{g_{ij}\}$ from taking on disproportionately large values \citep{agarwal2021neural}.
Given that the inputs and outputs are typically normalized, this constraint underpins the global input ranking of Section \ref{subsec:sr}, in which the columns of the linear transformation $\boldsymbol{T}$ serve as a proxy for predictive importance.

To further promote sparsity and disentanglement, we analyze the singular value decomposition (SVD) of $\boldsymbol{T}$:
\[
\boldsymbol{T} = \boldsymbol{Q}_1 \boldsymbol{\Sigma} \boldsymbol{Q}_2^\mathrm{T},
\]
where $\boldsymbol{Q}_1$ and $\boldsymbol{Q}_2$ are orthonormal matrices of dimensions $M \times M$ and $N \times N$, respectively, and $\boldsymbol{\Sigma}$ is an $M \times N$ diagonal matrix containing the $\min(M, N)$ singular values.
Drawing on the metric
\[
\Phi(\boldsymbol{R}_1, \boldsymbol{R}_2) = \left\|\boldsymbol{I} - \boldsymbol{R}_1 \boldsymbol{R}_2^\mathrm{T}\right\|_F = \sqrt{2 (3 - \mathrm{tr}(\boldsymbol{R}_1 \boldsymbol{R}_2^\mathrm{T}))}, \qquad
\boldsymbol{R}_1, \boldsymbol{R}_2 \in SO(3),
\]
used to measure the distance between two 3D rotations \citep{huynh2009metrics}, we define a joint penalty for the deviation of $\boldsymbol{Q}_1$ and $\boldsymbol{Q}_2$ from the identity matrix $\boldsymbol{I}$:
\begin{equation}
\label{eq:l3}
\ell_3 = \Phi(\boldsymbol{I}, \boldsymbol{Q}_1) + \Phi(\boldsymbol{I}, \boldsymbol{Q}_2) = \sqrt{2 (M - \mathrm{tr} \boldsymbol{Q}_1)} + \sqrt{2 (N - \mathrm{tr} \boldsymbol{Q}_2)}.
\end{equation}
This term effectively penalizes unnecessary couplings between the inputs; in isolation, $\ell_3$ is minimized when the transformation $\boldsymbol{T}$ is a rectangular diagonal matrix.
This shrinkage toward the axis-aligned NAM subclass---the most interpretable member of the PNAM family---acts as a structured prior: since $\ell_3$ is dominated by the empirical loss $\mathcal{L}$, cross-variable couplings are introduced only when they reduce $\mathcal{L}$ by more than the cost they incur.\footnote{\label{fn:l3_prime}An alternative penalty promoting orthogonality of the columns of $\boldsymbol{T}$, e.g., $\ell_3^\prime = \|\mathrm{offdiag}(\boldsymbol{T}^\mathrm{T} \boldsymbol{T})\|_F$, can be employed; it is cheaper to compute and avoids differentiating through the SVD. Appendix \ref{subsec:knot_classif} shows that replacing $\ell_3$ with $\ell_3^\prime$ yields the same accuracy but a slightly less stable feature ranking.}
Finally, we apply $L_1$ sparsity constraints on $\boldsymbol{T}$ and $\boldsymbol{\zeta}$ to drive nonessential interactions and bases to zero \citep{tibshirani1996regression, xu2023sparse, bahmani2024discovering}:
\begin{equation}
\label{eq:l4_l5}
\ell_4 = \|\boldsymbol{T}\|_1, \qquad
\ell_5 = \|\boldsymbol{\zeta}\|_1.
\end{equation}

\subsection{Post-processing and symbolic regression} \label{subsec:sr}
Following training, we apply three post-processing techniques to prune the PNAM and enhance interpretability \citep{murdoch2019definitions, cheng2024survey}.

The first technique exploits the regularizer $\ell_2$ in Eq.~\eqref{eq:l1_l2}.
The column-wise mean of the absolute entries in the linear transformation $\boldsymbol{T}$---defined as $\frac{1}{M} \sum_{j = 1}^M |T_{jk}|$ and termed the mean absolute coefficients---measures how strongly each input is used in the learned projections.
When the bases $\{g_{ij}\}$ are constrained by $\ell_2$ and the variables are normalized, $\{g_{ij}\}$ become comparable in magnitude across projections, so a column of $\boldsymbol{T}$ is large only if this reduces the empirical loss $\mathcal{L}$.
The mean absolute coefficients then serve as an empirical proxy for predictive importance (validated in Appendix \ref{subsec:knot_classif}) rather than a guaranteed property of the PNAM architecture.
We can thus retain only the top $n \leq N$ features and zero out the columns of $\boldsymbol{T}$ associated with the $(N - n)$ least significant inputs.
The second technique applies hard thresholding to the transformation $\boldsymbol{T}$ and weighting coefficients $\boldsymbol{\zeta}$, zeroing out entries where $|T_{jk}| < T_0$ and $|\zeta_{ij}| < \zeta_0$ for user-defined thresholds $T_0$ and $\zeta_0$.\footnote{These techniques may be used independently or in tandem. Like the choice of $M$ and the regularization weights $w_1, \dots, w_5$, the specific values for $n$, $T_0$, and $\zeta_0$ allow the user to navigate the Pareto front between accuracy and sparsity.}

Finally, the third technique leverages the single-variable nature of the bases $\{g_{ij}\}$ to convert them into explicit symbolic expressions.
While various SR algorithms are compatible (such as DSR \citep{petersen2019deep} or AI Feynman \citep{udrescu2020ai}), we employ PySR \citep{cranmer2023interpretable}, which utilizes genetic programming \citep{holland1992adaptation, koza1994genetic}, for its efficiency and ease of use.
To distill the $i^\mathrm{th}$ output of the $j^\mathrm{th}$ MLP into symbolic form, we minimize the following objective for fixed $i$ and $j$:
\begin{equation}
\label{eq:sr_loss}
\begin{aligned}
g_{ij}^\ast &= \argmin_{g_{ij}} \Bigg(\frac{1}{D} \sum_{\nu = 1}^D \Bigg((\zeta_{ij} \mathrm{MLP}_{ij}(z_j; \boldsymbol{\theta}_j)|_\nu - g_{ij}(z_j)|_\nu)^2 \\
&\quad + w_6 \bigg(\zeta_{ij} \frac{\mathrm{d} \mathrm{MLP}_{ij}(z_j; \boldsymbol{\theta}_j)}{\mathrm{d} z_j}\bigg|_\nu - \frac{\mathrm{d} g_{ij}(z_j)}{\mathrm{d} z_j}\bigg|_\nu\bigg)^2\Bigg)\Bigg),
\end{aligned}
\end{equation}
where $g_{ij}$ is a mathematical expression of a single variable $z_j$, and $w_6$ controls the weight placed on matching the derivatives of the learned MLP.
Summing the resulting $\{g_{ij}^\ast\}$ over $j$ and adding the bias $\epsilon_i$ recovers the global output $\widehat{y}_i$ as defined in Eq.~\eqref{eq:outputs_param}.
These post-processing steps can reduce tens of thousands of network parameters to as few as tens or hundreds, effectively bypassing the NP-hardness of multi-variable SR \citep{petersen2019deep, virgolin2022symbolic} while retaining the predictive power of deep neural networks.

\section{Numerical experiments} \label{sec:exp}
We illustrate the expressivity and interpretability of PNAMs through a benchmark study on an extensive dataset of mathematical knots from \citep{davies2021advancing}, covering (i) multi-class classification and (ii) single-output regression.
In both tasks, PNAMs recover the same key invariants identified by previous studies \citep{davies2021advancing, liu2025kan} and produce compact symbolic expressions without prior knowledge of the relevant invariants.
Additional results for knot theory are provided in Appendices \ref{subsec:knot_classif} and \ref{subsec:knot_regress}.
Experiments on the MNIST dataset \citep{lecun1998gradient}, provided in Appendix \ref{subsec:mnist}, further demonstrate the feature selection and dimensionality reduction capabilities of PNAMs for high-dimensional image classification.
Finally, we examine the benign overfitting behavior of PNAMs as the projection dimension $M$ increases using a phase field simulation of fracture propagation \citep{clayton2023rate} with a limited dataset (Appendix \ref{subsec:phase_field}).

Unless otherwise stated, we employ an 80/20 train-test split; the training set is further divided into training and validation subsets (80/20).
The PNAM is implemented in PyTorch \citep{paszke2019pytorch} using SiLU\footnote{While ReLU often yields a marginally lower empirical loss $\mathcal{L}$, the resulting bases $\{g_{ij}\}$ are less smooth, requiring higher complexity to approximate via Eq.~\eqref{eq:sr_loss}; SiLU provides the necessary smoothness for robust symbolic distillation.} \citep{hendrycks2016gaussian, elfwing2018sigmoid} as the activation function $A$ in Eq.~\eqref{eq:mlp}.
We initialize the projection dimension to $M = \min(8, N)$ and adjust it based on the complexity of the task.

Unless otherwise stated, the regularization weights $w_1, \dots, w_5$ in Eq.~\eqref{eq:nn_loss} are set to 0.01 for classification and 0.001 for regression tasks.\footnote{We find that these values provide a robust trade-off between accuracy and sparsity given the magnitude of $\mathcal{L}$ relative to the penalty terms $\ell_1, \dots, \ell_5$ for scaled variables.}
By default, the optional physical constraint term $\mathcal{L}_\mathrm{P}$ is omitted, and the regularization weight $w_6$ in Eq.~\eqref{eq:sr_loss} is set to zero.
We utilize a batch size of 256 samples\footnote{As is standard for mini-batch training, $\ell_1$ and $\ell_2$ in Eq.~\eqref{eq:nn_loss} are divided by the batch size rather than $D$.} and the Adam optimizer \citep{kingma2014adam} with an initial learning rate of 0.001 that decays by a factor of 0.995 per epoch to train all networks.
All training is conducted on a single NVIDIA A100-SXM4-40GB GPU.
Once trained, each basis $g_{ij}$ is evolved for one minute using 30 populations of 30 expressions with a maximum complexity of 30; all operators and leaf nodes are assigned a complexity of one.
Other relevant hyperparameters and training details are reported alongside the results.

\subsection{Benchmarking with knot theory} \label{subsec:knot_theory}
Established by a team of researchers mostly from Google DeepMind (DM) \citep{davies2021advancing}, the dataset of mathematical knots consists of 243,746 samples, each possessing 17 geometric invariants: adjoint torsion degree, torsion degree, short geodesic (real part), short geodesic (imaginary part), injectivity radius, Chern--Simons invariant, cusp volume, longitudinal translation, meridional translation (imaginary part), meridional translation (real part), volume, and six symmetry groups.
The goal is to use these 17 invariants to predict the knot signature, which takes on one of 14 discrete values (even integers ranging from $-12$ to 14).
To assess the versatility of PNAMs, we explore two distinct formulations: (i) a multi-class classification problem with 14 classes and (ii) a single-output regression problem.

\paragraph{Classification.}
The first two rows of Table \ref{tab:knot_theory_classif_compare} are reproduced from \citep{liu2025kan} and compare the performance of the MLP\footnote{\label{fn:dm_code}Inspections of the source code (\url{https://github.com/google-deepmind/mathematics_conjectures/blob/main/knot_theory.ipynb}) reveal that the original MLP training uses an early stopping patience of only one evaluation, which likely leads to underfitting.} implemented by \citet{davies2021advancing} against that of the KAN.
The subsequent rows present our implementations of the MLP, the NAM, and four parameterizations of the PNAM, all utilizing two hidden layers of 64 and 32 neurons.\footnote{The parameter counts for the MLP, the NAM, and the PNAM are estimated as $\mathcal{O}(L W^2)$, $\mathcal{O}(N L W^2)$, and $\mathcal{O}(M N + M L W^2)$, respectively, where $W$ is the number of neurons in the widest layer. Compared to an MLP of identical $L$ and $W$, the PNAM possesses approximately $M$ times more parameters and can be $M$ times slower to train. We intend to parallelize the evaluation of the bases in future work.}
We scale all inputs to have zero mean and unit variance and train the models for 50 epochs without early stopping.
All six models achieve test accuracies exceeding 90\%, with the NAM performing the worst and the PNAM with $M = 64$ matching our implementation of the MLP.
The accuracy of the PNAM increases monotonically with $M$ (from 8 to 64), consistent with Theorem \ref{thm:rate}.
Moreover, the accuracy gain of the PNAM over the NAM for all three values of $M$ in Table \ref{tab:knot_theory_classif_compare} suggests that the penalty $\ell_3$ in Eq.~\eqref{eq:l3} does not collapse the PNAM into a NAM in practice (see, e.g., the learned projection matrix $\boldsymbol{T}$ in Fig.~\ref{fig:knot_theory_regress}(b)).
Although the accuracy of PNAMs can be improved by increasing $M$ or by replacing the first hidden layer of the basis MLPs with a Fourier feature mapping (e.g., the last row of Table \ref{tab:knot_theory_classif_compare}, with $\boldsymbol{B}_j \in \mathbb{R}^{32}$ and $\sigma = 1$ in Eq.~\eqref{eq:fourier}), these modifications increase the complexity of the learned functions.

\begin{table}[!ht]
\caption{Performance comparison of different neural network architectures for the multi-class classification problem of predicting the signature of mathematical knots. The first two rows are reproduced from Table 3 of \citet{liu2025kan}; see \citep{liu2025kan} for definitions of $G$ and $k$. For the remaining rows, the mean and standard deviation of the test accuracy are computed over 10 runs. The last row incorporates the Fourier feature mapping described in Eq.~\eqref{eq:fourier}.}
\label{tab:knot_theory_classif_compare}
\centering
\begin{tabular}{llll}
\toprule
Method & Architecture & Parameter count & Test acc. \\
\midrule
DM's MLP & 4 layers: [17, 300, 300, 300, 14] & $3 \times 10^5$ & 78.0\% \\
KAN & 2 layers: [17, 1, 14] ($G = 3$, $k = 3$) & $2 \times 10^2$ & 81.6\% \\
\hline
Our MLP & 3 layers: [17, 64, 32, 14] & $1.2 \times 10^4$ & $95.8 \pm 0.1$\% \\
NAM & 3 layers: $17 \times [1, 64, 32, 14]$ & $2.1 \times 10^5$ & $92.4 \pm 0.2$\% \\
PNAM & 3 layers: $8 \times [1, 64, 32, 14]$ & $9.8 \times 10^4$ & $93.6 \pm 0.4$\% \\
PNAM & 3 layers: $17 \times [1, 64, 32, 14]$ & $2.1 \times 10^5$ & $95.0 \pm 0.2$\% \\
PNAM & 3 layers: $64 \times [1, 64, 32, 14]$ & $7.9 \times 10^5$ & $95.9 \pm 0.1$\% \\
PNAM & 3 layers: $8 \times [1, 2 (32), 32, 14]$ ($\sigma = 1$) & $9.8 \times 10^4$ & $94.3 \pm 0.3$\% \\
\bottomrule
\end{tabular}
\end{table}

\begin{table}[!ht]
\caption{Ranking of important features based on the mean absolute coefficients for the PNAM with $M = 8$ (without the Fourier feature mapping) in Table \ref{tab:knot_theory_classif_compare}. The mean and standard deviation of the test accuracy (associated with retaining only the top $n$ features) are computed over the runs in which the top $n$ features appeared.}
\label{tab:knot_theory_classif_rank}
\centering
\begin{tabular}{lllll}
\toprule
Rank & Input feature & Symbol & Frequency & Test acc. \\
\midrule
1 & Re(meridional translation) & $\chi_{10}$ & 10/10 & $55.1 \pm 3.5$\% \\
\hline
2 & Longitudinal translation & $\chi_8$ & 10/10 & $74.2 \pm 1.5$\% \\
\hline
\multirow{4}{*}{3} & Im(meridional translation) & $\chi_9$ & 7/10 & $78.1 \pm 2.4$\% \\
& Cusp volume & $\chi_7$ & 1/10 & 80.5\% \\
& Im(short geodesic) & $\chi_4$ & 1/10 & 73.5\% \\
& Volume & $\chi_{11}$ & 1/10 & 70.5\% \\
\bottomrule
\end{tabular}
\end{table}

Table \ref{tab:knot_theory_classif_rank} summarizes the three most important features identified by the mean absolute coefficients of the linear transformation $\boldsymbol{T}$ for the PNAM with $M = 8$ and without the Fourier feature mapping.
Across all 10 runs, the meridional translation (real part) and the longitudinal translation consistently have the largest and second largest mean absolute coefficients, respectively; the meridional translation (imaginary part) ranks third in seven runs.
The top two invariants are likewise ranked first and second in 10/10 runs for $M = 17$ and $M = 64$, indicating that the feature ranking remains stable as the capacity grows.
Retaining only the columns of $\boldsymbol{T}$ associated with the top $n = 3$ inputs in Table \ref{tab:knot_theory_classif_rank}, the PNAM still achieves a test accuracy of 78.1\% (refer to Fig.~\ref{fig:knot_theory_classif} for additional details).
These findings are consistent with Fig.~3 of \citet{davies2021advancing}\footnote{We note that \citet{davies2021advancing} swap the nomenclature for the real and imaginary parts of the meridional translation in their source code (see footnote \ref{fn:dm_code}) and figure.} and Fig.~4.3 of \citet{liu2025kan}.
Interestingly, PNAMs recover these key invariants through direct inspection of the projection weights, whereas \citet{davies2021advancing} employ gradient-based attribution \citep{sundararajan2017axiomatic} and \citet{liu2025kan} leverage a specific KAN architecture with a hidden dimension of one to extract feature importance.

\paragraph{Regression.}
\citet{davies2021advancing} and \citet{liu2025kan} leverage insights from the classification task to construct explicit mathematical expressions for the knot signature, now formulated as a single-output regression problem.
In Table \ref{tab:knot_theory_regress_eqs}, expression A corresponds to the equation handcrafted by \citet{davies2021advancing}, while expressions B to F are derived from post-processing steps of KANs trained specifically on the three most relevant invariants \citep{liu2025kan}.
Although \citet{davies2021advancing} originally report a test accuracy between 70 and 80\% for expression A, \citet{liu2025kan} report a test accuracy of 83.1\%.
Since the accuracy appears sensitive to the random data split employed, we evaluate all candidate expressions on our specific test set and the entire knot dataset (denoted as ``Our'' and ``Total acc.'' in Table \ref{tab:knot_theory_regress_eqs}), in addition to the values reported by \citet{liu2025kan} for expressions A to F.

We demonstrate that the PNAM with regularization weight $w = 0.001$ in Table \ref{tab:knot_theory_regress_compare}---trained on the full set of invariants without prior knowledge of their relevance---can be distilled into symbolic expressions with as few as tens of parameters.
We emphasize that expressions G to I in Table \ref{tab:knot_theory_regress_eqs} emerge from the PNAM pipeline after pruning and SR; their particular forms depend on the available data and would likely change as more invariants are considered.
The key contribution here is the ability of PNAMs to discover relevant relationships as new data and invariants are introduced.

Among the results summarized in Table \ref{tab:knot_theory_regress_rank}, expressions H and I are obtained from the run depicted in Fig.~\ref{fig:knot_theory_regress}, using $n = 3$ features with the cusp volume\footnote{The cusp volume is equivalent to the product of the longitudinal translation and the imaginary part of the meridional translation (see Fig.~4.4(b) of \citet{liu2025kan}).} ranked as the third most significant.
Although the cusp volume is not explicitly identified in \citep{davies2021advancing, liu2025kan} as a primary invariant for predictive accuracy, the PNAM independently identifies it as an invariant that potentially enhances performance.
Moreover, comparing expressions E and G suggests that PNAMs can match KANs in accuracy while utilizing fewer invariants without relying on additional assumptions.
For an analysis of how the regularization weights in Eq.~\eqref{eq:nn_loss} influence performance, see Appendix \ref{subsec:knot_regress}.

\section{Limitations} \label{sec:limit}

\paragraph{Interpretability scope.}
Although PNAMs provide global interpretability, they do not offer local interpretability.
Because the bases $\{g_{ij}\}$ operate on the projected variables $\{z_j\}$ rather than the original inputs $\{\chi_k\}$, examining $g_{ij}$ in isolation does not directly reveal the contribution of a specific input $\chi_k$ to the output.
Furthermore, the mean absolute coefficients of the linear transformation $\boldsymbol{T}$ measure feature usage in the learned projections, and they serve as a proxy for predictive importance only when the regularizer $\ell_2$ in Eq.~\eqref{eq:l1_l2} is active and appropriately calibrated.
In the absence of such a constraint, the resulting ranking may be unstable or erroneous (see, e.g., Fig.~\ref{fig:knot_theory_regress_w}(a)).
Even with this constraint, the ranking is an empirical property of the regularized models, validated in the settings evaluated here, as opposed to a general guarantee of the architecture.
Finally, we emphasize that the PNAM is a predictive, rather than a causal, model.
Large mean absolute coefficients indicate a strong predictive influence but do not imply a causal relationship, particularly when the model is trained on highly correlated inputs.

An alternative approach to interpretability combines MLPs with post-hoc attribution, pruning, or symbolic distillation.
Attribution explains the predictions and pruning reduces the model size, but the model itself remains a black box.
Symbolic distillation, moreover, amounts to the multi-variable SR problem whose intractability motivates our design.
The PNAM instead provides an integrated pipeline in which the same trained model supports feature ranking, pruning, and conversion into a closed-form expression, because the additive structure reduces SR to independent 1D problems.
However, the GAM/NAM extensions with explicit interaction terms \citep{lou2013accurate, chang2021node, yang2021gami, dubey2022scalable, enouen2022sparse} are more directly interpretable when the ground truth is dominated by a few low-order interactions of raw features.

\paragraph{Optimization and training.}
The universal approximation and rate results of Section \ref{subsec:approx} are only existential: they assert that accurate transformations and bases exist, not that gradient-based optimization recovers them---a characteristic shared with classical universal approximation theorems for MLPs \citep{hornik1989multilayer, leshno1993multilayer}.
Two observations nevertheless connect the theory to the learned models.
First, the transformation achieving an exact representation is highly nonunique (e.g., Example \ref{ex:two_dim}), so the optimization target is a large set of admissible transformations rather than a single isolated construction.
Second, performance improves monotonically with the projection dimension $M$ on average across all our experiments (see Tables \ref{tab:knot_theory_classif_compare}, \ref{tab:mnist_compare}, and \ref{tab:phase_field_compare}), tracking the capacity scaling predicted by Theorem \ref{thm:rate}.

In practice, the joint optimization over $\boldsymbol{T}$, $\boldsymbol{\Theta}$, $\boldsymbol{\zeta}$, and $\boldsymbol{\epsilon}$ in Eq.~\eqref{eq:nn_loss} is non-convex, and performance is sensitive to both $M$ and the regularization weights $w_1, \dots, w_5$.
Over-regularization leads to sparse but inaccurate models, while under-regularization results in dense projections that are significantly harder to interpret.
In addition, the SR step in Section \ref{subsec:sr} is run for only one minute per basis in the numerical experiments.
While this ensures computational efficiency, it may be insufficient for capturing the nuances of highly complex functions.
Consequently, the resulting symbolic expressions are approximations and should be rigorously validated before use in safety-critical applications.

\section{Conclusion} \label{sec:concl}
In this work, we establish the first theoretical foundation for PNAMs, proving universal approximation, characterizing the minimum number of projections $M$, and deriving an approximation rate of $\mathcal{O}(M^{-s / (N - 1)})$ for $\mathcal{C}^s$ functions.
Building on this foundation, we introduce structured regularization, input pruning, and symbolic regression techniques to recover sparse and interpretable models.
Across a diverse set of experiments, PNAMs consistently match or surpass similar architectures in predictive accuracy while promoting sparsity.
In the knot theory benchmark, PNAMs recover the same key invariants identified by prior studies and produce compact symbolic expressions with orders of magnitude fewer parameters than the trained networks.
Additional experiments on MNIST and phase field fracture mechanics demonstrate the feature selection capability of PNAMs for high-dimensional image classification and their robustness in the limited-data regime, respectively.
As a stand-alone model, the PNAM is not meant to replace the MLP, the NAM, or classical SR.
Instead, it offers a principled middle ground: achieving the accuracy of the MLP while retaining the additive structure of the NAM, at the cost of increased memory and training time proportional to $M$.
Future work includes parallelizing the evaluation of the 1D bases to reduce this overhead and exploring the use of PNAMs as backbones in graph neural networks and transformers, where their modularity could enable structural pruning of network parameters and accelerate inference.

\begin{ack}
NNP is supported by the SMART Scholarship Program and the SMART SEED Innovation Award, both administered by the U.S.~Army DEVCOM Army Research Laboratory.
WCS is supported by the Department of Energy (DOE), National Nuclear Security Administration, Predictive Science Academic Alliance Program (PSAAP) under Award Number DE-NA0003962.
Additional support is provided for NNP and WCS through the Dynamic Materials and Interactions Program from the Air Force Office of Scientific Research under grant contract FA9550-22-1-0310 and through the Army Research Office under grant contract W911NF-24-1-0116.
QD is supported in part by the DOE-ASCR award DE-SC0025347 and the National Science Foundation (NSF) award DMS-2309245.
Support is also provided by the NSF under grant contract CMMI-1846875 for the high-performance computing facility.
All of this support is gratefully acknowledged.
The views and conclusions contained in this document are those of the authors and should not be interpreted as representing the official policies, either expressed or implied, of the sponsors, including the Army Research Laboratory or the U.S.~Government.
The U.S.~Government is authorized to reproduce and distribute reprints for Government purposes, notwithstanding any copyright notation herein.
The authors declare that they have no known competing financial interests or personal relationships that could have appeared to influence the work reported in this paper.
\end{ack}

\bibliographystyle{unsrtnat}
\bibliography{neurips_2026}

@inproceedings{liu2025kan,
	author = {Liu, Z. and Wang, Y. and Vaidya, S. and Ruehle, F. and Halverson, J. and Soljacic, M. and Hou, T. and Tegmark, M.},
	booktitle = {International Conference on Learning Representations},
	pages = {70367--70413},
	title = {{KAN}: {Kolmogorov--Arnold} networks},
	year = {2025}}

@inproceedings{bastek2025physics,
	author = {Bastek, J.-H. and Sun, W.C. and Kochmann, D.},
	booktitle = {International Conference on Learning Representations},
	pages = {3360--3385},
	title = {Physics-informed diffusion models},
	year = {2025}}

@article{enouen2022sparse,
	author = {Enouen, J. and Liu, Y.},
	journal = {Advances in Neural Information Processing Systems},
	pages = {13908--13920},
	title = {Sparse interaction additive networks via feature interaction detection and sparse selection},
	volume = {35},
	year = {2022}}

@article{dubey2022scalable,
	author = {Dubey, A. and Radenovic, F. and Mahajan, D.},
	journal = {Advances in Neural Information Processing Systems},
	pages = {36748--36761},
	title = {Scalable interpretability via polynomials},
	volume = {35},
	year = {2022}}

@article{chang2021node,
	author = {Chang, C.-H. and Caruana, R. and Goldenberg, A.},
	journal = {arXiv preprint arXiv:2106.01613},
	title = {{NODE-GAM}: Neural generalized additive model for interpretable deep learning},
	year = {2021}}

@article{yang2021gami,
	author = {Yang, Z. and Zhang, A. and Sudjianto, A.},
	journal = {Pattern Recognition},
	pages = {108192},
	publisher = {Elsevier},
	title = {{GAMI-Net}: An explainable neural network based on generalized additive models with structured interactions},
	volume = {120},
	year = {2021}}

@inproceedings{lou2013accurate,
	author = {Lou, Y. and Caruana, R. and Gehrke, J. and Hooker, G.},
	booktitle = {Proceedings of the 19th ACM SIGKDD International Conference on Knowledge Discovery and Data Mining},
	pages = {623--631},
	title = {Accurate intelligible models with pairwise interactions},
	year = {2013}}

@article{yarotsky2017error,
	author = {Yarotsky, D.},
	journal = {Neural Networks},
	pages = {103--114},
	publisher = {Elsevier},
	title = {Error bounds for approximations with deep {ReLU} networks},
	volume = {94},
	year = {2017}}

@book{devore1993constructive,
	address = {Berlin},
	author = {DeVore, R. A. and Lorentz, G. G.},
	publisher = {Springer},
	series = {A Series of Comprehensive Studies in Mathematics},
	title = {Constructive Approximation},
	volume = {303},
	year = {1993}}

@article{maiorov1999best,
	author = {Maiorov, V. E.},
	journal = {Journal of Approximation Theory},
	number = {1},
	pages = {68--94},
	publisher = {Elsevier},
	title = {On best approximation by ridge functions},
	volume = {99},
	year = {1999}}

@article{alexander1995polynomial,
	author = {Alexander, J. and Hirschowitz, A.},
	journal = {Journal of Algebraic Geometry},
	number = {2},
	pages = {201--222},
	publisher = {Providence, RI: University Press, c1992-},
	title = {Polynomial interpolation in several variables},
	volume = {4},
	year = {1995}}

@book{landsberg2012tensors,
	address = {Providence, RI},
	author = {Landsberg, J. M.},
	publisher = {American Mathematical Society},
	series = {Graduate Studies in Mathematics},
	title = {Tensors: Geometry and Applications},
	volume = {128},
	year = {2012}}

@article{barata2012moore,
	author = {Barata, J. C. A. and Hussein, M. S.},
	journal = {Brazilian Journal of Physics},
	pages = {146--165},
	publisher = {Springer},
	title = {The {Moore--Penrose} pseudoinverse: A tutorial review of the theory},
	volume = {42},
	year = {2012}}

@article{odrzywolek2026all,
	author = {Odrzywo{\l}ek, A.},
	journal = {arXiv preprint arXiv:2603.21852},
	title = {All elementary functions from a single binary operator},
	year = {2026}}

@article{pinkus1999approximation,
	author = {Pinkus, A.},
	journal = {Acta Numerica},
	pages = {143--195},
	publisher = {Cambridge University Press},
	title = {Approximation theory of the {MLP} model in neural networks},
	volume = {8},
	year = {1999}}

@article{lecun1998gradient,
	author = {LeCun, Y. and Bottou, L. and Bengio, Y. and Haffner, P.},
	journal = {Proceedings of the IEEE},
	number = {11},
	pages = {2278--2324},
	publisher = {Ieee},
	title = {Gradient-based learning applied to document recognition},
	volume = {86},
	year = {1998}}

@book{molnar2019interpretable,
	author = {Molnar, C.},
	publisher = {Leanpub},
	title = {Interpretable Machine Learning: A Guide for Making Black Box Models Explainable},
	year = {2019}}

@article{rudin2022interpretable,
	author = {Rudin, C. and Chen, C. and Chen, Z. and Huang, H. and Semenova, L. and Zhong, C.},
	journal = {Statistics Surveys},
	pages = {1--85},
	publisher = {The American Statistical Association, the Bernoulli Society, the Institute~{\ldots}},
	title = {Interpretable machine learning: Fundamental principles and 10 grand challenges},
	volume = {16},
	year = {2022}}

@article{radenovic2022neural,
	author = {Radenovic, F. and Dubey, A. and Mahajan, D.},
	journal = {Advances in Neural Information Processing Systems},
	number = {612},
	pages = {8414--8426},
	title = {Neural basis models for interpretability},
	volume = {35},
	year = {2022}}

@article{hastie1986generalized,
	author = {Hastie, T. and Tibshirani, R.},
	journal = {Statistical Science},
	number = {3},
	pages = {297--310},
	publisher = {Institute of Mathematical Statistics},
	title = {Generalized additive models},
	volume = {1},
	year = {1986}}

@article{mallinar2022benign,
	author = {Mallinar, N. and Simon, J. and Abedsoltan, A. and Pandit, P. and Belkin, M. and Nakkiran, P.},
	journal = {Advances in Neural Information Processing Systems},
	number = {87},
	pages = {1182--1195},
	title = {Benign, tempered, or catastrophic: A taxonomy of overfitting},
	volume = {35},
	year = {2022}}

@article{miehe2010thermodynamically,
	author = {Miehe, C. and Welschinger, F. and Hofacker, M.},
	journal = {International Journal for Numerical Methods in Engineering},
	number = {10},
	pages = {1273--1311},
	publisher = {Wiley Online Library},
	title = {Thermodynamically consistent phase-field models of fracture: Variational principles and multi-field {FE} implementations},
	volume = {83},
	year = {2010}}

@article{davies2024signature,
	author = {Davies, A. and Juh{\'a}sz, A. and Lackenby, M. and Toma{\v{s}}ev, N.},
	journal = {Geometry \& Topology},
	number = {5},
	pages = {2313--2343},
	publisher = {Mathematical Sciences Publishers},
	title = {The signature and cusp geometry of hyperbolic knots},
	volume = {28},
	year = {2024}}

@inproceedings{fang2023depgraph,
	author = {Fang, G. and Ma, X. and Song, M. and Mi, M. B. and Wang, X.},
	booktitle = {Proceedings of the IEEE/CVF Conference on Computer Vision and Pattern Recognition},
	pages = {16091--16101},
	title = {{DepGraph}: Towards any structural pruning},
	year = {2023}}

@article{brunton2016discovering,
	author = {Brunton, S. L. and Proctor, J. L. and Kutz, J. N.},
	journal = {Proceedings of the National Academy of Sciences},
	number = {15},
	pages = {3932--3937},
	publisher = {National Academy of Sciences},
	title = {Discovering governing equations from data by sparse identification of nonlinear dynamical systems},
	volume = {113},
	year = {2016}}

@article{cranmer2020discovering,
	author = {Cranmer, M. and Sanchez-Gonzalez, A. and Battaglia, P. and Xu, R. and Cranmer, K. and Spergel, D. and Ho, S.},
	journal = {Advances in Neural Information Processing Systems},
	number = {1462},
	pages = {17429--17442},
	title = {Discovering symbolic models from deep learning with inductive biases},
	volume = {33},
	year = {2020}}

@article{chrysos2021deep,
	author = {Chrysos, G. G. and Moschoglou, S. and Bouritsas, G. and Deng, J. and Panagakis, Y. and Zafeiriou, S.},
	journal = {IEEE Transactions on Pattern Analysis and Machine Intelligence},
	number = {8},
	pages = {4021--4034},
	publisher = {IEEE},
	title = {Deep polynomial neural networks},
	volume = {44},
	year = {2022}}

@article{gurtin1996generalized,
	author = {Gurtin, M. E.},
	journal = {Physica D: Nonlinear Phenomena},
	number = {3-4},
	pages = {178--192},
	publisher = {Elsevier},
	title = {Generalized {Ginzburg--Landau} and {Cahn--Hilliard} equations based on a microforce balance},
	volume = {92},
	year = {1996}}

@inproceedings{sundararajan2017axiomatic,
	author = {Sundararajan, M. and Taly, A. and Yan, Q.},
	booktitle = {International Conference on Machine Learning},
	organization = {PMLR},
	pages = {3319--3328},
	title = {Axiomatic attribution for deep networks},
	volume = {70},
	year = {2017}}

@article{elfwing2018sigmoid,
	author = {Elfwing, S. and Uchibe, E. and Doya, K.},
	journal = {Neural Networks},
	pages = {3--11},
	publisher = {Elsevier},
	title = {Sigmoid-weighted linear units for neural network function approximation in reinforcement learning},
	volume = {107},
	year = {2018}}

@article{paszke2019pytorch,
	author = {Paszke, A. and Gross, S. and Massa, F. and Lerer, A. and Bradbury, J. and Chanan, G. and Killeen, T. and Lin, Z. and Gimelshein, N. and Antiga, L. and others},
	journal = {Advances in Neural Information Processing Systems},
	number = {721},
	pages = {8026--8037},
	title = {{PyTorch}: An imperative style, high-performance deep learning library},
	volume = {32},
	year = {2019}}

@article{hendrycks2016gaussian,
	author = {Hendrycks, D. and Gimpel, K.},
	journal = {arXiv preprint arXiv:1606.08415},
	title = {Gaussian error linear units ({GELUs})},
	year = {2016}}

@article{kingma2014adam,
	author = {Kingma, D. P. and Ba, J.},
	journal = {arXiv preprint arXiv:1412.6980},
	title = {Adam: A method for stochastic optimization},
	year = {2014}}

@article{virgolin2022symbolic,
	author = {Virgolin, M. and Pissis, S. P.},
	journal = {arXiv preprint arXiv:2207.01018},
	title = {Symbolic regression is {NP}-hard},
	year = {2022}}

@article{koza1994genetic,
	author = {Koza, J. R.},
	journal = {Statistics and Computing},
	pages = {87--112},
	publisher = {Springer},
	title = {Genetic programming as a means for programming computers by natural selection},
	volume = {4},
	year = {1994}}

@book{holland1992adaptation,
	address = {Cambridge, MA},
	author = {Holland, J. H.},
	publisher = {MIT Press},
	title = {Adaptation in Natural and Artificial Systems: An Introductory Analysis with Applications to Biology, Control, and Artificial Intelligence},
	year = {1992}}

@article{udrescu2020ai,
	author = {Udrescu, S.-M. and Tegmark, M.},
	journal = {Science Advances},
	number = {16},
	pages = {eaay2631},
	publisher = {American Association for the Advancement of Science},
	title = {{AI Feynman}: A physics-inspired method for symbolic regression},
	volume = {6},
	year = {2020}}

@article{petersen2019deep,
	author = {Petersen, B. K. and Landajuela, M. and Mundhenk, T. N. and Santiago, C. P. and Kim, S. K. and Kim, J. T.},
	journal = {arXiv preprint arXiv:1912.04871},
	title = {Deep symbolic regression: Recovering mathematical expressions from data via risk-seeking policy gradients},
	year = {2019}}

@article{cheng2024survey,
	author = {Cheng, H. and Zhang, M. and Shi, J. Q.},
	journal = {IEEE Transactions on Pattern Analysis and Machine Intelligence},
	number = {12},
	pages = {10558--10578},
	publisher = {IEEE},
	title = {A survey on deep neural network pruning: Taxonomy, comparison, analysis, and recommendations},
	volume = {46},
	year = {2024}}

@article{murdoch2019definitions,
	author = {Murdoch, W. J. and Singh, C. and Kumbier, K. and Abbasi-Asl, R. and Yu, B.},
	journal = {Proceedings of the National Academy of Sciences},
	number = {44},
	pages = {22071--22080},
	publisher = {National Academy of Sciences},
	title = {Definitions, methods, and applications in interpretable machine learning},
	volume = {116},
	year = {2019}}

@article{tibshirani1996regression,
	author = {Tibshirani, R.},
	journal = {Journal of the Royal Statistical Society Series B: Statistical Methodology},
	number = {1},
	pages = {267--288},
	publisher = {Oxford University Press},
	title = {Regression shrinkage and selection via the lasso},
	volume = {58},
	year = {1996}}

@article{krogh1991simple,
	author = {Krogh, A. and Hertz, J.},
	journal = {Advances in Neural Information Processing Systems},
	pages = {950--957},
	title = {A simple weight decay can improve generalization},
	volume = {4},
	year = {1991}}

@article{clayton2023rate,
	author = {Clayton, J. D. and Knap, J. and Leavy, R. B.},
	journal = {Mechanics of Materials},
	pages = {104606},
	publisher = {Elsevier},
	title = {On rate dependence and anisotropy in phase field modeling of polycrystalline fracture},
	volume = {180},
	year = {2023}}

@article{davies2021advancing,
	author = {Davies, A. and Veli{\v{c}}kovi{\'c}, P. and Buesing, L. and Blackwell, S. and Zheng, D. and Toma{\v{s}}ev, N. and Tanburn, R. and Battaglia, P. and Blundell, C. and Juh{\'a}sz, A. and others},
	journal = {Nature},
	pages = {70--74},
	publisher = {Nature Publishing Group UK London},
	title = {Advancing mathematics by guiding human intuition with {AI}},
	volume = {600},
	year = {2021}}

@article{cranmer2023interpretable,
	author = {Cranmer, M.},
	journal = {arXiv preprint arXiv:2305.01582},
	title = {Interpretable machine learning for science with {PySR} and {SymbolicRegression.jl}},
	year = {2023}}

@article{huynh2009metrics,
	author = {Huynh, D. Q.},
	journal = {Journal of Mathematical Imaging and Vision},
	pages = {155--164},
	publisher = {Springer},
	title = {Metrics for {3D} rotations: Comparison and analysis},
	volume = {35},
	year = {2009}}

@article{raissi2019physics,
	author = {Raissi, M. and Perdikaris, P. and Karniadakis, G. E.},
	journal = {Journal of Computational Physics},
	pages = {686--707},
	publisher = {Elsevier},
	title = {Physics-informed neural networks: A deep learning framework for solving forward and inverse problems involving nonlinear partial differential equations},
	volume = {378},
	year = {2019}}

@article{czarnecki2017sobolev,
	author = {Czarnecki, W. M. and Osindero, S. and Jaderberg, M. and Swirszcz, G. and Pascanu, R.},
	journal = {Advances in Neural Information Processing Systems},
	pages = {4281--4290},
	title = {Sobolev training for neural networks},
	volume = {30},
	year = {2017}}

@inproceedings{xu2023sparse,
	author = {Xu, S. and Bu, Z. and Chaudhari, P. and Barnett, I. J.},
	booktitle = {Joint European Conference on Machine Learning and Knowledge Discovery in Databases},
	organization = {Springer},
	pages = {343--359},
	title = {Sparse neural additive model: Interpretable deep learning with feature selection via group sparsity},
	volume = {3},
	year = {2023}}

@techreport{turner1966inverse,
	author = {Turner, L. R.},
	institution = {NASA Lewis Research Center, Cleveland, OH (United States)},
	title = {Inverse of the {Vandermonde} matrix with applications},
	year = {1966}}

@article{macon1958inverses,
	author = {Macon, N. and Spitzbart, A.},
	journal = {The American Mathematical Monthly},
	number = {2},
	pages = {95--100},
	publisher = {Taylor \& Francis},
	title = {Inverses of {Vandermonde} matrices},
	volume = {65},
	year = {1958}}

@article{cotter1990stone,
	author = {Cotter, N. E.},
	journal = {IEEE Transactions on Neural Networks},
	number = {4},
	pages = {290--295},
	publisher = {IEEE},
	title = {The {Stone--Weierstrass} theorem and its application to neural networks},
	volume = {1},
	year = {1990}}

@article{stone1948generalized,
	author = {Stone, M. H.},
	journal = {Mathematics Magazine},
	number = {4},
	pages = {167--184},
	publisher = {JSTOR},
	title = {The generalized {Weierstrass} approximation theorem},
	volume = {21},
	year = {1948}}

@article{stone1937applications,
	author = {Stone, M. H.},
	journal = {Transactions of the American Mathematical Society},
	number = {3},
	pages = {375--481},
	publisher = {JSTOR},
	title = {Applications of the theory of {Boolean} rings to general topology},
	volume = {41},
	year = {1937}}

@article{bahmani2024discovering,
	author = {Bahmani, B. and Suh, H. S. and Sun, W.C.},
	journal = {Computer Methods in Applied Mechanics and Engineering},
	pages = {116827},
	publisher = {Elsevier},
	title = {Discovering interpretable elastoplasticity models via the neural polynomial method enabled symbolic regressions},
	volume = {422},
	year = {2024}}

@article{tancik2020fourier,
	author = {Tancik, M. and Srinivasan, P. and Mildenhall, B. and Fridovich-Keil, S. and Raghavan, N. and Singhal, U. and Ramamoorthi, R. and Barron, J. and Ng, R.},
	journal = {Advances in Neural Information Processing Systems},
	number = {632},
	pages = {7537--7547},
	title = {Fourier features let networks learn high frequency functions in low dimensional domains},
	volume = {33},
	year = {2020}}

@article{leshno1993multilayer,
	author = {Leshno, M. and Lin, V. Y. and Pinkus, A. and Schocken, S.},
	journal = {Neural Networks},
	number = {6},
	pages = {861--867},
	publisher = {Elsevier},
	title = {Multilayer feedforward networks with a nonpolynomial activation function can approximate any function},
	volume = {6},
	year = {1993}}

@article{lu2017expressive,
	author = {Lu, Z. and Pu, H. and Wang, F. and Hu, Z. and Wang, L.},
	journal = {Advances in Neural Information Processing Systems},
	pages = {6232--6240},
	title = {The expressive power of neural networks: A view from the width},
	volume = {30},
	year = {2017}}

@article{hornik1989multilayer,
	author = {Hornik, K. and Stinchcombe, M. and White, H.},
	journal = {Neural Networks},
	number = {5},
	pages = {359--366},
	publisher = {Elsevier},
	title = {Multilayer feedforward networks are universal approximators},
	volume = {2},
	year = {1989}}

@article{agarwal2021neural,
	author = {Agarwal, R. and Melnick, L. and Frosst, N. and Zhang, X. and Lengerich, B. and Caruana, R. and Hinton, G. E.},
	journal = {Advances in Neural Information Processing Systems},
	number = {359},
	pages = {4699--4711},
	title = {Neural additive models: Interpretable machine learning with neural nets},
	volume = {34},
	year = {2021}}

@article{phan2025hydra,
	author = {Phan, N. N. and Sun, W.C. and Clayton, J. D.},
	journal = {Computer Methods in Applied Mechanics and Engineering},
	pages = {117792},
	publisher = {Elsevier},
	title = {{HYDRA}: Symbolic feature engineering of overparameterized {Eulerian} hyperelasticity models for fast inference time},
	volume = {437},
	year = {2025}}


\appendix
\section{Appendix}
\setcounter{figure}{0}\renewcommand{\thefigure}{A.\arabic{figure}}\renewcommand{\theHfigure}{A.\arabic{figure}}
\setcounter{table}{0}\renewcommand{\thetable}{A.\arabic{table}}\renewcommand{\theHtable}{A.\arabic{table}}
\setcounter{equation}{0}\renewcommand{\theequation}{A.\arabic{equation}}\renewcommand{\theHequation}{A.\arabic{equation}}

\subsection{Polynomial reproducing} \label{subsec:reprod}
In this section, we establish the polynomial reproducing property, namely, that any multi-variable polynomial $\mathcal{F}$ of $\boldsymbol{\chi} \in \mathbb{R}^N$ can be represented by a superposition of ridge functions.
Specifically, we show that there exists a linear transformation $\boldsymbol{T}$ such that $\mathcal{F}(\boldsymbol{\chi}) = \sum_{i = 1}^M g_i(z_i)$, where each $g_i$ is a single-variable polynomial of the projected input $z_i = \sum_{j = 1}^N T_{ij} \chi_j$.
This result is proved for arbitrary dimensions via mathematical induction on the input dimension $N$, starting with the base case for two dimensions ($N = 2$).

To facilitate the inductive step, we first present a lemma demonstrating the reproducing property for a specific bivariate case.
This lemma provides the necessary constructive foundation for the general $N$-dimensional result that follows.
\begin{lemma}[Monomial representation]
\label{lem:monomial}
For any bivariate monomial $\chi_1^p \chi_2^q$, there exist a linear transformation $\boldsymbol{T}: \mathbb{R}^2 \rightarrow \mathbb{R}^M$, where $M \leq p + q + 1$, with $T_{i1} = 1$ and $T_{i2} = c_i$, and a set of coefficients $\{\zeta_i\}_{i = 1}^M$ such that
\begin{equation}
\label{eq:monomial}
\chi_1^p \chi_2^q = \sum_{i = 1}^M \zeta_i \left(\sum_{j = 1}^2 T_{ij} \chi_j\right)^{p + q} = \sum_{i = 1}^M \zeta_i (\chi_1 + c_i \chi_2)^{p + q}.
\end{equation}
\end{lemma}

\begin{proof}
Applying the binomial expansion to each term on the right-hand side of Eq.~\eqref{eq:monomial} and reordering the summations, we obtain
\[
\begin{aligned}
\sum_{i = 1}^M \zeta_i (\chi_1 + c_i \chi_2)^{p + q} &= \sum_{i = 1}^M \zeta_i \sum_{m = 0}^{p + q} \binom{p + q}{m} \chi_1^m (c_i \chi_2)^{p + q - m} \\
&= \sum_{m = 0}^{p + q} \binom{p + q}{m} \left(\sum_{i = 1}^M \zeta_i c_i^{p + q - m}\right) \chi_1^m \chi_2^{p + q - m}.
\end{aligned}
\]
To satisfy the equality in Eq.~\eqref{eq:monomial}, the coefficients of the resulting polynomial must match the target monomial $\chi_1^p \chi_2^q$.
Specifically, the coefficient must be 1 when $m = p$ and 0 otherwise.
This leads to the following system of $M$ linear equations for $\{\zeta_i\}$:
\[
\binom{p + q}{m} \sum_{i = 1}^M \zeta_i c_i^{p + q - m} = \widetilde{\delta}_{mp}, \qquad
m \in \{0, 1, \dots, p + q\},
\]
where $\widetilde{\delta}_{mp}$ is the Kronecker delta.
Written in matrix form, this system is
\[
\begin{bmatrix}
c_1^{p + q} & c_2^{p + q} & \hdots & c_M^{p + q} \\
c_1^{p + q - 1} & c_2^{p + q - 1} & \hdots & c_M^{p + q - 1} \\
\vdots & \vdots & \ddots & \vdots \\
1 & 1 & \hdots & 1
\end{bmatrix}
\begin{bmatrix}
\zeta_1 \\
\zeta_2 \\
\vdots \\
\zeta_M
\end{bmatrix} = \begin{bmatrix}
0 \\
\vdots \\
\binom{p + q}{p}^{-1} \\
\vdots \\
0
\end{bmatrix}.
\]
We note that the coefficient matrix forms a so-called Vandermonde matrix \citep{macon1958inverses, turner1966inverse}.
It is a standard result that a Vandermonde matrix is invertible if and only if the constants $\{c_i\}_{i = 1}^M$ are distinct.
By choosing $M = p + q + 1$ mutually different values for $\{c_i\}$, we guarantee a unique solution for the coefficients $\{\zeta_i\}$.
This establishes that $M = p + q + 1$ projections suffice, thus proving the bound $M \leq p + q + 1$.
\end{proof}

Even for this simple case, we may observe the nonuniqueness in the construction of the linear transformation $\boldsymbol{T}$.
Consequently, it is reasonable to expect that for specific machine learning applications, $\boldsymbol{T}$ may be optimized to achieve effective representations in a reduced-dimensional feature space.
This potential for dimensionality reduction is illustrated in the following example.
\begin{example}[Dimensionality reduction]
\label{ex:two_dim}
We present instances where the required projection dimension $M$ is strictly smaller than the upper bound $p + q + 1$ established in Lemma \ref{lem:monomial}.
For $p = q = 1$, a variation of the polarization identity (see Theorem \ref{thm:polar}) gives
\[
\chi_1 \chi_2 = \left(\chi_1 + \frac{1}{4} \chi_2\right)^2 - \left(\chi_1 - \frac{1}{4} \chi_2\right)^2,
\]
which corresponds to $M = 2$, whereas the lemma suggests an upper bound of 3.
In the trivial case where $p = 0$ or $q = 0$, the monomial is univariate, and we may simply use the identity transformation with $M = 1$.
Furthermore, $M$ can be smaller than the input dimension $N$.
For example, the quadratic form $\chi_1^2 + 2 \chi_1 \chi_2 + \chi_2^2$ can be represented as $z_1^2$ with $z_1 = \chi_1 + \chi_2$, yielding $M = 1 < 2 = N$.
These cases demonstrate the efficacy of utilizing a learned transformation to achieve dimensionality reduction through an appropriate latent space.
\end{example}

Next, we consider the extension of Lemma \ref{lem:monomial}.
Since a product of two single-variable polynomials $f_1(\chi_1)$ and $f_2(\chi_2)$ can be expressed as a linear combination of bivariate monomials $\{\chi_1^p \chi_2^q\}$, Lemma \ref{lem:monomial} can be generalized to represent polynomials with a product form.
\begin{lemma}[Product-form representation]
\label{lem:product}
For any bivariate polynomial of $(\chi_1, \chi_2) \in \mathbb{R}^2$ having the product form $f_1(\chi_1) f_2(\chi_2)$, there exist a linear transformation $\boldsymbol{T}: \mathbb{R}^2 \rightarrow \mathbb{R}^M$, where $M \leq \deg(f_1) + \deg(f_2) + 1$, with $\deg(\cdot)$ denoting the degree of a polynomial, and a set of single-variable polynomials $\{g_i: \mathbb{R} \rightarrow \mathbb{R}\}_{i = 1}^M$ such that
\begin{equation}
\label{eq:product}
f_1(\chi_1) f_2(\chi_2) = \sum_{i = 1}^M g_i(T_{i1} \chi_1 + T_{i2} \chi_2).
\end{equation}
\end{lemma}

\begin{proof}
We begin by expanding the single-variable polynomials in their respective power bases:
\[
f_1(\chi_1) = \sum_{p = 0}^{\deg(f_1)} a_p \chi_1^p, \qquad
f_2(\chi_2) = \sum_{q = 0}^{\deg(f_2)} b_q \chi_2^q.
\]
Multiplying these expressions yields a bivariate polynomial represented as a linear combination of monomials:
\[
f_1(\chi_1) f_2(\chi_2) = \sum_{p = 0}^{\deg(f_1)} \sum_{q = 0}^{\deg(f_2)} a_p b_q \chi_1^p \chi_2^q.
\]
To apply Lemma \ref{lem:monomial} using a single unified linear transformation $\boldsymbol{T}$ for all terms, we fix $M = \deg(f_1) + \deg(f_2) + 1$ distinct constants $\{c_i\}_{i = 1}^M$.
For any given monomial $\chi_1^p \chi_2^q$ in the expansion, the total degree satisfies $p + q \leq \deg(f_1) + \deg(f_2) = M - 1$.

The matching of coefficients for a specific monomial generates a Vandermonde system of $p + q + 1$ equations with $M$ unknowns $\{\zeta_{pqi}\}_{i = 1}^M$.
Since $p + q + 1 \leq M$ and the constants $\{c_i\}$ are distinct, the resulting $(p + q + 1) \times M$ Vandermonde matrix has full row rank.
This guarantees that the system is consistent and admits at least one valid solution set $\{\zeta_{pqi}\}_{i = 1}^M$ for each pair $(p, q)$.
While the solution is nonunique for $p + q + 1 < M$, we may fix any particular solution, such as the minimum-norm solution obtained via the Moore--Penrose pseudoinverse \citep{barata2012moore} of the Vandermonde matrix.

By solving for $\{\zeta_{pqi}\}$ using the same $M$ constants for every monomial, we can employ a global transformation $\boldsymbol{T}$ with $T_{i1} = 1$ and $T_{i2} = c_i$.
Substituting the monomial representations back into the expansion and reordering the summations, we obtain
\[
f_1(\chi_1) f_2(\chi_2) = \sum_{i = 1}^M \left(\sum_{p = 0}^{\deg(f_1)} \sum_{q = 0}^{\deg(f_2)} a_p b_q \zeta_{pqi} (\chi_1 + c_i \chi_2)^{p + q}\right).
\]
This result matches the form of Eq.~\eqref{eq:product} by defining each single-variable polynomial $g_i$ as
\[
g_i(z_i) = \sum_{p = 0}^{\deg(f_1)} \sum_{q = 0}^{\deg(f_2)} a_p b_q \zeta_{pqi} z_i^{p + q},
\]
where $z_i = T_{i1} \chi_1 + T_{i2} \chi_2$ represents a projected input.
This completes the proof.
\end{proof}

Another consequence is the reproducing property for arbitrary polynomials in two dimensions, which follows from Lemma \ref{lem:product} by observing that each monomial $\chi_1^p \chi_2^q$ is itself a product of two single-variable polynomials.
\begin{lemma}[Bivariate polynomial representation]
\label{lem:bivariate}
For any bivariate polynomial $\mathcal{F}$ of $(\chi_1, \chi_2) \in \mathbb{R}^2$, there exist a linear transformation $\boldsymbol{T}: \mathbb{R}^2 \rightarrow \mathbb{R}^M$, where $M \leq \deg(\mathcal{F}) + 1$, and a set of single-variable polynomials $\{g_i: \mathbb{R} \rightarrow \mathbb{R}\}_{i = 1}^M$ such that
\begin{equation}
\label{eq:bivariate}
\mathcal{F}(\chi_1, \chi_2) = \sum_{i = 1}^M g_i(T_{i1} \chi_1 + T_{i2} \chi_2).
\end{equation}
\end{lemma}

\begin{proof}
Note that $\deg(\mathcal{F})$ refers to the degree of the multi-variable polynomial, defined as the maximum value of $p + q$ among the monomials $\{\chi_1^p \chi_2^q\}$ appearing in $\mathcal{F}$ with nonzero coefficients.
Expanding $\mathcal{F}$ in the monomial basis, we have
\[
\mathcal{F}(\chi_1, \chi_2) = \sum_{p + q \leq \deg(\mathcal{F})} a_{pq} \chi_1^p \chi_2^q.
\]
The key observation is that each monomial $\chi_1^p \chi_2^q$ is a specific instance of the product form $f_1(\chi_1) f_2(\chi_2)$ treated in Lemma \ref{lem:product}, where $\deg(f_1) + \deg(f_2) = p + q \leq \deg(\mathcal{F})$.
Lemma \ref{lem:product} therefore applies to each monomial individually, utilizing a unified set of $M = \deg(\mathcal{F}) + 1$ projections shared globally across all monomials.
The remainder of the proof follows identically to that of Lemma \ref{lem:product}.
\end{proof}

Finally, we employ mathematical induction to extend the result to any input dimension.
\begin{theorem}[Polynomial representation via induction]
\label{thm:induction}
For any multi-variable polynomial $\mathcal{F}$ of $(\chi_1, \dots, \chi_N) \in \mathbb{R}^N$, there exist a linear transformation $\boldsymbol{T}: \mathbb{R}^N \rightarrow \mathbb{R}^M$, for some $M$ that can be chosen to depend only on $\deg(\mathcal{F})$ and $N$ (e.g., $M = (\deg(\mathcal{F}) + 1)^{N - 1}$), and a set of single-variable polynomials $\{g_i: \mathbb{R} \rightarrow \mathbb{R}\}_{i = 1}^M$ such that
\begin{equation}
\label{eq:induction}
\mathcal{F}(\chi_1, \dots, \chi_N) = \sum_{i = 1}^M g_i\left(\sum_{j = 1}^N T_{ij} \chi_j\right).
\end{equation}
\end{theorem}

\begin{proof}
We proceed by mathematical induction on the input dimension $N$.
For the base case $N = 2$, Lemma \ref{lem:bivariate} establishes that a single transformation $\boldsymbol{T}: \mathbb{R}^2 \rightarrow \mathbb{R}^M$ suffices for any bivariate polynomial of degree $\deg(\mathcal{F})$, as the choice of projection constants $\{c_i\}$ depends only on $\deg(\mathcal{F})$.

Now, assume that Theorem \ref{thm:induction} holds uniformly for the space of polynomials of degree $\deg(\mathcal{F})$ in $N$ variables, that is, there exists a global transformation $\boldsymbol{T}_N: \mathbb{R}^N \rightarrow \mathbb{R}^{M_N}$ that can represent any such polynomial by varying only the single-variable functions.
We then consider a polynomial $\mathcal{F}(\chi_1, \dots, \chi_N, \chi_{N + 1})$ of degree $\deg(\mathcal{F})$ defined on $\mathbb{R}^{N + 1}$.
It can be expanded as
\[
\mathcal{F}(\chi_1, \dots, \chi_N, \chi_{N + 1}) = \sum_{r = 0}^R \mathcal{F}_r(\chi_1, \dots, \chi_N) \chi_{N + 1}^r,
\]
for some polynomials $\{\mathcal{F}_r\}_{r = 0}^R$, where $R \leq \deg(\mathcal{F})$.
Since $\mathcal{F}$ has degree $\deg(\mathcal{F})$, each term $\mathcal{F}_r(\chi_1, \dots, \chi_N) \chi_{N + 1}^r$ must satisfy $\deg(\mathcal{F}_r) \leq \deg(\mathcal{F}) - r$.
In particular, because $\deg(\mathcal{F}_r) \leq \deg(\mathcal{F})$ for all $r$, the inductive hypothesis allows us to use the same global transformation $\boldsymbol{T}_N$ for all $r \in \{0, \dots, R\}$.
Thus, there exists a set of single-variable polynomials $\{\{\widetilde{g}_{ri}\}_{i = 1}^{M_N}\}_{r = 0}^R$ such that
\begin{equation}
\label{eq:hypo}
\mathcal{F}_r(\chi_1, \dots, \chi_N) = \sum_{i = 1}^{M_N} \widetilde{g}_{ri}(z_{N, i}), \qquad
z_{N, i} = \sum_{j = 1}^N T_{N, ij} \chi_j.
\end{equation}
Substituting Eq.~\eqref{eq:hypo} into the expansion and rearranging the summations, we obtain
\[
\mathcal{F}(\chi_1, \dots, \chi_N, \chi_{N + 1}) = \sum_{i = 1}^{M_N} \left( \sum_{r = 0}^R \widetilde{g}_{ri}(z_{N, i}) \chi_{N + 1}^r \right).
\]
Let $\mathcal{H}_i(z_{N, i}, \chi_{N + 1}) = \sum_{r = 0}^R \widetilde{g}_{ri}(z_{N, i}) \chi_{N + 1}^r$.
Each $\mathcal{H}_i$ is a bivariate polynomial in the variables $z_{N, i}$ and $\chi_{N + 1}$ of total degree at most $\deg(\mathcal{F})$.
By Lemma \ref{lem:bivariate}, each $\mathcal{H}_i$ admits a representation involving at most $\deg(\mathcal{F}) + 1$ projections of the form $z_{N + 1, ik} = z_{N, i} + c_k \chi_{N + 1}$.

Given that the choice of $\{c_k\}$ is independent of the specific polynomials $\{\mathcal{H}_i\}$, we can share these $\deg(\mathcal{F}) + 1$ constants across all $i \in \{1, \dots, M_N\}$, resulting in a global transformation $\boldsymbol{T}_{N + 1}: \mathbb{R}^{N + 1} \rightarrow \mathbb{R}^{M_{N + 1}}$ for $N + 1$ dimensions.
Therefore, the total number of projections is
\[
M_{N + 1} \leq M_N \cdot (\deg(\mathcal{F}) + 1).
\]
Solving this recurrence relation with the base case $M_2 = \deg(\mathcal{F}) + 1$ yields
\[
M_N \leq (\deg(\mathcal{F}) + 1)^{N - 1}.
\]
As a result, Theorem \ref{thm:induction} holds for the input dimension $N + 1$.
This completes the induction process and the proof.
\end{proof}

Theorem \ref{thm:induction} shows that to represent a polynomial $\mathcal{F}$ of degree $P = \deg(\mathcal{F})$ in $N$ variables, the number of required projections satisfies $M \leq (P + 1)^{N - 1}$; thus, for large $P$, the bound scales as $M \asymp P^{N - 1}$.
This bound is consistent with classical ridge function approximation theory (see, e.g., \citep{pinkus1999approximation}), which establishes the same $P^{N - 1}$ scaling.
While Theorem \ref{thm:induction} achieves this scaling constructively via mathematical induction using a Vandermonde system of equations, \citet{pinkus1999approximation} leverages a non-constructive dimension argument on the space of homogeneous polynomials to guarantee the existence of suitable directions without specifying them.
Alternatively, an estimate on $M$ can be obtained via a complementary construction based on the polarization identity \citep{landsberg2012tensors}.
\begin{theorem}[Polynomial representation via polarization]
\label{thm:polar}
Let $\mathcal{P}_{N, P}$ be the space of multi-variable polynomials in $N$ variables of degree at most $P$.
Any $\mathcal{F} \in \mathcal{P}_{N, P}$ can be represented exactly by a PNAM with $M$ projections, where $M$ is bounded by the total number of terms generated by the polarization of each basis monomial, that is,
\[
M \leq \sum_{d = 1}^P \binom{N + d - 1}{d} 2^{d - 1}.
\]
\end{theorem}

\begin{proof}
Any polynomial $\mathcal{F}(\boldsymbol{\chi})$ of degree $P$, where $\boldsymbol{\chi} = (\chi_1, \dots, \chi_N) \in \mathbb{R}^N$, can be written as a linear combination of monomials using multi-index notation:
\[
\mathcal{F}(\boldsymbol{\chi}) = \sum_{|\alpha| \leq P} a_\alpha \boldsymbol{\chi}^\alpha, \qquad
\boldsymbol{\chi}^\alpha = \chi_1^{\alpha_1} \cdots \chi_N^{\alpha_N}.
\]
The number of monomials of degree exactly $d$ is $\binom{N + d - 1}{d}$, and the total dimension of the polynomial space is $\dim(\mathcal{P}_{N, P}) = \binom{N + P}{P}$.
Note that constant monomials ($|\alpha| = 0$) require no projections, as their contribution can be absorbed into the constant term of any single-variable polynomial in the representation; hence, they do not contribute to $M$.

\paragraph{Step 1: Polarization of a single monomial.}
For each monomial $\boldsymbol{\chi}^\alpha$ of degree $d = |\alpha| \geq 1$, we apply the polarization identity.
We encode the factored structure of $\boldsymbol{\chi}^\alpha = \chi_1^{\alpha_1} \cdots \chi_N^{\alpha_N}$ as a product of $d$ linear factors by introducing $d$ vectors $\boldsymbol{v}_1, \dots, \boldsymbol{v}_d \in \mathbb{R}^N$, where $\alpha_j$ of them equal the $j^\mathrm{th}$ standard basis vector $\boldsymbol{e}_j$, so that $\boldsymbol{v}_i \cdot \boldsymbol{\chi} = \chi_j$ for the corresponding factor.
The polarization identity then gives
\[
\boldsymbol{\chi}^{\alpha} = \frac{1}{d! 2^d} \sum_{\boldsymbol{\delta} \in \{-1, 1\}^d} \left(\prod_{i = 1}^d \delta_i\right) \left(\sum_{i = 1}^d \delta_i \boldsymbol{v}_i \cdot \boldsymbol{\chi}\right)^d.
\]
Pairing each sign vector $\boldsymbol{\delta}$ with its negation $-\boldsymbol{\delta}$ yields equal contributions, since
\[
\begin{aligned}
\prod_{i = 1}^d (-\delta_i) \left(-\sum_{i = 1}^d \delta_i \boldsymbol{v}_i \cdot \boldsymbol{\chi}\right)^d &= (-1)^d (-1)^d \prod_{i = 1}^d \delta_i \left(\sum_{i = 1}^d \delta_i \boldsymbol{v}_i \cdot \boldsymbol{\chi}\right)^d \\
&= \prod_{i = 1}^d \delta_i \left(\sum_{i = 1}^d \delta_i \boldsymbol{v}_i \cdot \boldsymbol{\chi}\right)^d.
\end{aligned}
\]
As a result, the $2^d$ terms collapse to $2^{d - 1}$ distinct projected power functions of the form $\zeta_k (\boldsymbol{t}_k \cdot \boldsymbol{\chi})^d$, where $\zeta_k = \frac{\prod_{i = 1}^d \delta_{ki}}{d! 2^{d - 1}}$ and $\boldsymbol{t}_k = \sum_{i = 1}^d \delta_{ki} \boldsymbol{v}_i$ are the scalar weight and projection direction associated with the $k^\mathrm{th}$ sign vector pair $(\boldsymbol{\delta}_k, -\boldsymbol{\delta}_k)$, respectively.

\paragraph{Step 2: Summation over the polynomial.}
Substituting the polarized form of each monomial back into the polynomial expansion yields
\[
\mathcal{F}(\boldsymbol{\chi}) = \sum_{\substack{|\alpha| \leq P \\ |\alpha| \geq 1}} a_\alpha \left(\sum_{k = 1}^{2^{|\alpha| - 1}} \zeta_{\alpha k} (\boldsymbol{t}_{\alpha k} \cdot \boldsymbol{\chi})^{|\alpha|}\right).
\]
By linearity, rearranging the summations gives
\begin{equation}
\label{eq:polar}
\mathcal{F}(\boldsymbol{\chi}) = \sum_{\alpha \neq \boldsymbol{0}} \sum_k g_{\alpha k}(\boldsymbol{t}_{\alpha k} \cdot \boldsymbol{\chi}),
\end{equation}
where $g_{\alpha k}(z) = a_\alpha \zeta_{\alpha k} z^{|\alpha|}$ is a single-variable polynomial (specifically a monomial).

\paragraph{Step 3: Consolidation.}
Based on the above construction, the total number of projections $M$ is bounded by the sum over all monomial degrees of the number of projections per monomial:
\[
M \leq \sum_{d = 1}^P \binom{N + d - 1}{d} 2^{d - 1}.
\]
This completes the proof.
\end{proof}

Theorems \ref{thm:induction} and \ref{thm:polar} each provide a constructive upper bound for $M$, but with complementary limitations: Theorem \ref{thm:induction} scales as $P^{N - 1}$ and is exponential in $N$, while Theorem \ref{thm:polar} scales as $(2^{P - 1} N^P) / P!$ and is exponential in $P$.
A bound that is simultaneously polynomial in both $N$ and $P$ can be obtained by a non-constructive argument via the Waring rank \citep{landsberg2012tensors}.
\begin{theorem}[Bound via Waring rank]
\label{thm:waring}
The minimum number of projections $M$ required to represent any multi-variable polynomial $\mathcal{F}$ in $N$ variables of degree $P$ is bounded by the dimension of the polynomial space:
\[
M \leq \binom{N + P}{P}.
\]
\end{theorem}

\begin{proof}
The set of all multi-variable polynomials of degree at most $P$ in $N$ variables, $\mathcal{P}_{N, P}$, is a vector space with dimension $\dim(\mathcal{P}_{N, P}) = \binom{N + P}{P}$.
The set of projected power functions $\{(\boldsymbol{t} \cdot \boldsymbol{\chi})^d \mid \boldsymbol{t} \in \mathbb{R}^N, 1 \leq d \leq P\}$ spans $\mathcal{P}_{N, P}$, as every monomial can be recovered via the polarization identity in Theorem \ref{thm:polar}.
Since this spanning set covers the entire space, one can extract a basis of size $\dim(\mathcal{P}_{N, P}) = \binom{N + P}{P}$.
Any $\mathcal{F} \in \mathcal{P}_{N, P}$ can therefore be expressed as a linear combination of at most $\binom{N + P}{P}$ projected functions.
\end{proof}

As noted, the Waring bound $\binom{N + P}{P} \sim P^N / N!$ scales polynomially in both $N$ and $P$, making it tighter than Theorem \ref{thm:polar} for large $P$ regardless of $N$.
This bound is looser than Theorem \ref{thm:induction} when $P$ is large relative to $N$, but tighter than Theorem \ref{thm:induction} when $N$ is large relative to $P$.
For large $P$, the Waring bound can be further refined for generic polynomials via the notion of generic Waring rank, as formalized by the Alexander--Hirschowitz theorem \citep{alexander1995polynomial, landsberg2012tensors}.
The following corollary summarizes the resulting bounds on $M$, including a parameter-counting lower bound.
\begin{corollary}[Projection complexity]
\label{cor:proj}
For the space $\mathcal{P}_{N, P}$ of multi-variable polynomials in $N$ variables of degree at most $P$, the minimum number of projections $M$ in the exact polynomial representation satisfies
\[
\frac{1}{N + P + 1}\binom{N + P}{P} \leq M \leq \binom{N + P}{P},
\]
where the lower bound follows from a parameter-counting argument and the upper bound from Theorem \ref{thm:waring}.
In the exact polynomial representation setting, each of the $M$ projections contributes $N$ free parameters from the linear transformation $\boldsymbol{T}$ and $P + 1$ coefficients from the single-variable polynomial $g_i$, giving $M (N + P + 1)$ total free parameters, which must be at least $\dim(\mathcal{P}_{N, P}) = \binom{N + P}{P}$.
For generic polynomials with large $P$, the Alexander--Hirschowitz theorem \citep{alexander1995polynomial, landsberg2012tensors} tightens the upper bound to
\[
M_\mathrm{generic} = \left\lceil\frac{1}{N}\binom{N + P}{P}\right\rceil,
\]
which differs from the lower bound by a factor of $\frac{N + P + 1}{N}$.
\end{corollary}

The bounds established in this section---from the constructive bound of Theorem \ref{thm:induction} to the parameter-counting lower bound of Corollary \ref{cor:proj}---characterize the minimum $M$ required for exact polynomial representation.
In practice, however, PNAMs do not aim to represent a fixed polynomial exactly, but rather to learn an effective approximation of an unknown function from data.
As demonstrated in Example \ref{ex:two_dim}, the nonuniqueness of the transformation $\boldsymbol{T}$ means that the theoretical upper bounds on $M$ are not tight in general; different constructions can achieve the same representation with fewer projections.
This suggests that, through learning and optimization, PNAMs can discover representations that are more compact than the theoretical bounds would predict, adapting $\boldsymbol{T}$ to the specific structure of the target function rather than to the worst case.

\subsection{Additional results for knot theory: classification} \label{subsec:knot_classif}
\begin{figure}[!ht]
\centering
{(a)}{\includegraphics[width=0.45\textwidth]{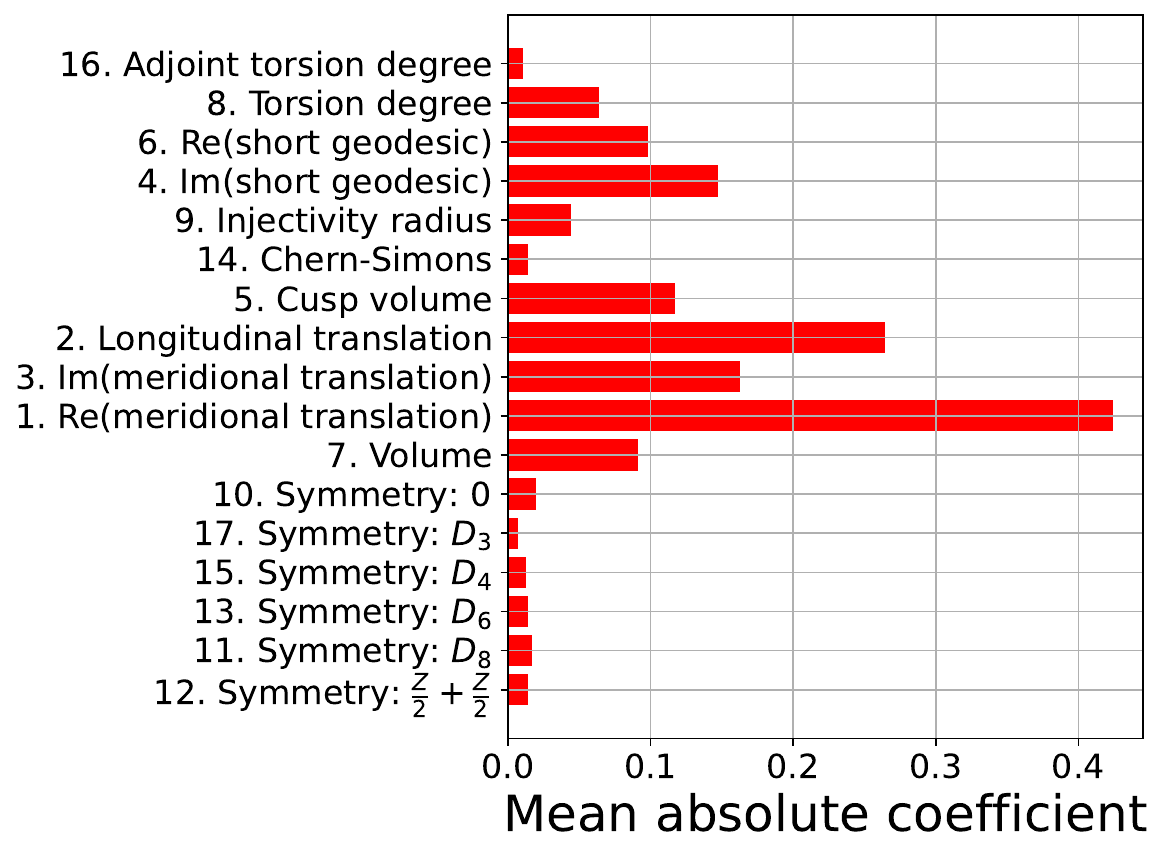}}
{(b)}{\includegraphics[width=0.45\textwidth]{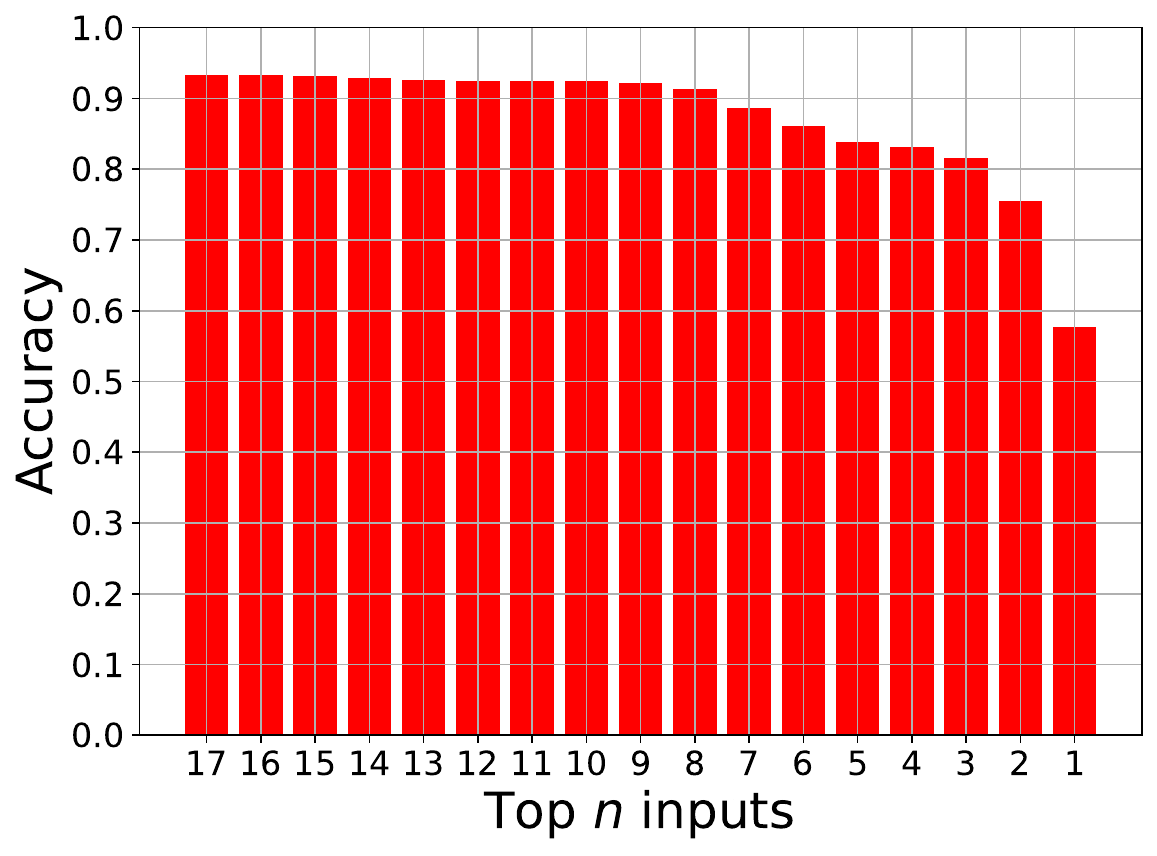}}
\caption{One complete instance of the ranking and accuracy reported in Table \ref{tab:knot_theory_classif_rank}. (a) Relative importance of the input features based on the mean absolute coefficients, and (b) test accuracy associated with using the top $n$ features in (a).}
\label{fig:knot_theory_classif}
\end{figure}

Figure \ref{fig:knot_theory_classif} shows a complete ranking of the inputs and the associated test accuracy of using the top $n$ inputs from one of the runs summarized in Table \ref{tab:knot_theory_classif_rank}.

\paragraph{Validation of the importance ranking.}
To test whether the ranking by the mean absolute coefficients of the linear transformation $\boldsymbol{T}$ captures predictive importance as opposed to mere feature usage, we perform four post-training checks on the PNAM with $M = 8$ and without the Fourier feature mapping in Table \ref{tab:knot_theory_classif_compare}.
First, columns of $\boldsymbol{T}$ are zeroed out according to the ranking of each run.
Table \ref{tab:knot_theory_classif_removal} reveals that removing the single top-ranked feature reduces the test accuracy from 93.6 to 30.9\%, whereas removing the three bottom-ranked features has no measurable effect.
Second, starting from the transformation $\boldsymbol{T}$ with all columns zeroed out, a greedy selection reactivates one column at a time, taking the column with the largest gain in accuracy as the next most significant input.
This procedure measures predictive importance without reference to the coefficients of $\boldsymbol{T}$, and it recovers the same top two invariants as the mean absolute coefficients in 10/10 runs and the same third invariant in 5/10 runs, where the third position varies between the meridional translation (imaginary part) and the cusp volume (which also appears at rank 3 in Table \ref{tab:knot_theory_classif_rank}).
However, the greedy selection requires $\mathcal{O}(N^2)$ evaluations of the test set, so the ranking by the mean absolute coefficients remains the practical choice when the number of inputs $N$ is large (e.g., $N = 784$ for the MNIST dataset).

\begin{table}[!ht]
\caption{Test accuracy of the PNAM with $M = 8$ (without the Fourier feature mapping) in Table \ref{tab:knot_theory_classif_compare} after zeroing out the columns of the linear transformation $\boldsymbol{T}$ associated with the top or bottom $n$ features of each run. The mean and standard deviation are computed over 10 runs without retraining.}
\label{tab:knot_theory_classif_removal}
\centering
\begin{tabular}{lll}
\toprule
\multirow{2}{*}{Remove $n$} & \multicolumn{2}{c}{Test acc.} \\
\cmidrule(lr){2-3}
& Top & Bottom \\
\midrule
0 (baseline) & $93.6 \pm 0.4$\% & $93.6 \pm 0.4$\% \\
1 & $30.9 \pm 2.1$\% & $93.7 \pm 0.4$\% \\
2 & $28.8 \pm 1.2$\% & $93.6 \pm 0.4$\% \\
3 & $25.2 \pm 2.8$\% & $93.5 \pm 0.4$\% \\
\bottomrule
\end{tabular}
\end{table}

Third, permutation importance---the drop in accuracy when the values of one input are shuffled across the test set---correlates with the mean absolute coefficients, yielding a Spearman rank correlation of $\rho = 0.939 \pm 0.023$; the two rankings agree on the most important feature in 10/10 runs.
Fourth, the gradient sensitivity $\mathbb{E} |\partial \widehat{y}_i / \partial \chi_k|$ over the test set, which incorporates the derivatives of the bases $\{g_{ij}\}$ and cancellations across projections, gives $\rho = 0.977 \pm 0.014$, agreeing on the top-ranked feature in 9/10 runs, where the single disagreement selects the second-ranked feature.
These checks complement the accuracies of retaining the top $n$ features in Table \ref{tab:knot_theory_classif_rank} and Fig.~\ref{fig:knot_theory_classif}(b).

\paragraph{Group-lasso MLP baseline.}
As an alternative to the ranking by the mean absolute coefficients, we retrain the MLP in Table \ref{tab:knot_theory_classif_compare} under an otherwise identical protocol with the added penalty $(w^\prime / N) \sum_{k} \|\boldsymbol{W}^{(1)}_k\|_2$, where $\boldsymbol{W}^{(1)}_k$ is the $k^\mathrm{th}$ column of the first-layer weight matrix $\boldsymbol{W}^{(1)}$, rank the input features by $\|\boldsymbol{W}^{(1)}_k\|_2$, and sweep $w^\prime$ over $\{0, 0.001, 0.01, 0.1, 1\}$.
At every $w^\prime$, this baseline is a weaker feature selector than the PNAM.
At the representative value $w^\prime = 0.01$, it matches the unregularized MLP in accuracy ($95.7 \pm 0.1$\%) and its most important feature agrees with that of the PNAM in 10/10 runs, but the features ranked second and third vary across runs and never include the longitudinal translation or the meridional translation (imaginary part).
Retaining its top three features produces a test accuracy of $50.2 \pm 3.8$\%, against $78.1 \pm 2.4$\% for the PNAM under the same input pruning (Table \ref{tab:knot_theory_classif_rank}).
This behavior persists for all $w^\prime \leq 0.1$.
At $w^\prime = 1$, the selection improves, recovering the longitudinal translation among its top three features in 9/10 runs with a top-three accuracy of $75.5 \pm 6.3$\%, but the accuracy with the full feature set falls to $86.8 \pm 1.2$\%, below even that of the NAM.
In the MLP, the deeper layers freely recombine the first-layer outputs, so the column norms become informative only once a sufficiently strong penalty concentrates the first-layer weights on a few inputs, by which point the accuracy has already degraded.

\paragraph{Ablation of the projection matrix.}
Table \ref{tab:knot_theory_classif_ablation} isolates the roles of learning and regularizing the linear transformation $\boldsymbol{T}$ in one matched setting: knot classification with $M = 17$, where all models use 17 basis MLPs of identical architecture under the protocol of Table \ref{tab:knot_theory_classif_compare}.
First, random mixing is worse than no mixing at all, so the performance gain of PNAMs cannot be attributed to the projection mechanism alone or to the additional parameters of $\boldsymbol{T}$.
The gain comes from learning the projection itself: the learned transformation with regularization weights $w_1 = \dots = w_5 = w = 0$ outperforms the random transformation fixed at initialization by 6.5\% and the identity transformation by 2.9\%.
Second, regularizing $\boldsymbol{T}$ costs 0.3\% in accuracy but, in exchange, stabilizes the feature ranking.
In the regularized models, the three most significant invariants of Table \ref{tab:knot_theory_classif_rank} occupy ranks 1, 2, and 3 in 10/10, 10/10, and 8/10 runs, respectively.
In the unregularized models, the top-ranked feature is stable, but the second-ranked feature varies across runs.
The regularization weight $w$ thus allows the user to navigate the trade-off between accuracy and ranking reliability discussed in Section \ref{sec:limit}.

\begin{table}[!ht]
\caption{Ablation of the projection matrix $\boldsymbol{T}$ on knot classification with $M = 17$. All models use 17 basis MLPs of identical architecture under the protocol of Table \ref{tab:knot_theory_classif_compare}; the mean and standard deviation of the test accuracy are computed over 10 runs.}
\label{tab:knot_theory_classif_ablation}
\centering
\begin{tabular}{ll}
\toprule
$\boldsymbol{T}$ & Test acc. \\
\midrule
Random, fixed at initialization & $88.8 \pm 0.5$\% \\
Identity (= NAM, Table \ref{tab:knot_theory_classif_compare}) & $92.4 \pm 0.2$\% \\
Learned, regularized (Table \ref{tab:knot_theory_classif_compare}) & $95.0 \pm 0.2$\% \\
Learned, unregularized ($w = 0$) & $95.3 \pm 0.2$\% \\
\bottomrule
\end{tabular}
\end{table}

\paragraph{Alternative to the penalty $\ell_3$.}
To assess the sensitivity of our results to the form of the penalty $\ell_3$, we retrain the PNAM with $M = 8$ (without the Fourier feature mapping) in Table \ref{tab:knot_theory_classif_compare} under an otherwise identical protocol, with $\ell_3$ replaced by $\ell_3^\prime$ in footnote \ref{fn:l3_prime}.
This substitution yields the same test accuracy ($93.6 \pm 0.4$\%) and the same top two invariants in 10/10 runs, with the invariant ranked third slightly more stable under $\ell_3$ (7/10 runs compared to 4/10).
Our conclusions are thus insensitive to this design choice.

\subsection{Additional results for knot theory: regression} \label{subsec:knot_regress}
In Table \ref{tab:knot_theory_regress_compare}, we compare the performance of the MLP and the NAM with 17 bases against that of the PNAM with eight bases under three values of the regularization weights $w_1 = \dots = w_5 = w$.
Both the NAM and the PNAM can be converted into symbolic expressions after training.
Without regularization ($w = 0$), the PNAM is within 3.2\% of the MLP.
The larger gap at $w = 0.001$ is the price of a sparser model with a more reliable ranking, and this model is chosen for symbolic distillation below.
Given the poor performance of the NAM, we proceed with just the PNAM.
Similar to Table \ref{tab:knot_theory_classif_rank}, Table \ref{tab:knot_theory_regress_rank} suggests that the meridional translation (real part) and the longitudinal translation are the most and second most important invariants, respectively.
Nevertheless, solely comparing the mean absolute coefficients may not be sufficient, as unimportant features could be incorrectly ranked as important.
We can, however, eliminate these features by examining the accuracy associated with including them, since including an unimportant feature does not improve accuracy compared to the corresponding case with one fewer input.

\begin{table}[!ht]
\caption{Performance comparison of different neural network architectures for the single-output regression problem of predicting the signature of mathematical knots. The mean and standard deviation of the test accuracy are computed over 10 runs. The three PNAMs use regularization weights $w_1 = \dots = w_5 = w$; Figs.~\ref{fig:knot_theory_regress_loss} and \ref{fig:knot_theory_regress_w} show their corresponding training and validation histories and the effect of $w$ on the input ranking, respectively.}
\label{tab:knot_theory_regress_compare}
\centering
\begin{tabular}{llll}
\toprule
Method & Architecture & Parameter count & Test acc. \\
\midrule
MLP & 3 layers: [17, 64, 32, 1] & $1.2 \times 10^4$ & $94.7 \pm 0.2$\% \\
NAM & 3 layers: $17 \times [1, 64, 32, 1]$ & $2.1 \times 10^5$ & $65.0 \pm 0.2$\% \\
PNAM ($w = 0$) & 3 layers: $8 \times [1, 64, 32, 1]$ & $9.8 \times 10^4$ & $91.5 \pm 0.4$\% \\
PNAM ($w = 0.001$) & 3 layers: $8 \times [1, 64, 32, 1]$ & $9.8 \times 10^4$ & $85.5 \pm 0.3$\% \\
PNAM ($w = 0.01$) & 3 layers: $8 \times [1, 64, 32, 1]$ & $9.8 \times 10^4$ & $69.9 \pm 6.4$\% \\
\bottomrule
\end{tabular}
\end{table}

\begin{table}[!ht]
\caption{Ranking of important features based on the mean absolute coefficients for the PNAM with $w = 0.001$ in Table \ref{tab:knot_theory_regress_compare}. The mean and standard deviation of the test accuracy (associated with retaining only the top $n$ features) are computed over the runs in which the top $n$ features appeared.}
\label{tab:knot_theory_regress_rank}
\centering
\begin{tabular}{lllll}
\toprule
Rank & Input feature & Symbol & Frequency & Test acc. \\
\midrule
1 & Re(meridional translation) & $\chi_{10}$ & 10/10 & $55.7 \pm 1.2$\% \\
\hline
\multirow{2}{*}{2} & Longitudinal translation & $\chi_8$ & 6/10 & $74.8 \pm 1.4$\% \\
& Torsion degree & $\chi_2$ & 4/10 & $54.9 \pm 0.5$\% \\
\hline
\multirow{5}{*}{3} & Torsion degree & $\chi_2$ & 3/10 & $76.4 \pm 0.4$\% \\
& Re(short geodesic) & $\chi_3$ & 3/10 & $61.8 \pm 7.3$\% \\
& Longitudinal translation & $\chi_8$ & 2/10 & $76.5 \pm 0.2$\% \\
& Cusp volume & $\chi_7$ & 1/10 & 81.3\% \\
& Im(short geodesic) & $\chi_4$ & 1/10 & 74.3\% \\
\bottomrule
\end{tabular}
\end{table}

\begin{figure}[!ht]
\centering
{(a)}{\includegraphics[width=0.3\textwidth]{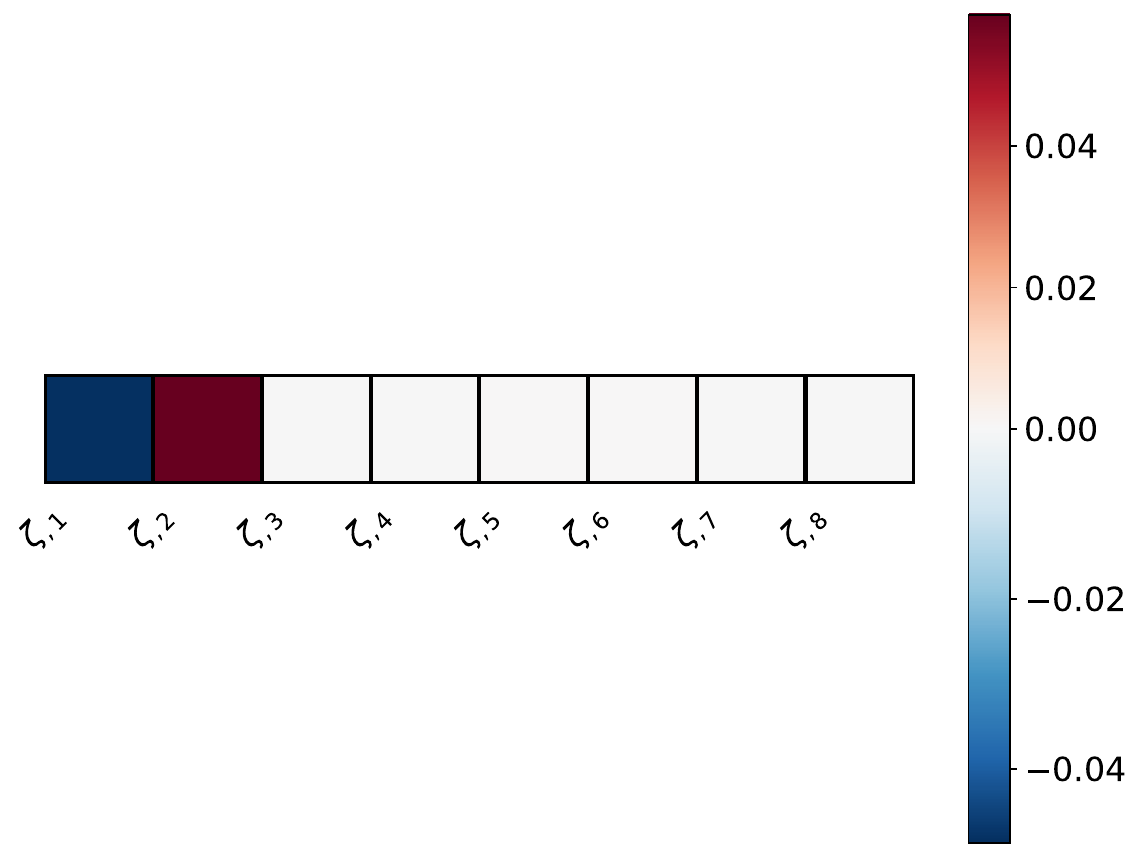}}
{(b)}{\includegraphics[width=0.3\textwidth]{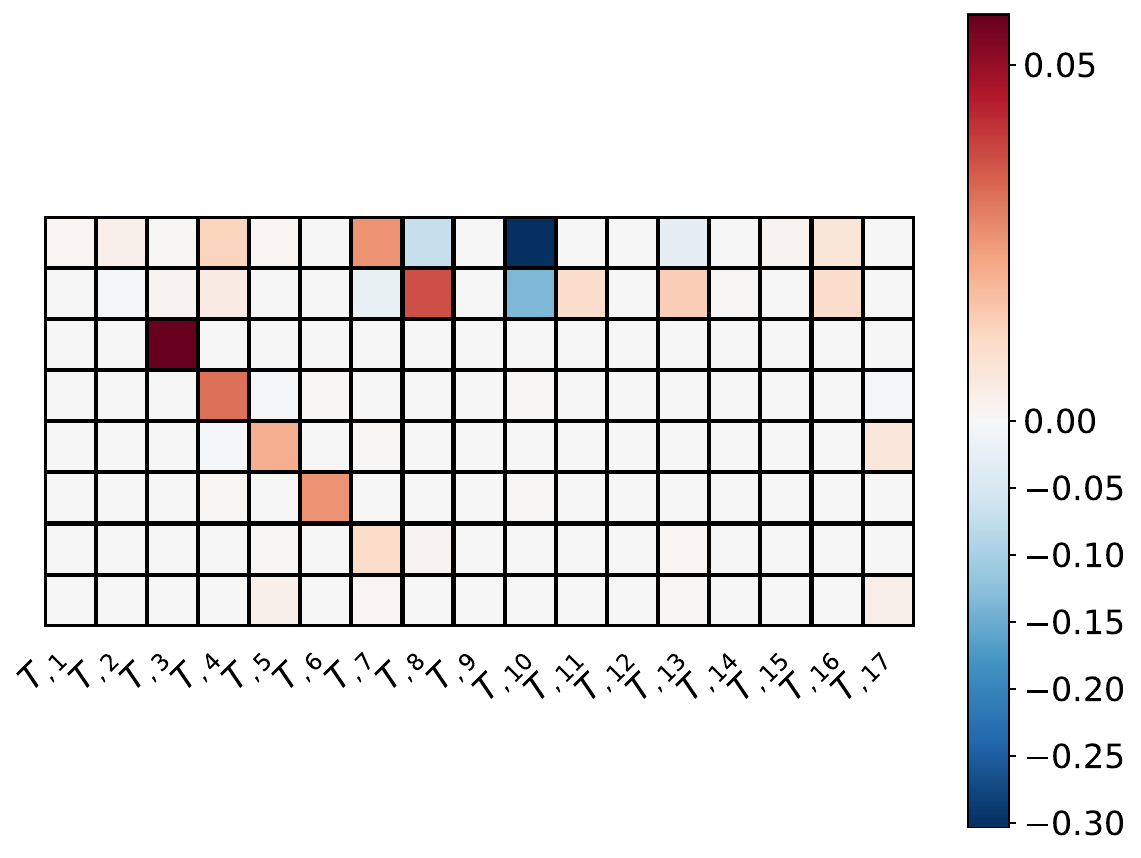}}
{(c)}{\includegraphics[width=0.3\textwidth]{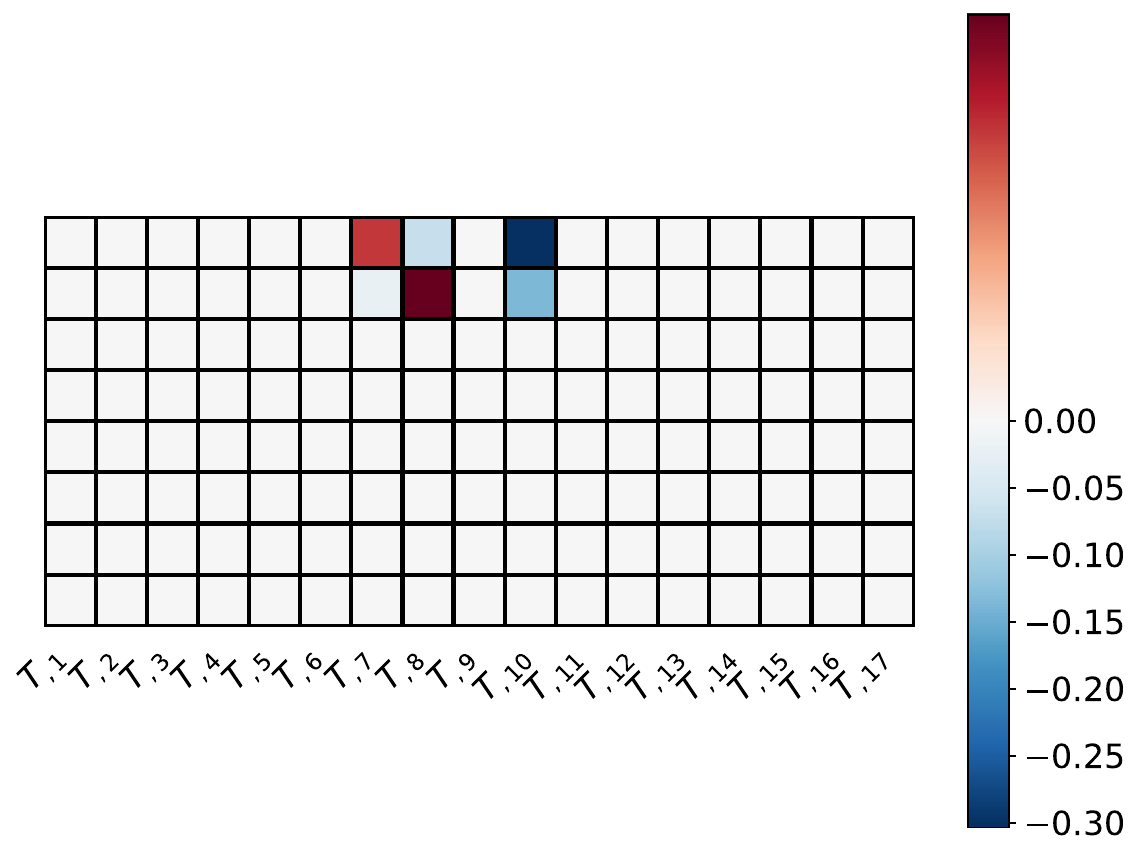}}
{(d)}{\includegraphics[width=0.45\textwidth]{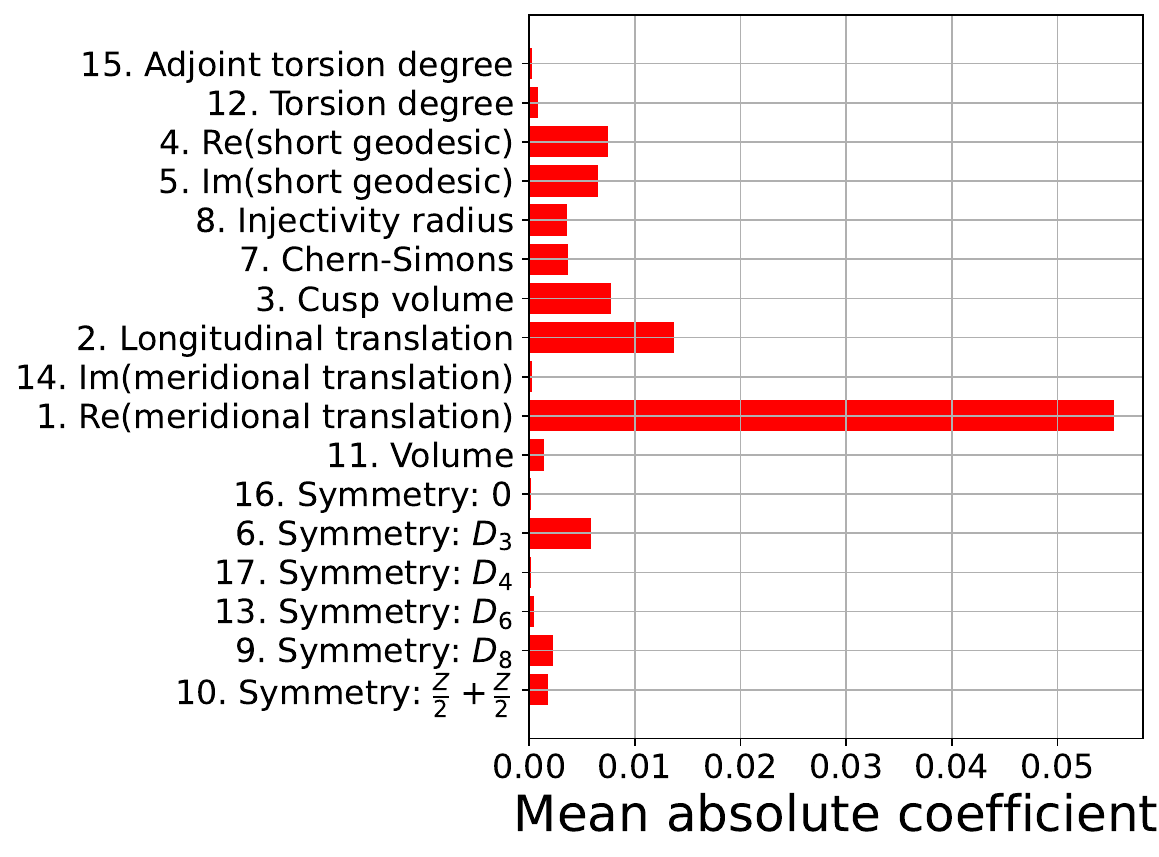}}
{(e)}{\includegraphics[width=0.45\textwidth]{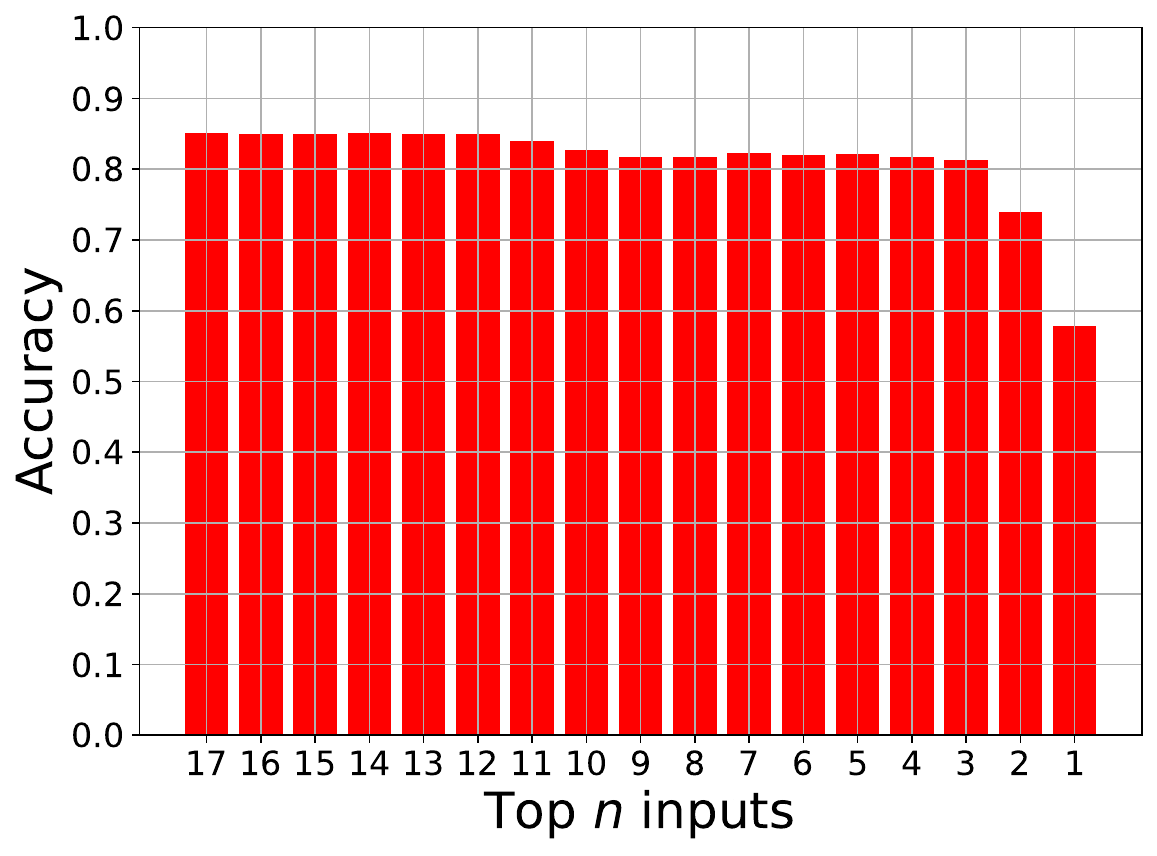}}
\caption{One complete instance of the ranking and accuracy reported in Table \ref{tab:knot_theory_regress_rank}, along with optimized parameters of the PNAM. (a) Weighting coefficients $\boldsymbol{\zeta}$ prior to post-processing, (b) linear transformation $\boldsymbol{T}$ prior to post-processing, (c) post-processed $\boldsymbol{T}$, retaining only the top $n = 3$ features and zeroing out the rows that correspond to the zero-valued columns of $\boldsymbol{\zeta}$ in (a), (d) relative importance of the input features based on the mean absolute coefficients of $\boldsymbol{T}$ in (b), and (e) test accuracy associated with using the top $n$ features in (d).}
\label{fig:knot_theory_regress}
\end{figure}

\begin{figure}[!ht]
\centering
{(a)}{\includegraphics[width=0.4\textwidth]{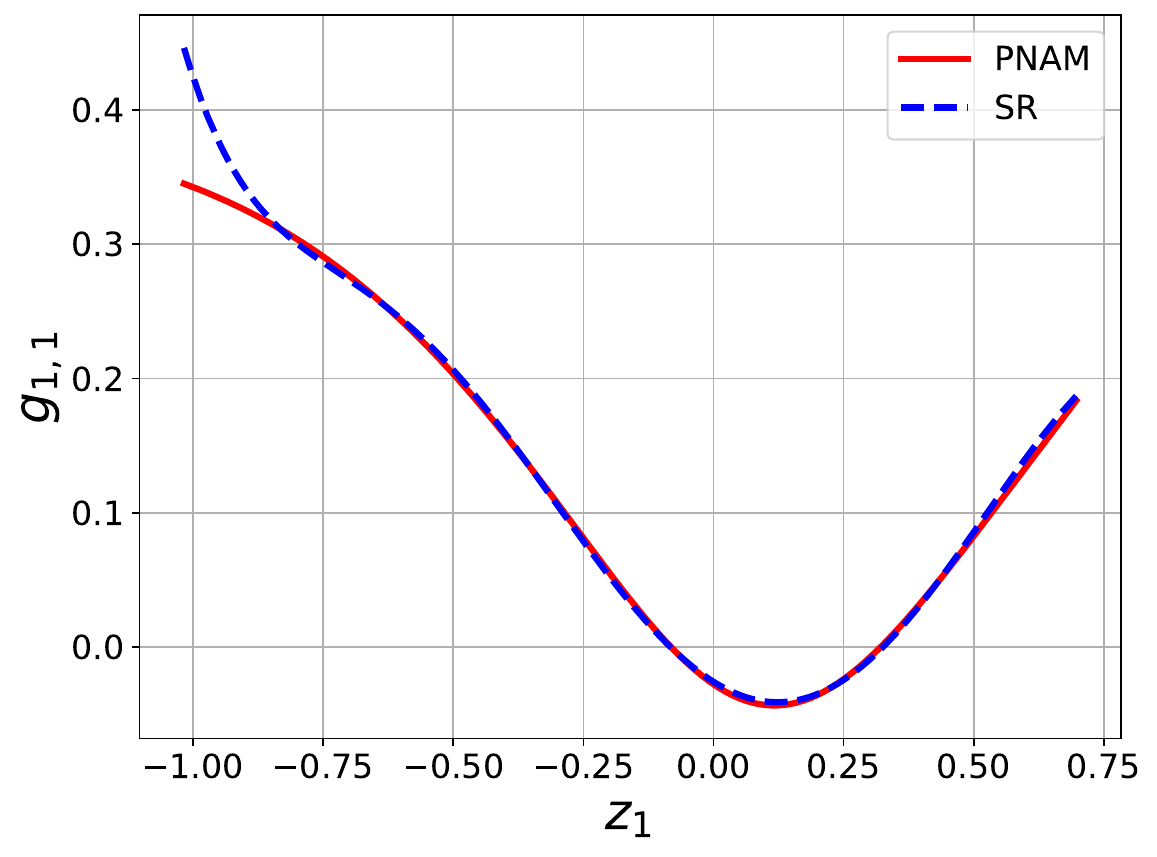}}
\includegraphics[width=0.4\textwidth]{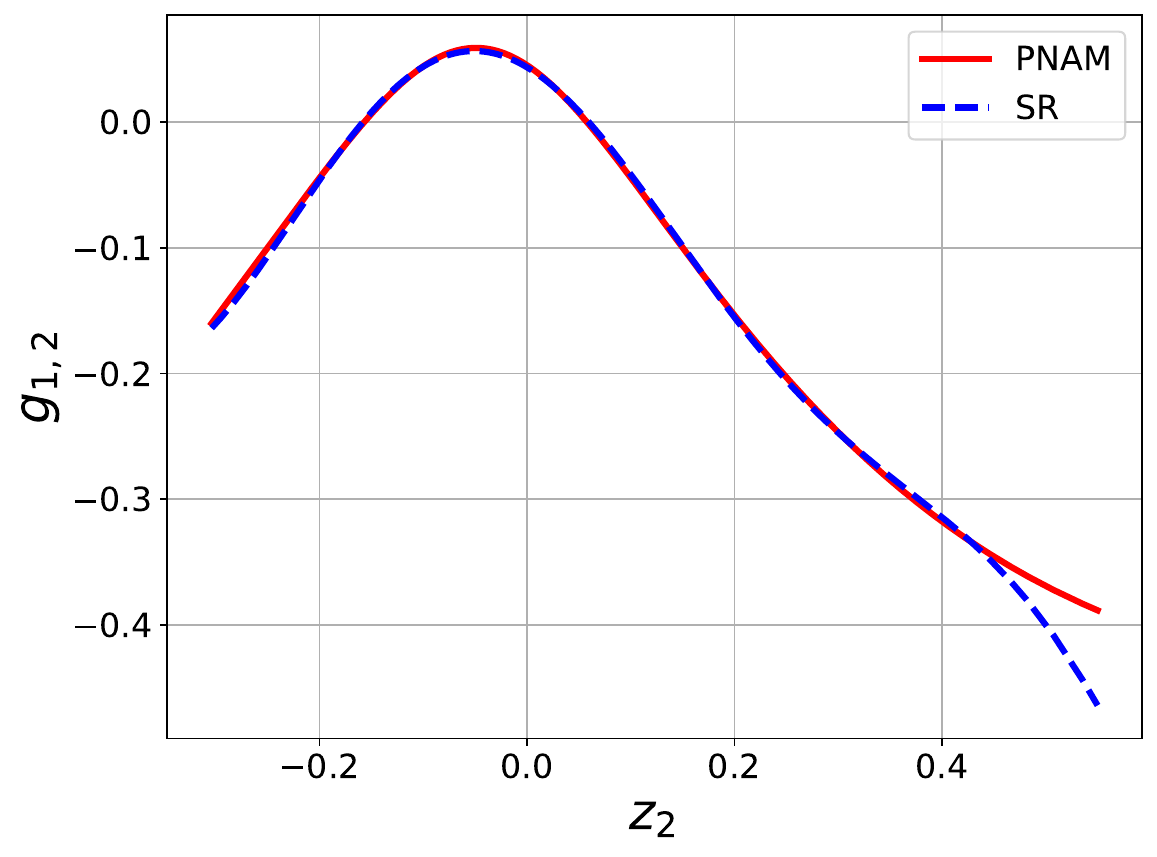}
{(b)}{\includegraphics[width=0.4\textwidth]{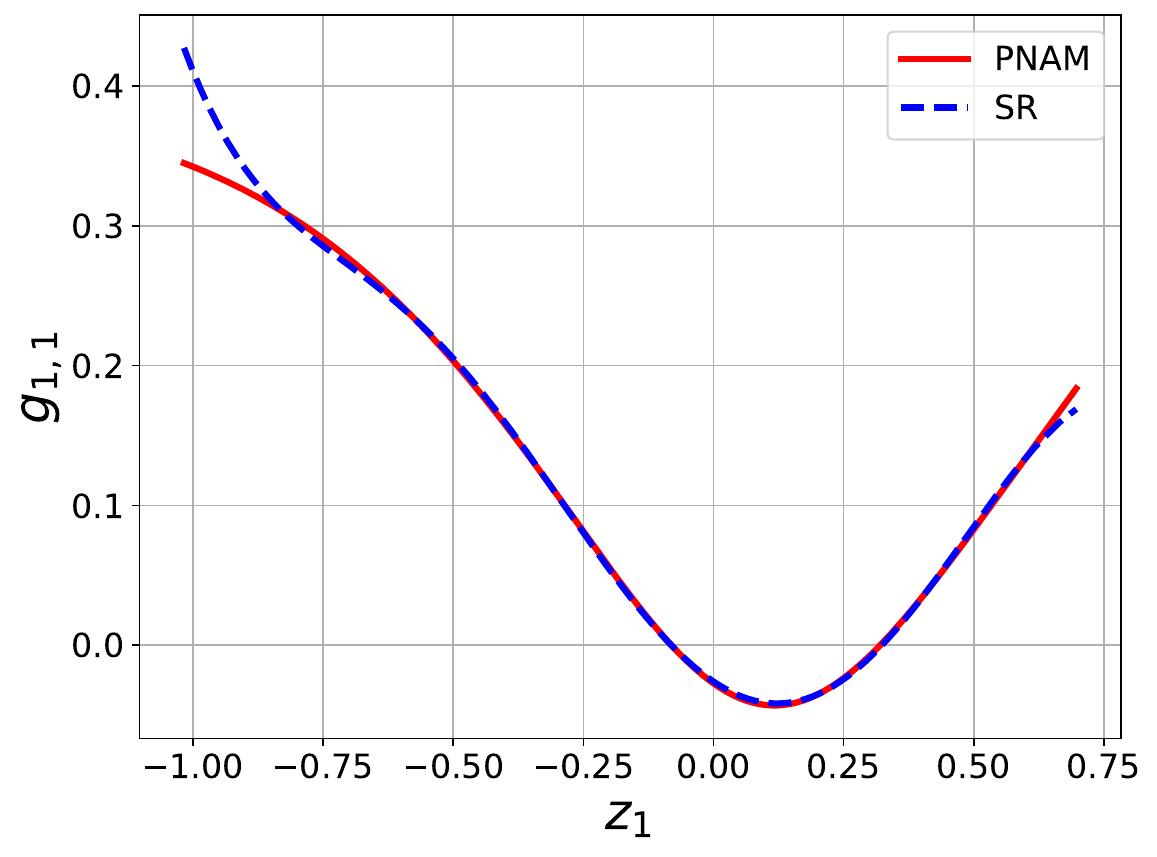}}
\includegraphics[width=0.4\textwidth]{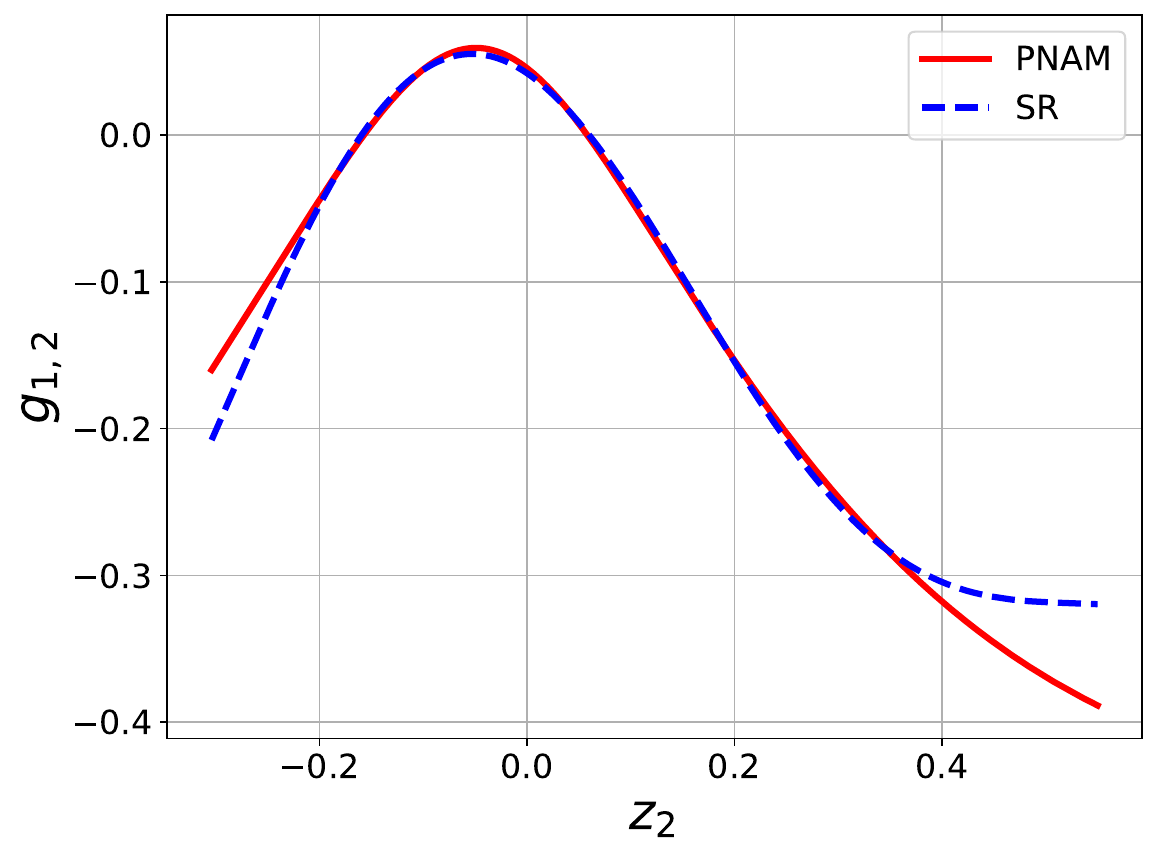}
\caption{Reproducibility of SR for learning 1D functions. Both sets of bases (PNAM) are from the run in Fig.~\ref{fig:knot_theory_regress}, with (a) the two bases (SR) comprising expression H and (b) the two bases (SR) comprising expression I in Table \ref{tab:knot_theory_regress_eqs}. The neural network parameterization of the PNAM, expression H, and expression I all use $n = 3$ features and achieve test accuracies of about 81\%.}
\label{fig:knot_theory_regress_g1}
\end{figure}

Figure \ref{fig:knot_theory_regress} shows a complete ranking of the inputs and the associated test accuracy of using the top $n$ inputs from one of the runs summarized in Table \ref{tab:knot_theory_regress_rank}.
The ranking in Fig.~\ref{fig:knot_theory_regress}(d) is determined from the mean absolute coefficients of the linear transformation $\boldsymbol{T}$ in Fig.~\ref{fig:knot_theory_regress}(b).
As intended by the regularization in Section \ref{subsec:constraints}, the learned $\boldsymbol{T}$ (i) is predominantly diagonal due to the penalty $\ell_3$ in Eq.~\eqref{eq:l3} and (ii) is sparse due to the constraint $\ell_4$ in Eq.~\eqref{eq:l4_l5}, while retaining the essential off-diagonal entries.
The post-processed transformation $\boldsymbol{T}$ with $n = 3$ in Fig.~\ref{fig:knot_theory_regress}(c) and the weighting coefficients $\boldsymbol{\zeta}$ of just two nonzero bases in Fig.~\ref{fig:knot_theory_regress}(a) are then leveraged to construct expressions H and I in Table \ref{tab:knot_theory_regress_eqs}.
Although each basis is evolved for only one minute using a few basic operators, Fig.~\ref{fig:knot_theory_regress_g1} demonstrates that SR can consistently approximate the single-variable bases because of their simple 1D nature.
Expressions H and I achieve test accuracies of 81.2 and 81.4\%, respectively, comparable to the test accuracy of 81.3\% in Fig.~\ref{fig:knot_theory_regress}(e) for $n = 3$ inputs.
Moreover, Table \ref{tab:knot_theory_regress_time} reveals that both expressions are more than two times faster for inference than the fully connected MLP.

\begin{table}[!ht]
\caption{Mathematical expressions for the knot dataset. Inputs $\chi_7$, $\chi_8$, $\chi_9$, and $\chi_{10}$ are the cusp volume, longitudinal translation, meridional translation (imaginary part), and meridional translation (real part), respectively. Expressions A to F are reproduced from Table 4 of \citet{liu2025kan}. A factor of $\nicefrac{1}{2}$ is added to expression A for consistency with DeepMind's findings \citep{davies2021advancing, davies2024signature}. Expression D has missing parentheses, so it cannot be evaluated. A factor of $\nicefrac{1}{2}$ is added to expression E for consistency with expression A. For expressions G and H, bases $\{g_{ij}\}$ in Eq.~\eqref{eq:sr_loss} use addition, subtraction, multiplication, and square as operators; they additionally use exponential, sine, and tangent for expression I. Every constant in the expressions is counted as one parameter. See our code (footnote \ref{fn:code}) to reproduce these results.}
\label{tab:knot_theory_regress_eqs}
\centering
\resizebox{\textwidth}{!}{
\begin{threeparttable}
\begin{tabular}{lllllll}
\toprule
\multirow{2}{*}{ID} & \multirow{2}{*}{Expression} & \multirow{2}{*}{PC\tnote{$\dagger$}} & \multirow{2}{*}{Discovered by} & \multicolumn{2}{c}{Eval. of test acc.} & \multirow{2}{*}{Total acc.} \\
\cmidrule(lr){5-6}
& & & & Reported & Our & \\
\midrule
\multirow{2}{*}{A} & \multirow{2}{*}{$\frac{\chi_8 \chi_{10}}{2 (\chi_{10}^2 + \chi_9^2)}$} & \multirow{2}{*}{3} & \multirow{2}{*}{Human (DM)} & \multirow{2}{*}{83.1\%} & \multirow{2}{*}{74.5\%} & \multirow{2}{*}{73.8\%} \\
& & & & & \\
\hline
\multirow{2}{*}{B} & $-0.02 \sin(4.98 \chi_9 + 0.85) + 0.08 |4.02 \chi_{10} + 6.28| - 0.52 \ -$ & \multirow{2}{*}{12} & \multirow{2}{*}{[3, 1] KAN} & \multirow{2}{*}{62.6\%} & \multirow{2}{*}{27.0\%} & \multirow{2}{*}{26.8\%} \\
& $0.04 e^{-0.88 (1 - 0.45 \chi_8)^2}$ & & & & \\
\hline
\multirow{2}{*}{C} & $0.17 \tan(-1.51 + 0.1 e^{-1.43 (1 - 0.4 \chi_9)^2 + 0.09 e^{-0.06 (1 - 0.21 \chi_8)^2}} \ +$ & \multirow{2}{*}{17} & \multirow{2}{*}{[3, 1, 1] KAN} & \multirow{2}{*}{71.9\%} & \multirow{2}{*}{41.7\%} & \multirow{2}{*}{41.5\%} \\
& $1.32 e^{-3.18 (1 - 0.43 \chi_{10})^2})$ & & & & \\
\hline
\multirow{4}{*}{D} & $-0.09 + 1.04 \exp(-9.59 (-0.62 \sin(0.61 \chi_{10} + 7.26)) \ -$ & \multirow{4}{*}{29} & \multirow{4}{*}{[3, 2, 1] KAN} & \multirow{4}{*}{84.0\%} & \multirow{4}{*}{--} & \multirow{4}{*}{--} \\
& $0.32 \tan(0.03 \chi_8 - 6.59) + 1 - 0.11 e^{-1.77 (0.31 - \chi_9)^2)^2} \ -$ & & & & & \\
& $1.09 e^{-7.6 (0.65 (1 - 0.01 \chi_8)^3} + 0.27 \arctan(0.53 \chi_9 - 0.6) \ +$ & & & & & \\
& $0.09 + \exp(-2.58 (1 - 0.36 \chi_{10})^2))$ & & & & & \\
\hline
\multirow{2}{*}{E} & \multirow{2}{*}{$\frac{4.76 \chi_8 \chi_{10}}{2 (3.09 \chi_9 + 6.05 \chi_{10}^2 + 3.54 \chi_9^2)}$} & \multirow{2}{*}{7} & [3, 2, 1] KAN & \multirow{2}{*}{82.8\%} & \multirow{2}{*}{79.3\%} & \multirow{2}{*}{79.3\%} \\
& & & $+$ Pad\'{e} approx. & & & \\
\hline
\multirow{2}{*}{F} & \multirow{2}{*}{$\frac{2.94 - 2.92 (1 - 0.10 \chi_{10})^2}{0.32 (0.18 - \chi_{10})^2 + 5.36 (1 - 0.04 \chi_8)^2 + 0.50}$} & \multirow{2}{*}{13} & \multirow{2}{*}{$\frac{[3, 1] \ \mathrm{KAN}}{[3, 1] \ \mathrm{KAN}}$} & \multirow{2}{*}{77.8\%} & \multirow{2}{*}{27.0\%} & \multirow{2}{*}{26.8\%} \\
& & & & & \\
\hline
\multirow{5}{*}{G} & $12.766 (0.132 (-\chi_{10} + 0.035 \chi_8 + 0.157)^2 \ +$ & \multirow{5}{*}{31} & & \multirow{5}{*}{--} & \multirow{5}{*}{75.9\%} & \multirow{5}{*}{75.7\%} \\
& $0.592 (-0.23 \chi_{10} + 0.008 \chi_8 + 1)^2 (0.162 \chi_{10} - 0.006 \chi_8 \ +$ & & $8 \times [1, 64, 32, 1]$ & & & \\
& $0.076) - 1)^4 + 7.202 (0.871 \chi_{10} + 0.029 \chi_8 - (0.229 \chi_{10} \ +$ & & PNAM ($n = 2$) & & & \\
& $0.008 \chi_8 - 0.103) (0.229 \chi_{10} + 0.008 \chi_8 + 0.159 (\chi_{10} \ +$ & & & & & \\
& $0.033 \chi_8 - 0.449)^2 + 0.625) - 0.173)^2 - 10.643$ & & & & & \\
\hline
\multirow{8}{*}{H} & $26 ((0.096 \chi_{10} - 0.002 \chi_7 + 0.004 \chi_8 - 0.237) (0.267 \chi_{10} \ -$ & \multirow{8}{*}{52} & & \multirow{8}{*}{--} & \multirow{8}{*}{81.2\%} & \multirow{8}{*}{80.9\%} \\
& $0.004 \chi_7 + 0.011 \chi_8 + 0.119 (\chi_{10} - 0.016 \chi_7 + 0.04 \chi_8 \ -$ & & & & & \\
& $0.316)^2 - 1)^2 + 0.133) (0.734 \chi_{10} - 0.012 \chi_7 + 0.029 \chi_8 \ +$ & & & & & \\
& $0.418) + 2.054 (0.589 \chi_{10} + 0.018 \chi_7 - 0.027 \chi_8 + 1)^2 \ -$ & & $8 \times [1, 64, 32, 1]$ & & \\
& $4.056 (\chi_{10} + 0.031 \chi_7 - 0.046 \chi_8 + 0.195)^2 (0.107 \chi_{10} \ +$ & & PNAM ($n = 3$) & & \\
& $0.003 \chi_7 - 0.005 \chi_8 - 0.064 (\chi_{10} + 0.031 \chi_7 - 0.046 \chi_8 \ +$ & & & & \\
& $0.087 (\chi_{10} + 0.031 \chi_7 - 0.046 \chi_8 - 0.165 (\chi_{10} + 0.031 \chi_7 \ -$ & & & & \\
& $0.046 \chi_8 + 0.195)^2 + 0.195)^2 + 0.195)^2 + 1)^2 - 0.895$ & & & & \\
\hline
\multirow{6}{*}{I} & $2.574 \chi_{10} + 0.078 \chi_7 - 0.13 \chi_8 + 26 (0.233 (-\chi_{10} \ +$ & \multirow{6}{*}{34} & & \multirow{6}{*}{--} & \multirow{6}{*}{81.4\%} & \multirow{6}{*}{81.0\%} \\
& $0.016 \chi_7 - 0.04 \chi_8 - 0.597)^2 \sin^2(0.367 \chi_{10} - 0.006 \chi_7 \ +$ & & & & & \\
& $0.015 \chi_8 - 0.719) + \sin(0.777 \chi_{10} - 0.013 \chi_7 + 0.031 \chi_8 \ +$ & & $8 \times [1, 64, 32, 1]$ & & & \\
& $0.262)) (0.168 \chi_{10} - 0.003 \chi_7 + 0.007 \chi_8 - 0.053) \ +$ & & PNAM ($n = 3$) & & \\
& $15.626 (-\sin(0.54 \chi_{10} + 0.017 \chi_7 - 0.025 \chi_8 + 0.105) \ -$ & & & & \\
& $0.049)^2 (-0.165 \chi_{10} - 0.005 \chi_7 + 0.008 \chi_8 - 0.645) + 0.509$ & & & & \\
\bottomrule
\end{tabular}
\begin{tablenotes}
\item[$\dagger$] Parameter count is abbreviated as PC.
\end{tablenotes}
\end{threeparttable}}
\end{table}

\begin{table}[!ht]
\caption{Training and inference times of the neural networks in Table \ref{tab:knot_theory_regress_compare}, along with inference times of expressions G to I in Table \ref{tab:knot_theory_regress_eqs}. Each basis comprising expressions G to I is evolved for one minute. Training times are obtained from an NVIDIA A100-SXM4-40GB GPU, while inference times are obtained from an Apple M1 CPU with 8 GB of memory. The test accuracy is repeated to aid comparison.}
\label{tab:knot_theory_regress_time}
\centering
\begin{tabular}{llll}
\toprule
Method & Training time (s/epoch) & Inference time (ns/sample) & Test acc. \\
\midrule
MLP & 2.1 & 290 & $94.7 \pm 0.2$\% \\
NAM & 14 & 4300 & $65.0 \pm 0.2$\% \\
PNAM & 8.9 & 2100 & $85.5 \pm 0.3$\% \\
Expression G & -- & 78 & 75.9\% \\
Expression H & -- & 130 & 81.2\% \\
Expression I & -- & 120 & 81.4\% \\
\bottomrule
\end{tabular}
\end{table}

\begin{figure}[!ht]
\centering
\includegraphics[width=0.5\textwidth]{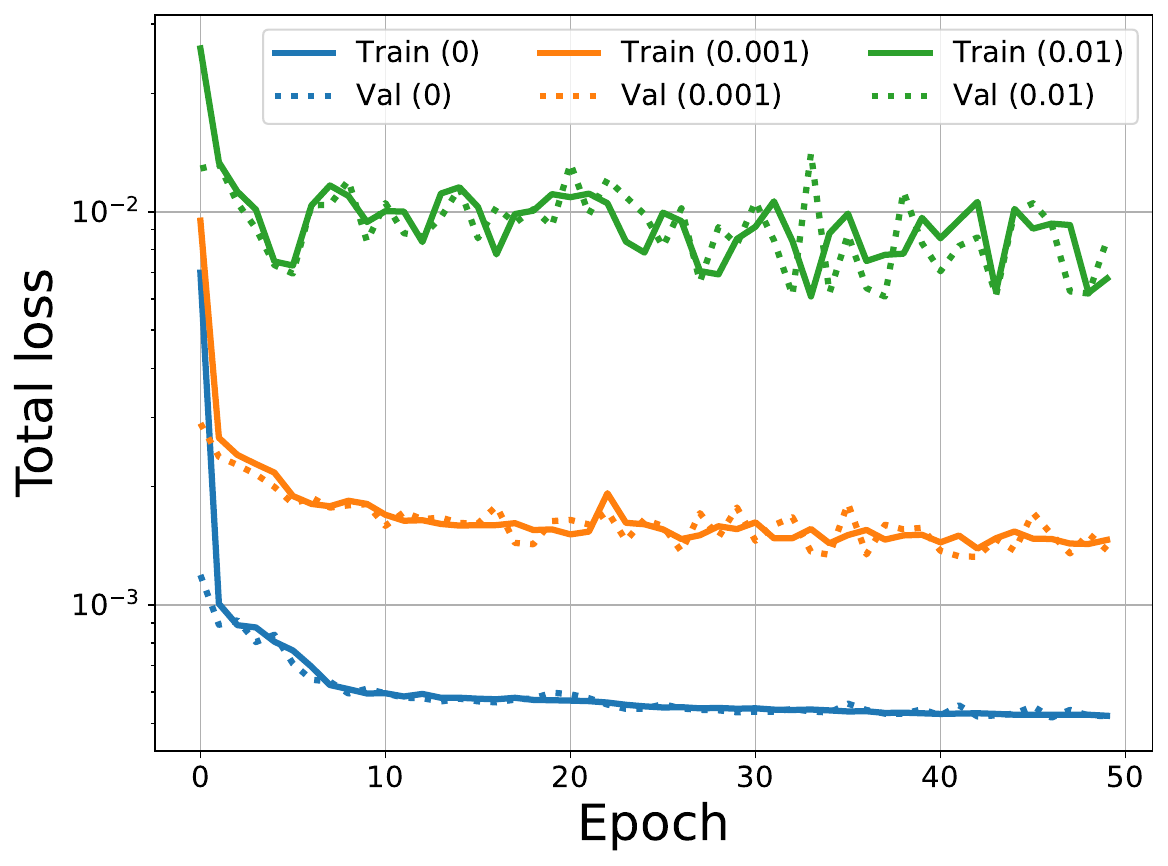}
\caption{Effects of regularization weights $w_1 = \dots = w_5 = w$ on the convergence of PNAMs for the knot dataset. Total loss corresponds to the loss function in Eq.~\eqref{eq:nn_loss}. The three PNAMs use the same projection dimension $M = 8$, and each basis MLP has two hidden layers of 64 and 32 neurons.}
\label{fig:knot_theory_regress_loss}
\end{figure}

\begin{figure}[!ht]
\centering
{(a)}{\includegraphics[width=0.45\textwidth]{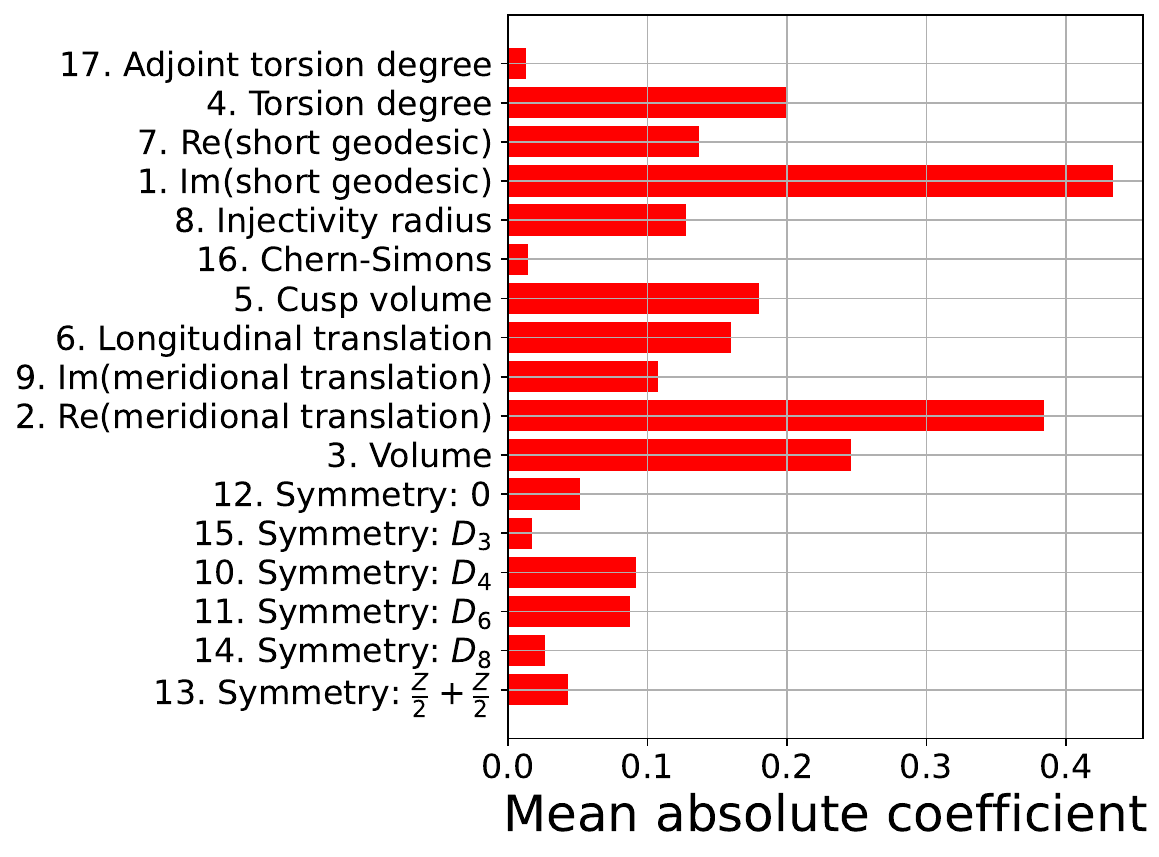}}
{(b)}{\includegraphics[width=0.45\textwidth]{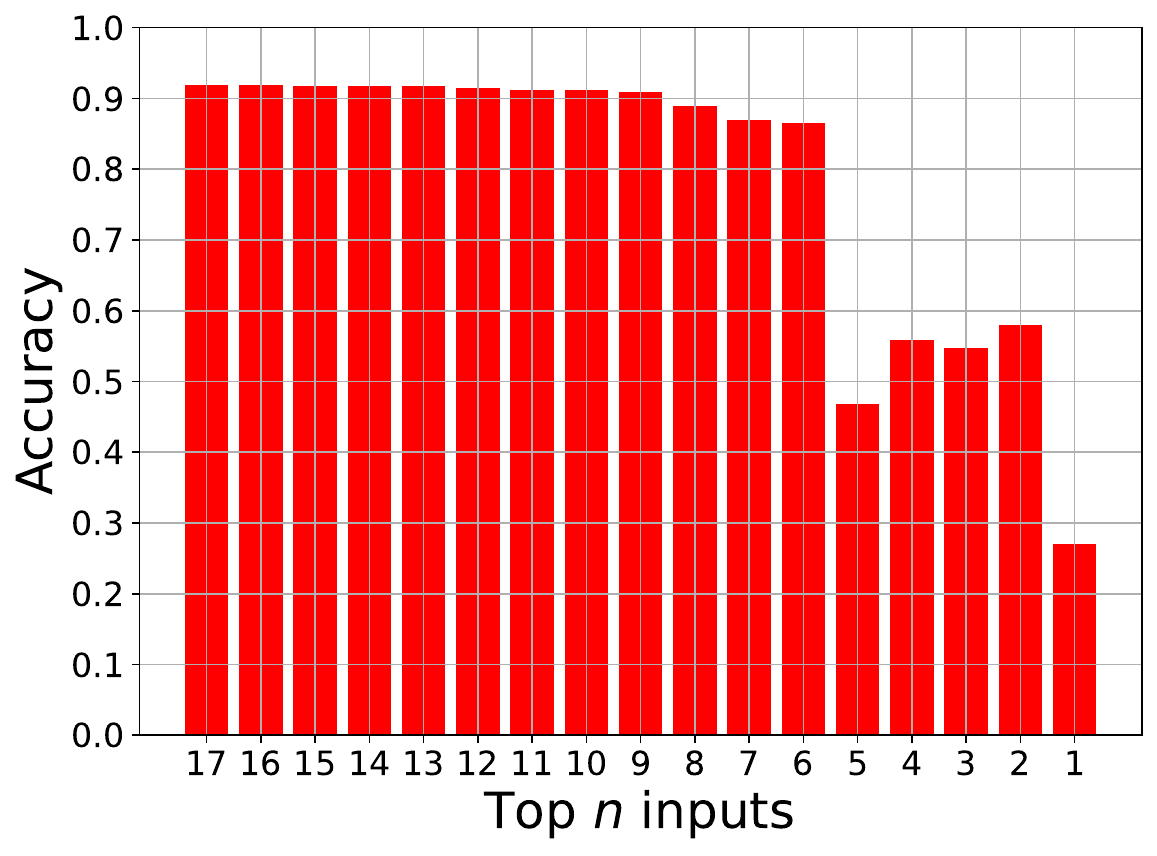}}
{(c)}{\includegraphics[width=0.45\textwidth]{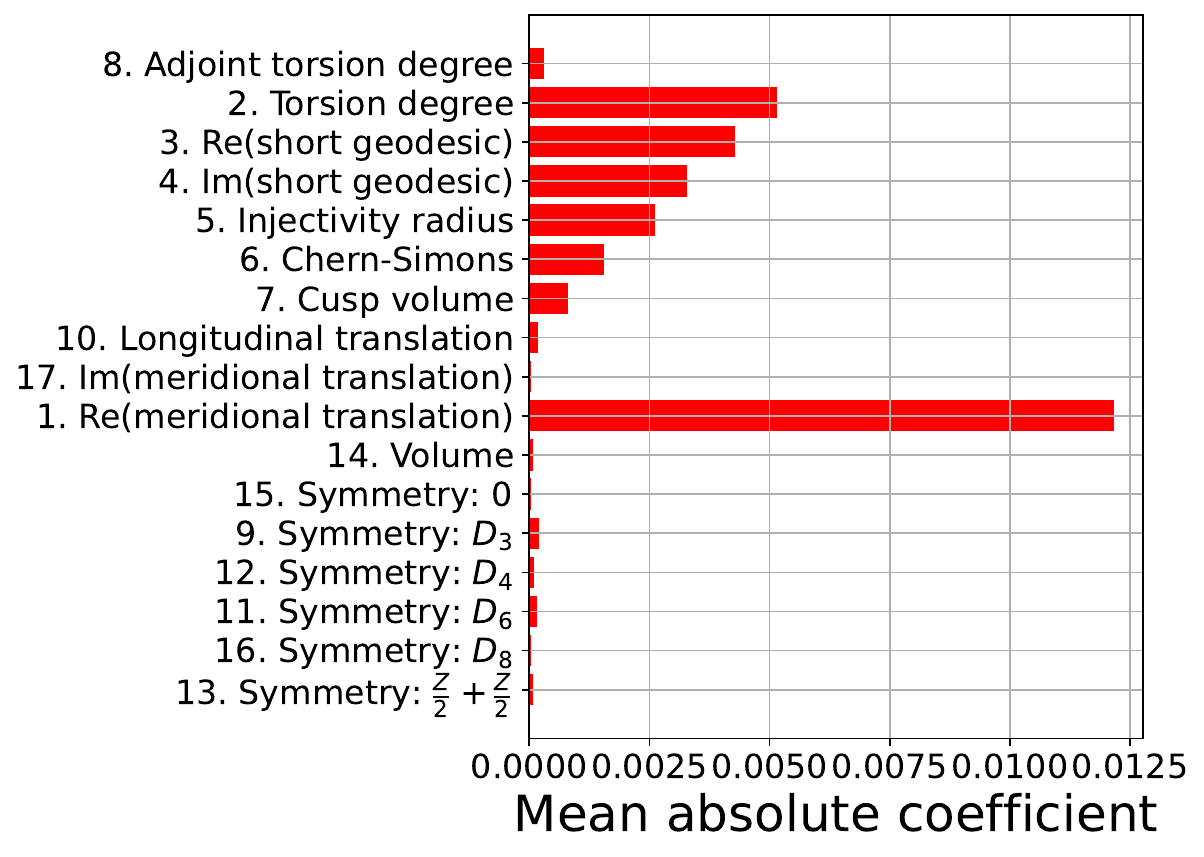}}
{(d)}{\includegraphics[width=0.45\textwidth]{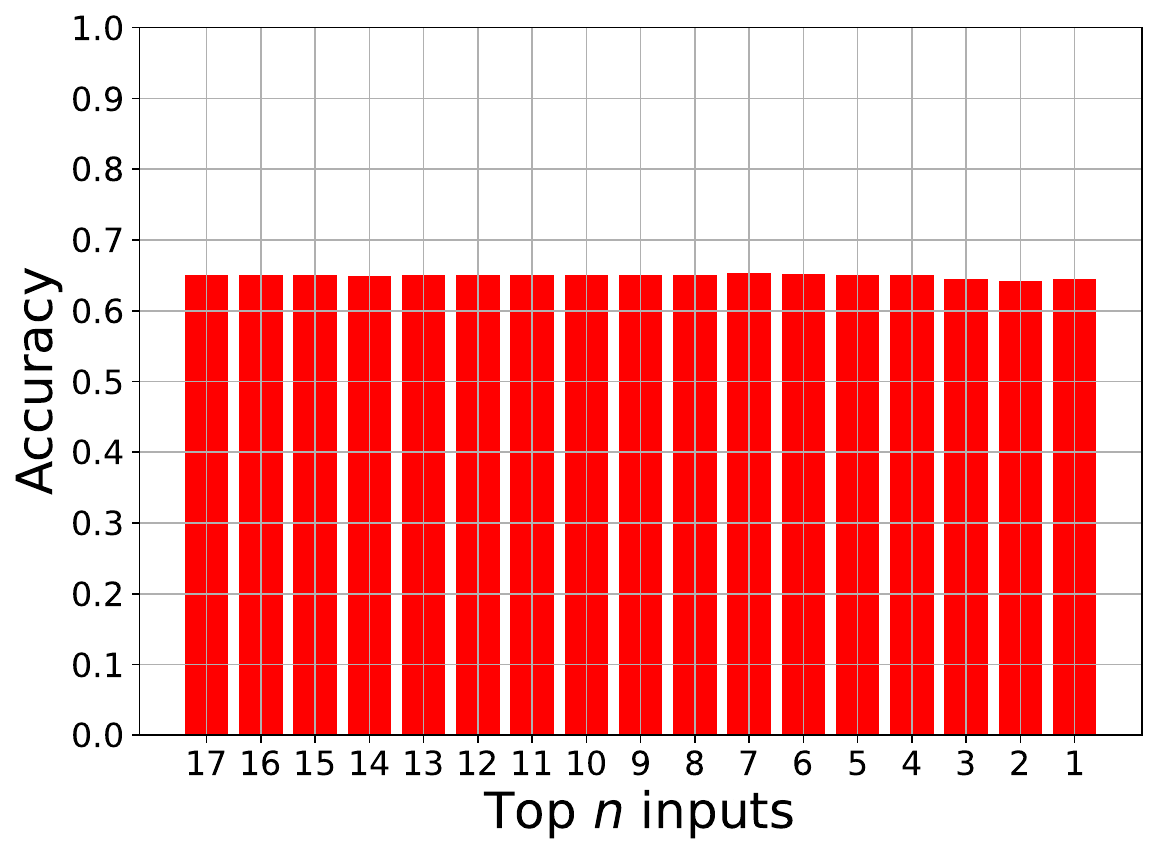}}
\caption{Effects of regularization weights $w_1 = \dots = w_5 = w$ on the input ranking and test accuracy of PNAMs for the knot dataset. Regularization weight $w = 0$ for (a) and (b); $w = 0.01$ for (c) and (d). (a) and (c): Relative importance of the input features based on the mean absolute coefficients. (b) and (d): Test accuracy associated with using the top $n$ features in (a) and (c), respectively. Results for $w = 0.001$ are provided in Fig.~\ref{fig:knot_theory_regress}(d) and (e).}
\label{fig:knot_theory_regress_w}
\end{figure}

Recall that the above results correspond to the regularization weight $w = 0.001$, chosen to balance the trade-off between accuracy and sparsity.
While increasing $w$ can make PNAMs sparser, Fig.~\ref{fig:knot_theory_regress_loss} indicates that overprioritizing sparsity can lead to less stable training, resulting in poorer performance.
Specifically, setting $w = 0.01$ results in such a sparse $\boldsymbol{T}$ (see the small coefficient values in Fig.~\ref{fig:knot_theory_regress_w}(c)) that the accuracy in Fig.~\ref{fig:knot_theory_regress_w}(d) is poor regardless of $n$.
On the other hand, Fig.~\ref{fig:knot_theory_regress_w}(b) suggests that setting $w = 0$---thereby deactivating the regularization constraints---improves the accuracy for large $n$, due to the extensive number of coefficients with large values in Fig.~\ref{fig:knot_theory_regress_w}(a), but reduces it for small $n$, due to the erroneous ranking of the input features.
Although the exact ranking may not be correct, this PNAM can still narrow down a set of variables (e.g., $n = 6$) that is relevant for predictive accuracy while remaining competitive with the MLP in Table \ref{tab:knot_theory_regress_compare}.

\subsection{High-dimensional image classification with the MNIST dataset} \label{subsec:mnist}
To demonstrate that PNAMs can efficiently handle high-dimensional problems with numerous inputs, we evaluate their performance on the MNIST dataset \citep{lecun1998gradient}, highlighting their dimensionality reduction capability and providing visual insight into their input pruning mechanism.
The MNIST dataset contains 70,000 images of handwritten digits from 0 to 9, each of size $28 \times 28$ pixels, partitioned into 60,000 images for training/validation and 10,000 images for testing.
Each image is reshaped into a 784-dimensional feature vector of raw pixel intensities, resulting in a high-dimensional classification task where the model must map dense feature vectors to their corresponding digit labels.
This experiment illustrates the ability of PNAMs to preserve interpretability while achieving strong predictive accuracy by identifying and utilizing only the most informative inputs, effectively pruning irrelevant features.

\begin{figure}[!ht]
\centering
{(a)}{\includegraphics[width=0.45\textwidth]{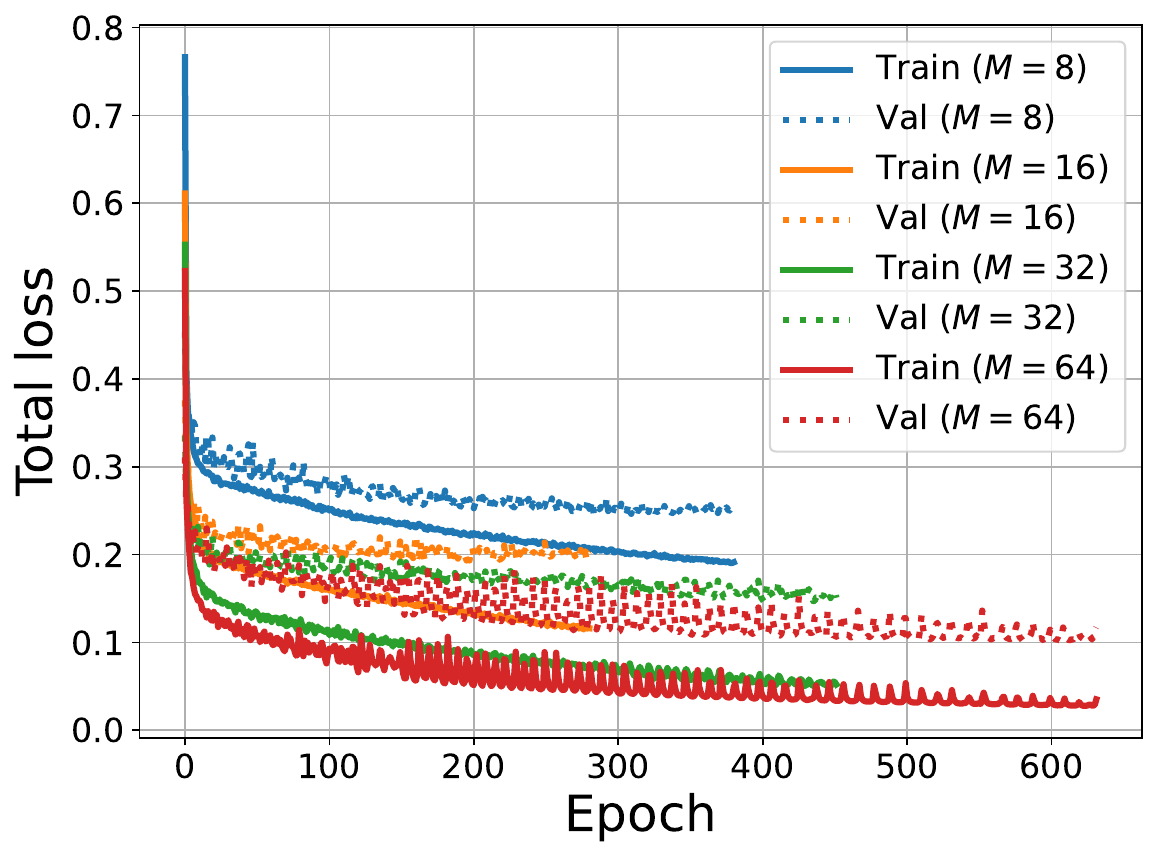}}
{(b)}{\includegraphics[width=0.45\textwidth]{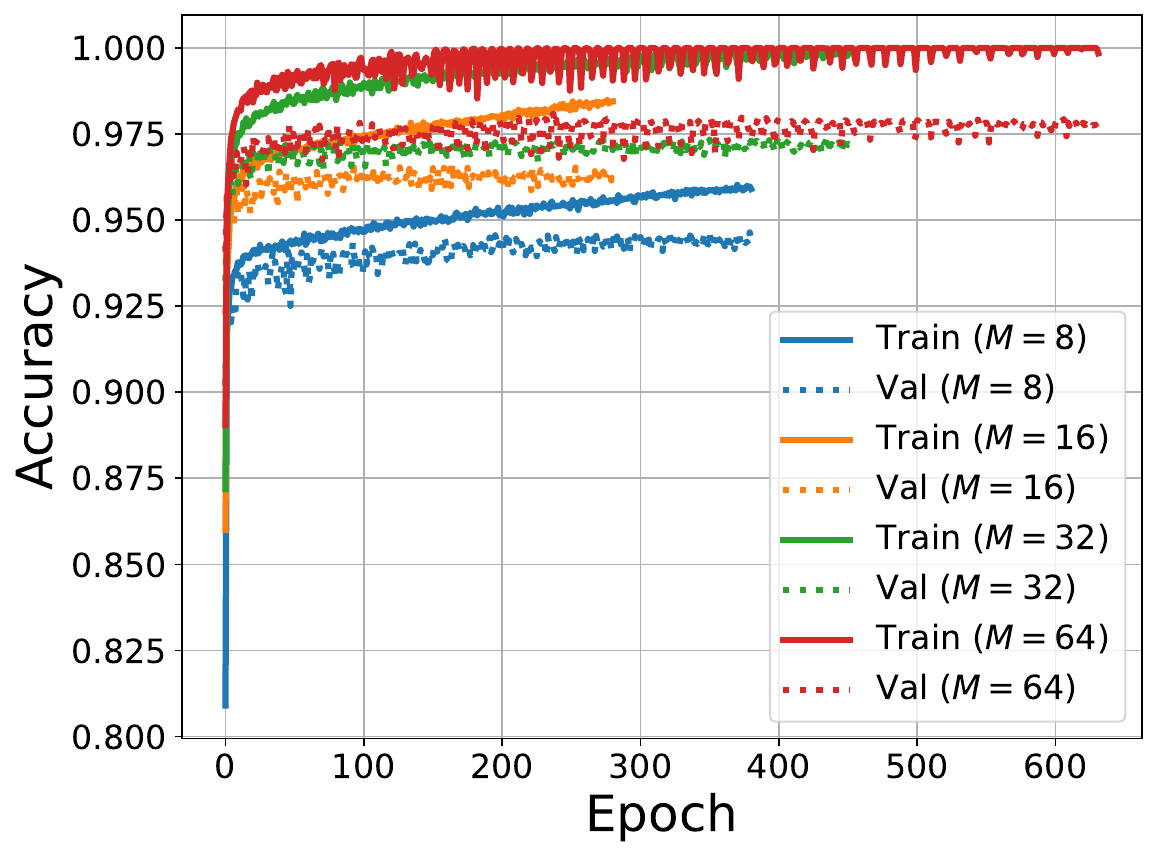}}
\caption{Effects of the projection dimension $M$ on the convergence and accuracy of PNAMs for the MNIST dataset. Training and validation histories of (a) the loss function in Eq.~\eqref{eq:nn_loss} and (b) the accuracy. Each basis MLP has two hidden layers of 64 and 32 neurons; all models are trained using regularization weights $w_1 = \dots = w_5 = 0.01$.}
\label{fig:mnist_M}
\end{figure}

In Table \ref{tab:mnist_compare}, we compare PNAMs against a standard NAM and fully connected MLPs, all of which are trained for up to 1000 epochs with an early stopping patience of 50 epochs.
First, the linear transformation $\boldsymbol{T}$ allows us to project the original 784-dimensional vector into a lower-dimensional feature space of size $M$.
Even with a small $M = 8$, the resulting PNAM outperforms the standard NAM, despite the latter requiring one basis per input and having nearly 100 times more parameters.
Second, increasing $M$ enhances the expressivity of PNAMs, allowing them to fit the training data perfectly (see Fig.~\ref{fig:mnist_M}) and achieve a test accuracy of 98.1\% for $M = 64$.
Third, although PNAMs perform similarly to MLPs, suggesting comparable expressive power in practice, they retain the key advantage of interpretability.
Unlike black-box MLPs, PNAMs offer insight into which variables are crucial for predictive accuracy, enabling model transparency and post-hoc analysis.

\begin{table}[!ht]
\caption{Performance comparison of different neural network architectures for the multi-class classification problem of predicting handwritten digits from the MNIST dataset. Training and validation histories of the PNAMs are provided in Fig.~\ref{fig:mnist_M}.}
\label{tab:mnist_compare}
\centering
\begin{tabular}{lllll}
\toprule
Method & Architecture & Parameter count & Training acc. & Test acc. \\
\midrule
MLP & 3 layers: [784, 64, 32, 10] & $1.2 \times 10^4$ & 99.9\% & 96.1\% \\
MLP & 3 layers: [784, 128, 64, 10] & $4.9 \times 10^4$ & 99.9\% & 96.3\% \\
MLP & 4 layers: [784, 128, 64, 32, 10] & $6.6 \times 10^4$ & 99.9\% & 96.7\% \\
MLP & 4 layers: [784, 256, 128, 64, 10] & $2.6 \times 10^5$ & 99.9\% & 97.1\% \\
NAM & 3 layers: $784 \times [1, 64, 32, 10]$ & $9.6 \times 10^6$ & 95.0\% & 93.0\% \\
PNAM & 3 layers: $8 \times [1, 64, 32, 10]$ & $1.0 \times 10^5$ & 96.0\% & 94.7\% \\
PNAM & 3 layers: $16 \times [1, 64, 32, 10]$ & $2.1 \times 10^5$ & 98.5\% & 96.7\% \\
PNAM & 3 layers: $32 \times [1, 64, 32, 10]$ & $4.2 \times 10^5$ & 100\% & 97.5\% \\
PNAM & 3 layers: $64 \times [1, 64, 32, 10]$ & $8.4 \times 10^5$ & 100\% & 98.1\% \\
\bottomrule
\end{tabular}
\end{table}

\begin{figure}[!ht]
\centering
\includegraphics[width=0.5\textwidth]{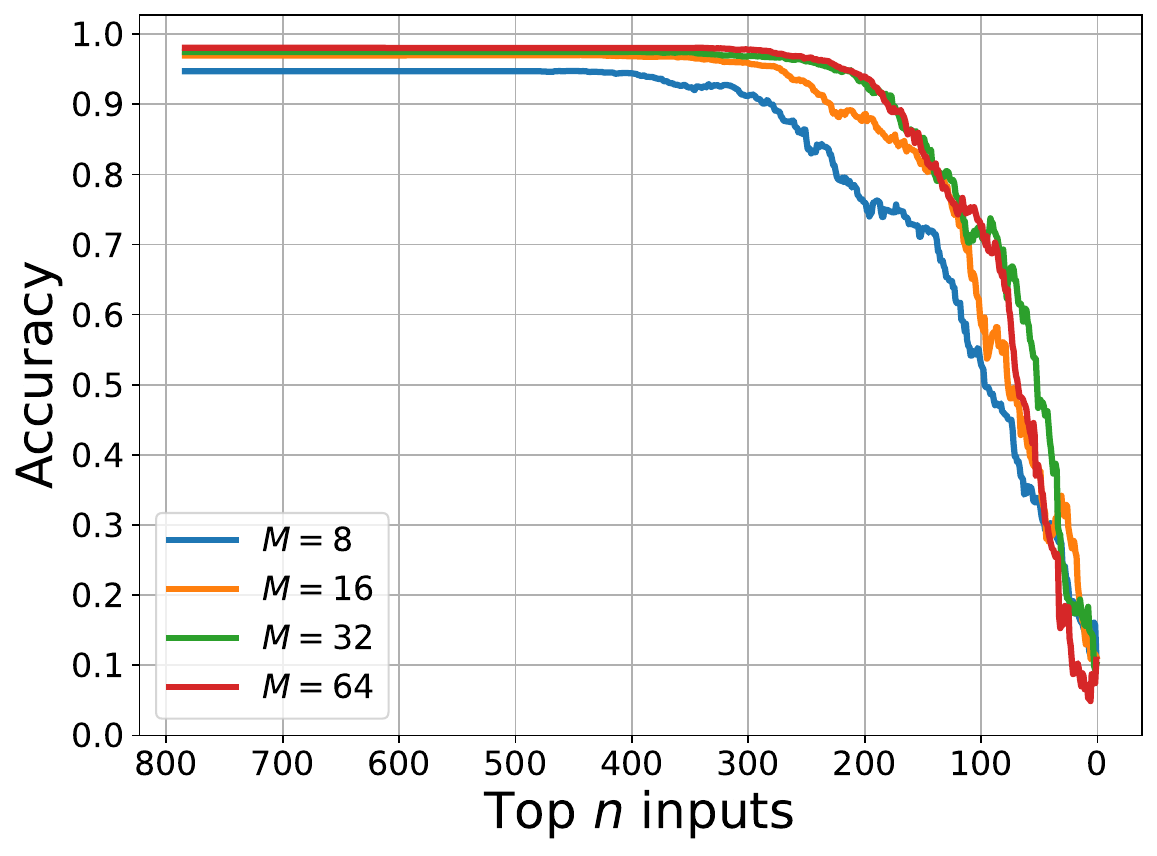}
\caption{Impact of input pruning on the test accuracy for the PNAMs in Table \ref{tab:mnist_compare}. Unimportant inputs are pruned according to the mean absolute coefficients (see Section \ref{subsec:sr}).}
\label{fig:mnist_M_acc}
\end{figure}

\begin{figure}[!ht]
\centering
{(a)}{\includegraphics[width=0.45\textwidth]{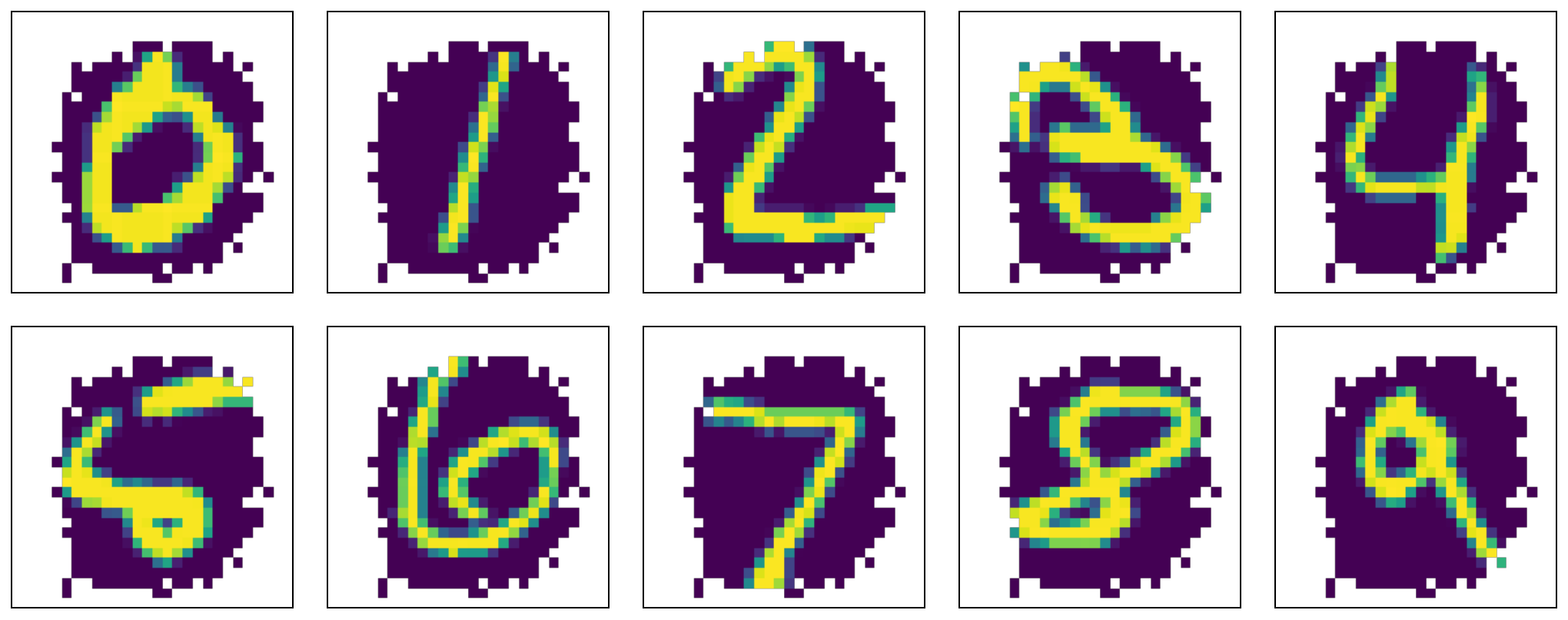}}
{(b)}{\includegraphics[width=0.45\textwidth]{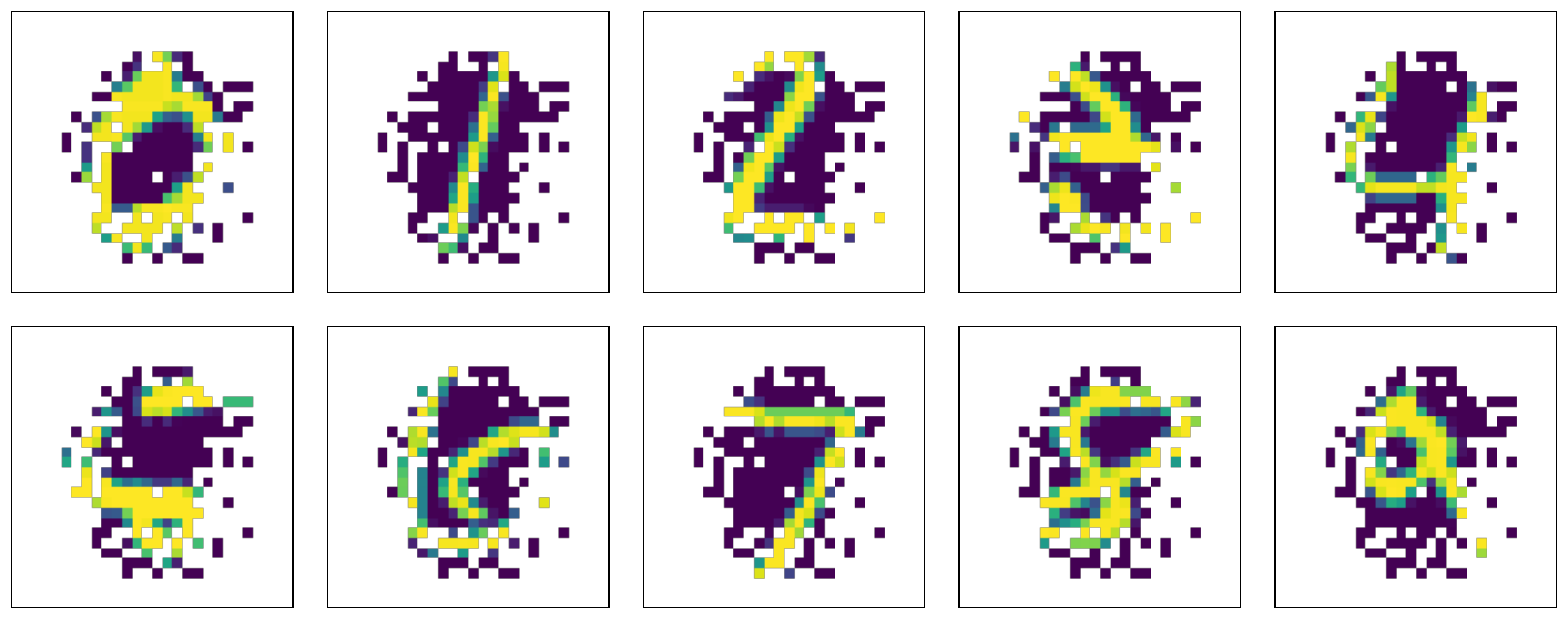}}
\caption{Demonstration of important inputs for the PNAM with $M = 64$ in Fig.~\ref{fig:mnist_M_acc}. Shown are the first 10 unique handwritten digits in the MNIST test set, arranged from 0 to 9, using (a) the $n = 400$ ($\sim$50\%) and (b) the $n = 200$ ($\sim$25\%) most relevant pixels, which achieve test accuracies of 98.0 and 94.0\%, respectively.}
\label{fig:mnist_digits}
\end{figure}

\begin{figure}[!ht]
\centering
\includegraphics[width=0.5\textwidth]{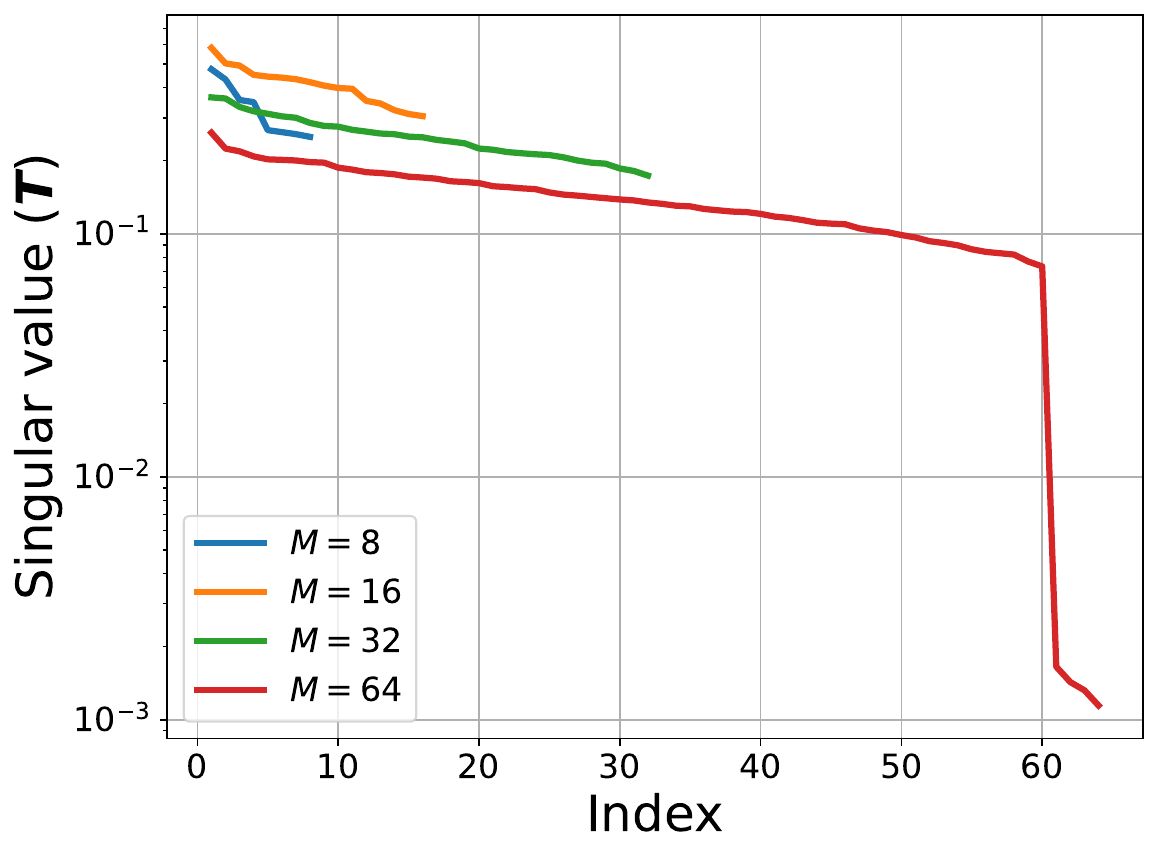}
\caption{Singular values of the linear transformation $\boldsymbol{T}$ for the PNAMs in Table \ref{tab:mnist_compare}. The linear decay and sudden drop in magnitude of the singular values on a log scale suggest that we can prune coefficients in $\boldsymbol{T}$ without severely degrading performance.}
\label{fig:mnist_S}
\end{figure}

Figure \ref{fig:mnist_M_acc} illustrates that PNAMs produce a consistent ranking of feature importance across different values of $M$, demonstrating robustness in their learned feature sensitivity.
This consistency supports effective pruning of nonessential inputs with minimal impact on accuracy.
Specifically, for the PNAM with $M = 64$, we remove the least important pixels and visualize two progressively smaller subsets in Fig.~\ref{fig:mnist_digits}.
Retaining only the top 400 and 200 relevant pixels, this model still achieves test accuracies of 98.0 and 94.0\%, respectively.
Visual inspection of the remaining pixels reveals that they occupy semantically meaningful regions, allowing human observers to easily identify the underlying digit classes.
Further insight is provided by Fig.~\ref{fig:mnist_S}, which shows that the majority of predictive information is concentrated in a small number of dominant singular values, enabling both dimensionality reduction and feature pruning.

\begin{figure}[!ht]
\centering
{(a)}{\includegraphics[width=0.45\textwidth]{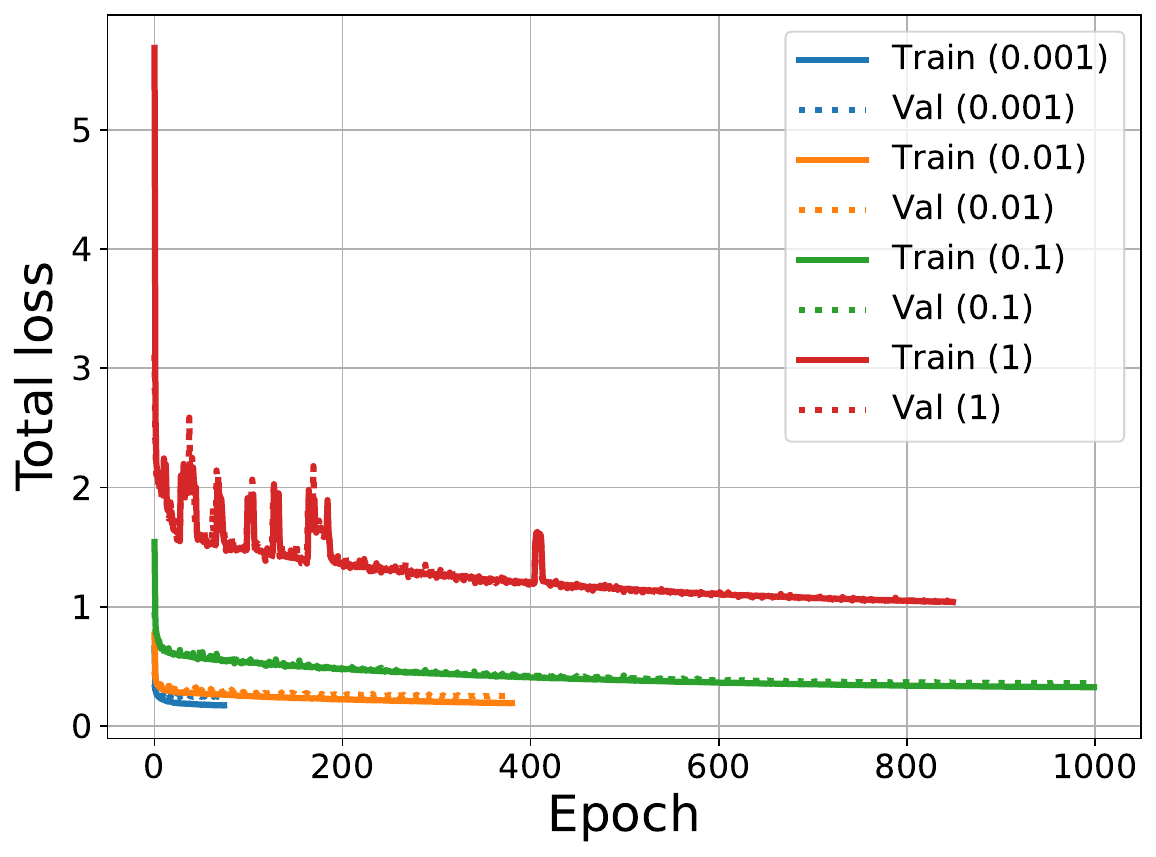}}
{(b)}{\includegraphics[width=0.45\textwidth]{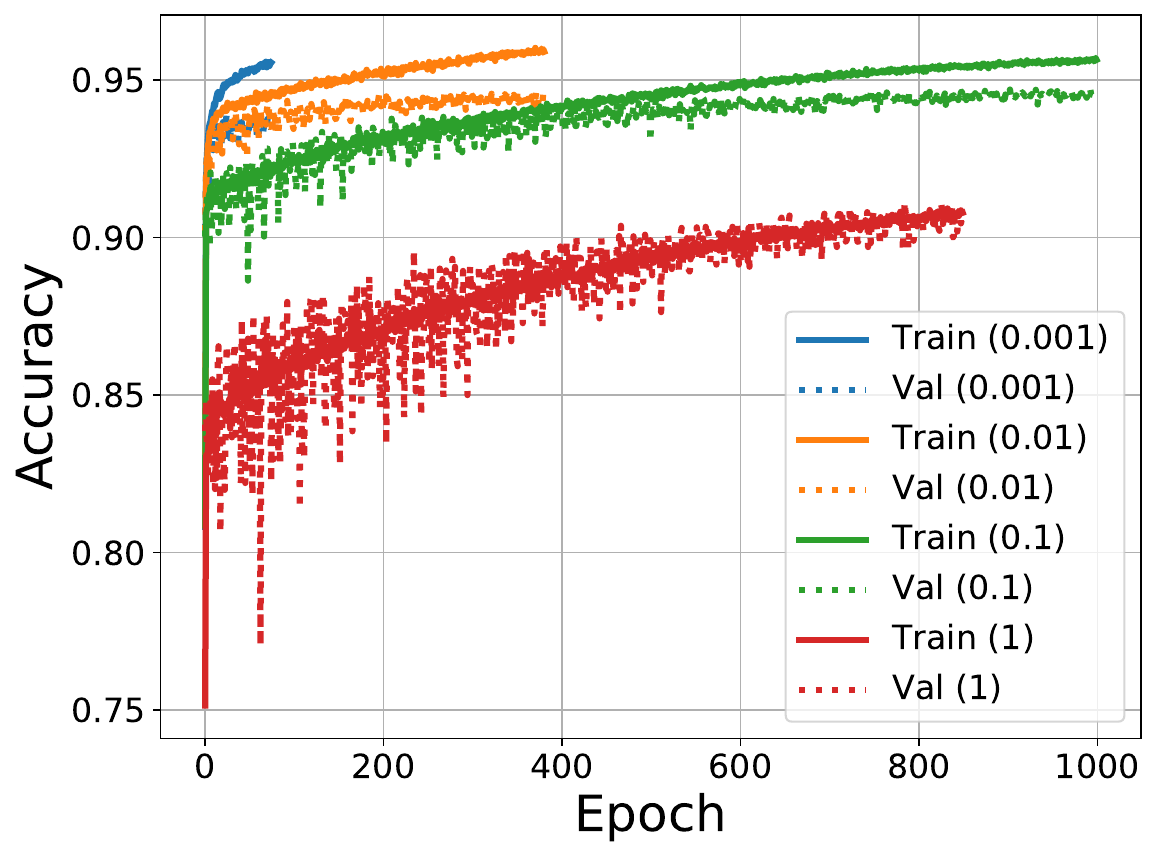}}
\caption{Effects of regularization weights $w_1 = \dots = w_5 = w$ on the convergence and accuracy of PNAMs for the MNIST dataset. Training and validation histories of (a) the loss function in Eq.~\eqref{eq:nn_loss} and (b) the accuracy. The four PNAMs use the same projection dimension $M = 8$, and each basis MLP has two hidden layers of 64 and 32 neurons.}
\label{fig:mnist_w}
\end{figure}

Finally, we examine the impact of the regularization weights $w_1, \dots, w_5$ on the performance of PNAMs.
Similar to the trend observed in Fig.~\ref{fig:knot_theory_regress_loss}, Fig.~\ref{fig:mnist_w} indicates that increasing the regularization weights leads to less stable training as the optimization objective shifts from prioritizing predictive accuracy to enforcing sparsity.
Nevertheless, as shown in Table \ref{tab:mnist_compare_w}, these regularization weights can be tuned to obtain a desired trade-off between accuracy and sparsity, tailored to the specific needs of the user.

\begin{table}[!ht]
\caption{Trade-off between accuracy and sparsity due to regularization weights $w_1 = \dots = w_5 = w$ for the PNAMs in Fig.~\ref{fig:mnist_w}. An element with $|T_{jk}| < 10^{-4}$ is considered zero, since setting such an element to zero has little to no effect on the test accuracy.}
\label{tab:mnist_compare_w}
\centering
\begin{tabular}{llll}
\toprule
$w$ & Training acc. & Test acc. & Percent of zero $T_{jk}$ \\
\midrule
0.001 & 95.6\% & 94.3\% & 14.1\% \\
0.01 & 96.0\% & 94.7\% & 32.8\% \\
0.1 & 95.7\% & 94.7\% & 68.9\% \\
1 & 90.9\% & 91.3\% & 76.5\% \\
\bottomrule
\end{tabular}
\end{table}

\subsection{Phase field theory for fracture of brittle solids} \label{subsec:phase_field}
We consider a phase field simulation of fracture in boron carbide (B$_4$C) with isotropic elasticity and isotropic fracture energy from \citep{clayton2023rate}, for quasi-static extension up to peak load.
The dataset contains 96 samples,\footnote{The phase field dataset is open-source with our code (footnote \ref{fn:code}) and can be used to study the benign or catastrophic overfitting of overparameterized neural networks \citep{mallinar2022benign} for a physical system.} homogenized from this simulation.
These samples are snapshots along a single loading path, and the 80/20 train-test split of Section \ref{sec:exp} is sampled randomly along this path, so the reported mean squared error (MSE) measures interpolation within the loading path.
Given the homogenized values of the axial strain, order parameter, and magnitude of the material gradient of the order parameter, the goal is to predict the average strain energy, phase energy, and axial stress.
Among the inputs, the order parameter measures local damage from 0 (undamaged) to 1 (fully fractured), and the magnitude of its material gradient measures how localized the damage is.
Physically, the strain energy is the elastic energy stored by deformation, the phase energy is the energy expended in creating fracture surfaces, and the stress is their total rate of change with strain.

Since the stress is obtained by differentiating the sum of the two energies, we frame this task as a two-output regression problem.
The MSE is employed as $\mathcal{L}$ in Eq.~\eqref{eq:nn_loss} to predict the strain energy and phase energy; to predict the stress, we use the following form for the physical constraint $\mathcal{L}_\mathrm{P}$:\footnote{Ginzburg--Landau kinetics \citep{gurtin1996generalized}, or its quasi-static reduction in the present case, could be added as a constraint if one has access to the loading rate and the Laplacian of the order parameter. See \citep{miehe2010thermodynamically} for background on phase field fracture mechanics and corresponding numerical models.}
\[
\mathcal{L}_\mathrm{P} = \frac{w_{\mathrm{P}1}}{D} \sum_{\nu = 1}^D \left((y_{1, 1}^\prime + y_{2, 1}^\prime)|_\nu - \frac{\partial (\widehat{y}_1(\boldsymbol{\chi}; \boldsymbol{T}, \boldsymbol{\Theta}, \boldsymbol{\zeta}, \boldsymbol{\epsilon}) + \widehat{y}_2(\boldsymbol{\chi}; \boldsymbol{T}, \boldsymbol{\Theta}, \boldsymbol{\zeta}, \boldsymbol{\epsilon}))}{\partial \chi_1}\bigg|_\nu\right)^2,
\]
where $y_{1, 1}^\prime$ and $y_{2, 1}^\prime$ are the derivatives of the strain energy and phase energy, respectively, with respect to the strain $\chi_1$.
We set the regularization weight $w_{\mathrm{P}1} = 1$.\footnote{In the implementation, the stress residuals are concatenated with the two energy residuals and a single MSE is taken over all three outputs.}
In addition, every training and validation batch is augmented with one reference point at the undeformed state (zero inputs with zero targets), anchoring the predicted energies and stress at the origin.

\begin{figure}[!ht]
\centering
\includegraphics[width=0.24\textwidth]{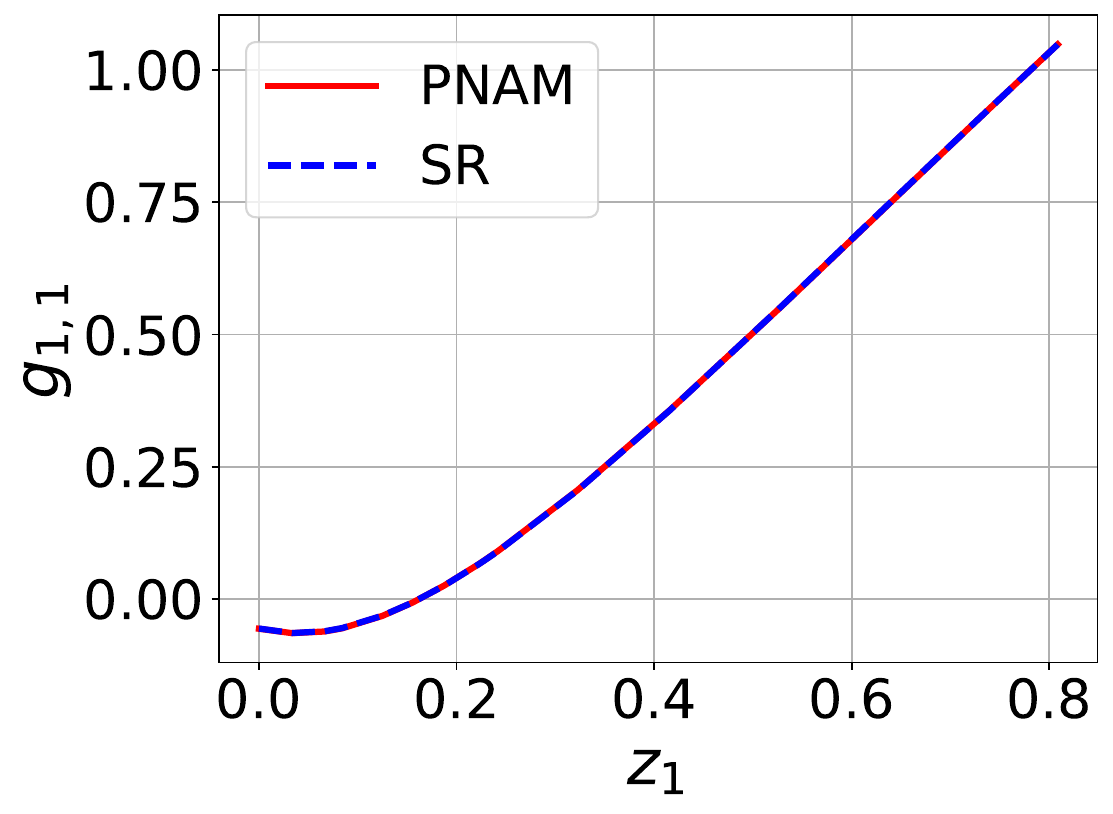}
\includegraphics[width=0.24\textwidth]{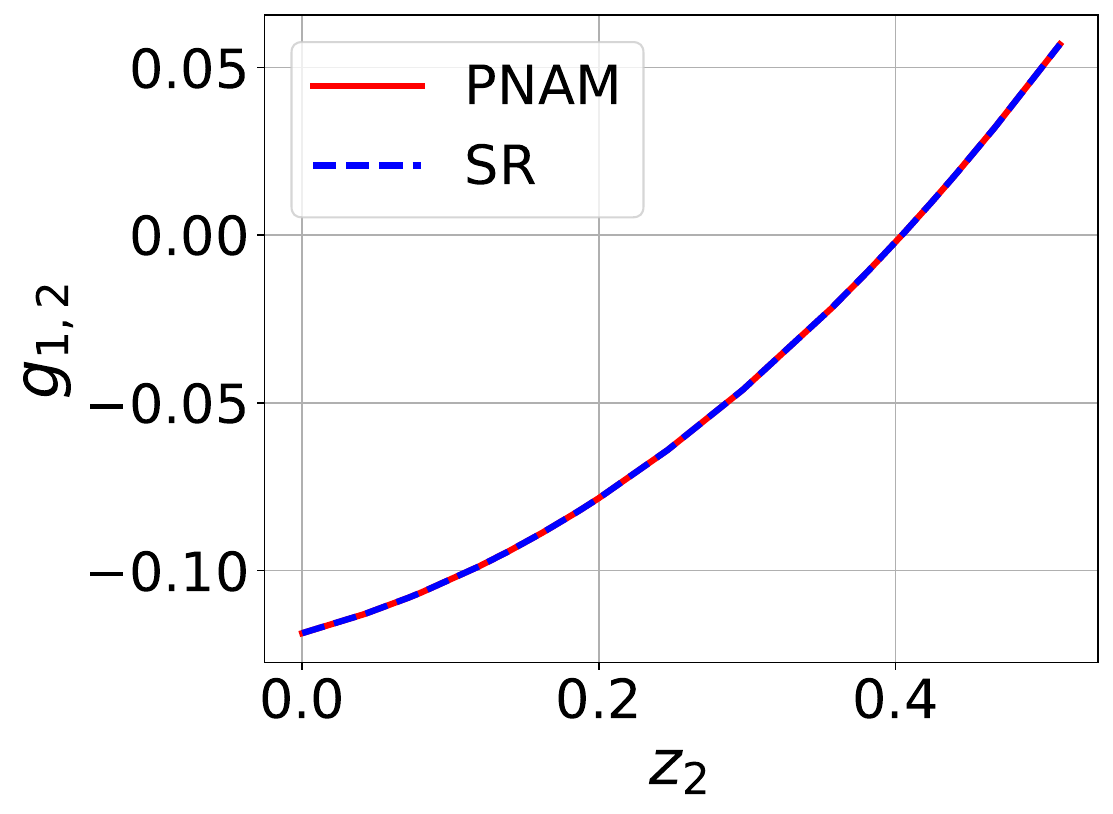}
\includegraphics[width=0.24\textwidth]{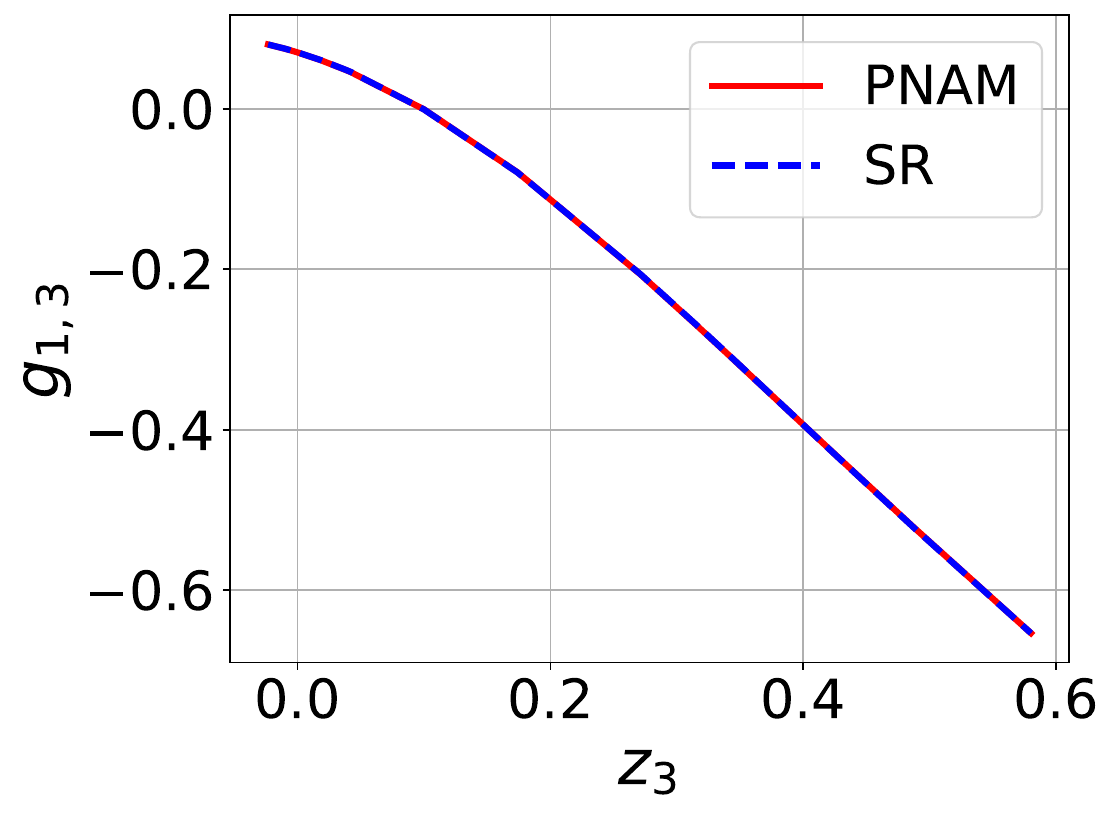}
\includegraphics[width=0.24\textwidth]{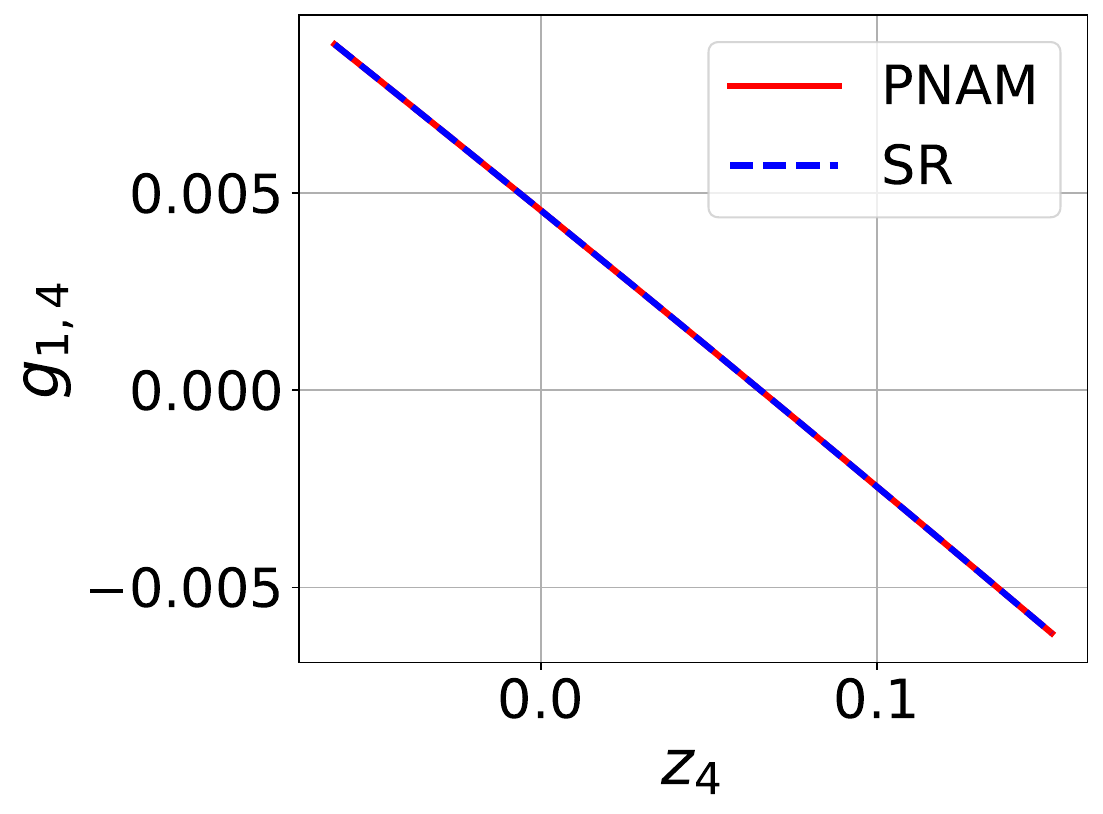}
\includegraphics[width=0.24\textwidth]{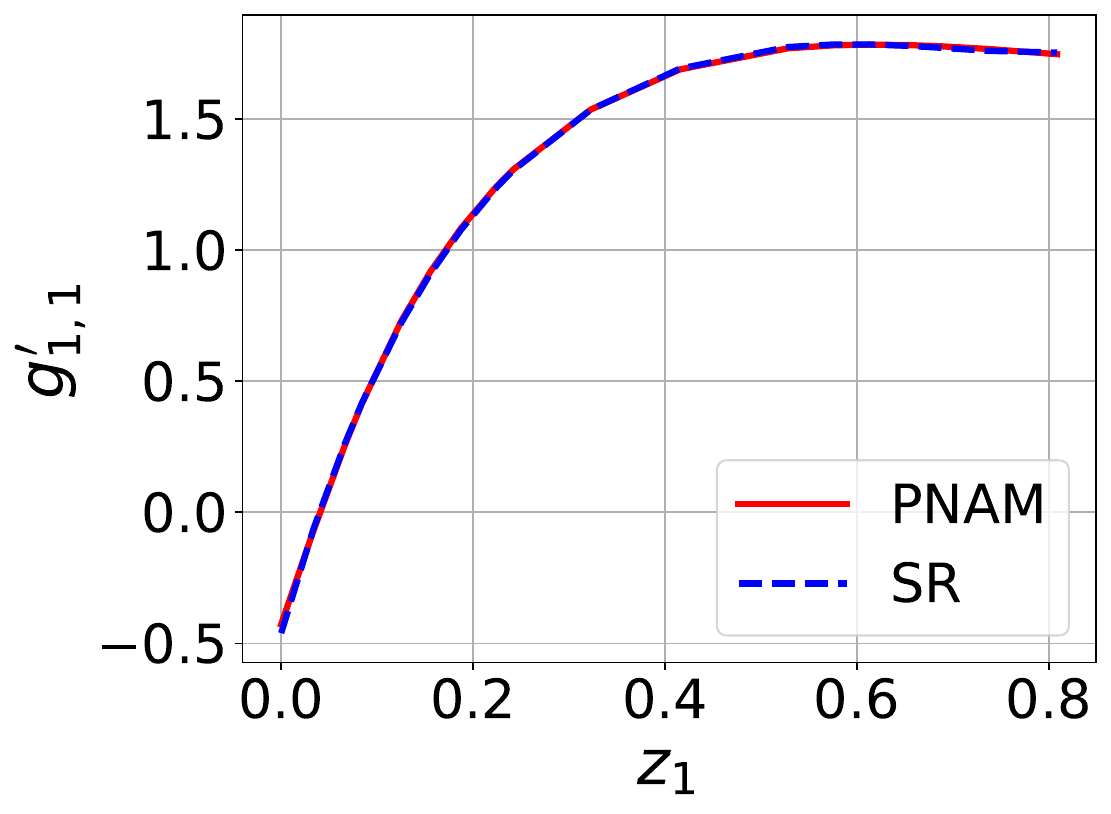}
\includegraphics[width=0.24\textwidth]{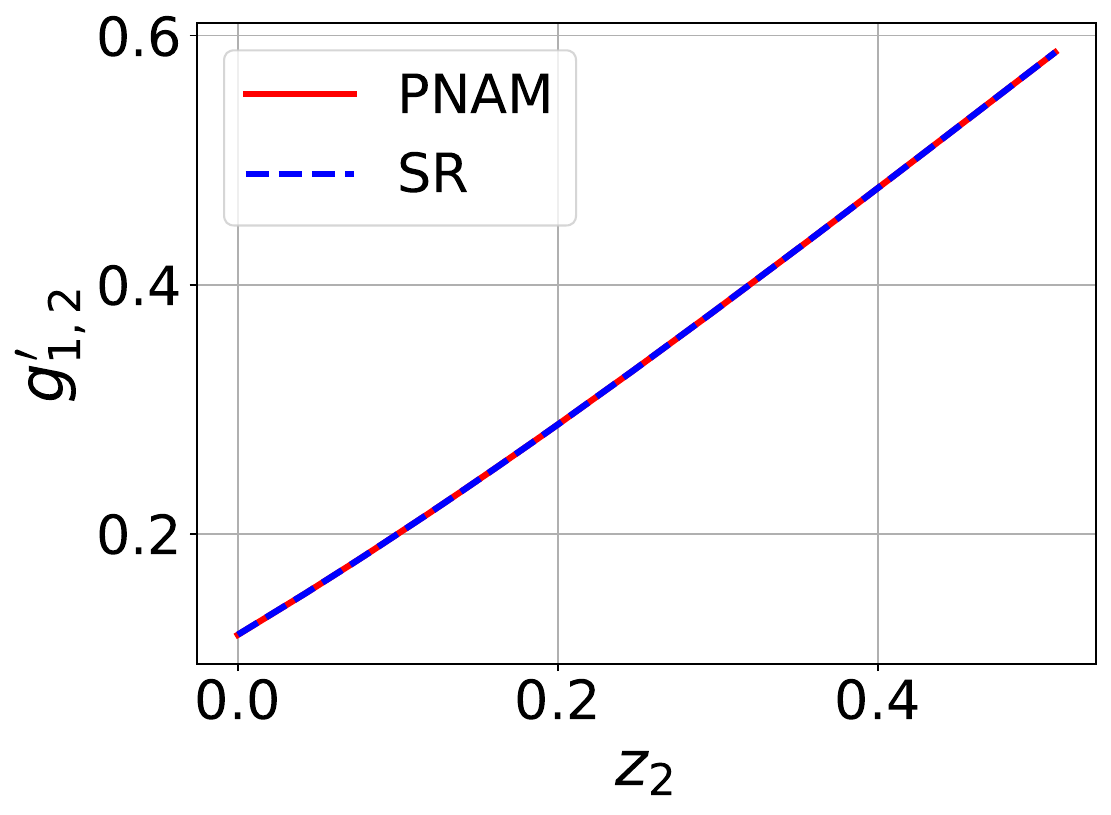}
\includegraphics[width=0.24\textwidth]{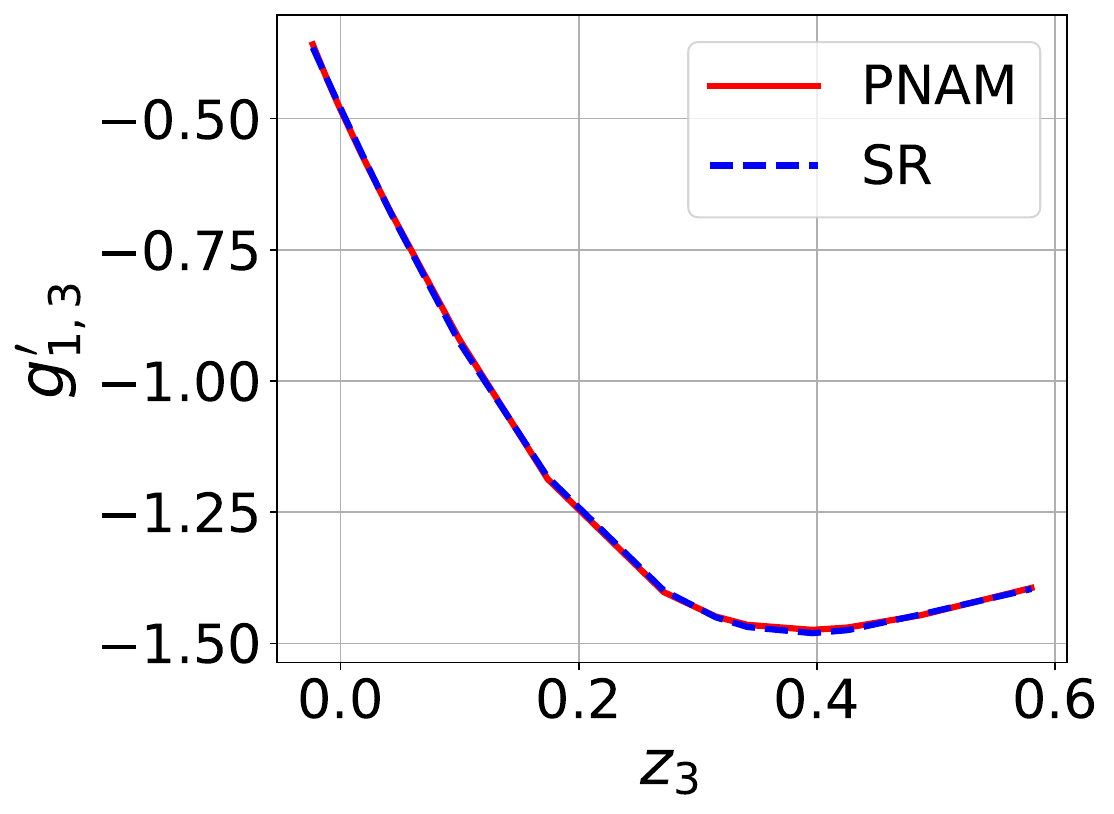}
\includegraphics[width=0.24\textwidth]{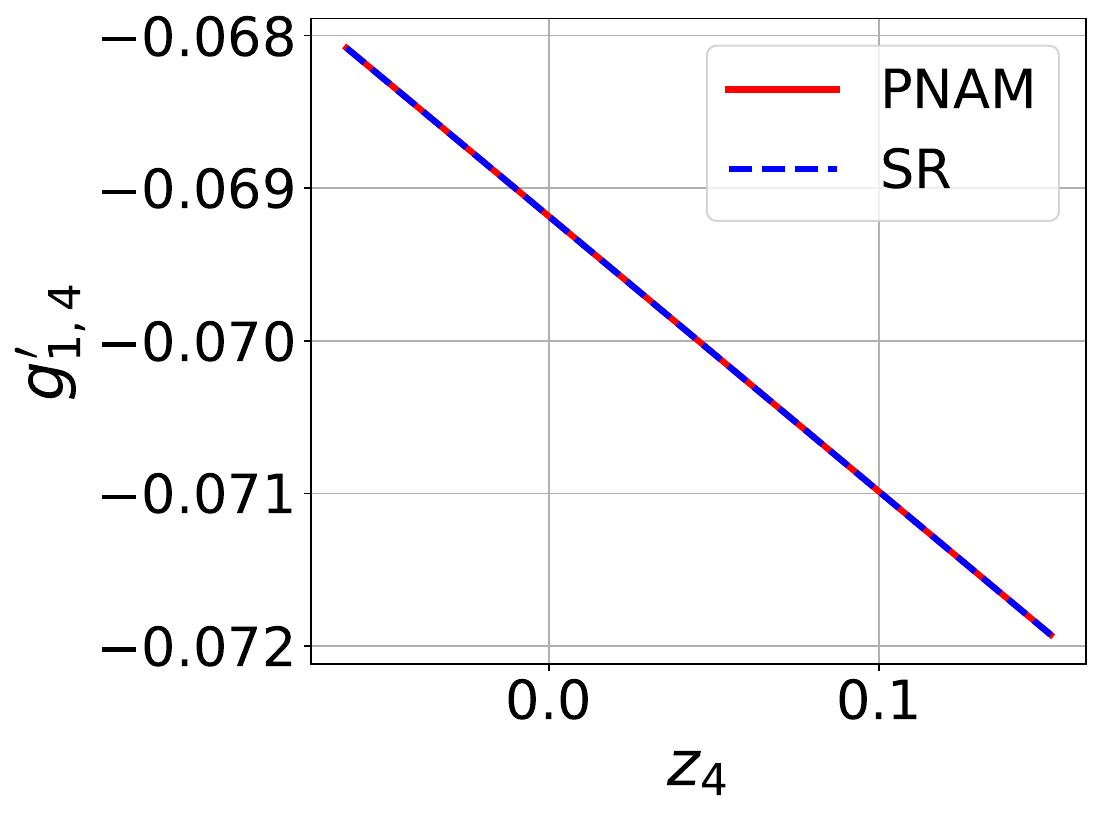}
\includegraphics[width=0.24\textwidth]{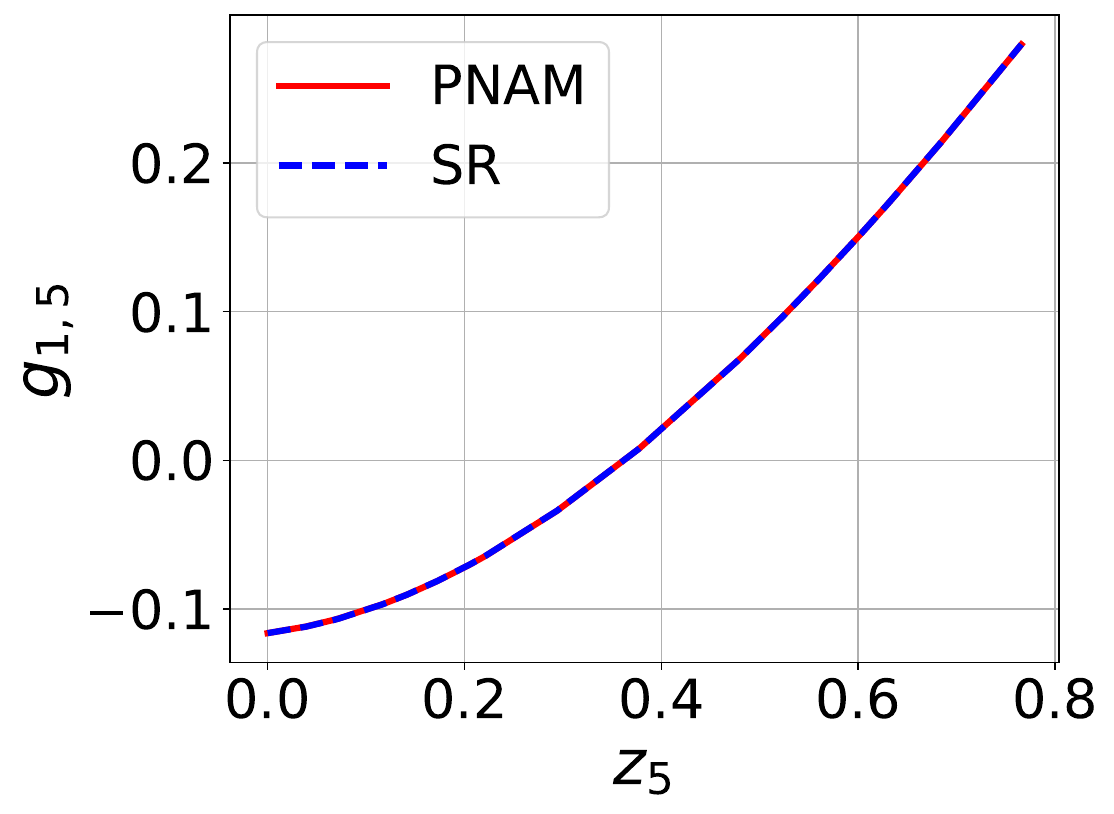}
\includegraphics[width=0.24\textwidth]{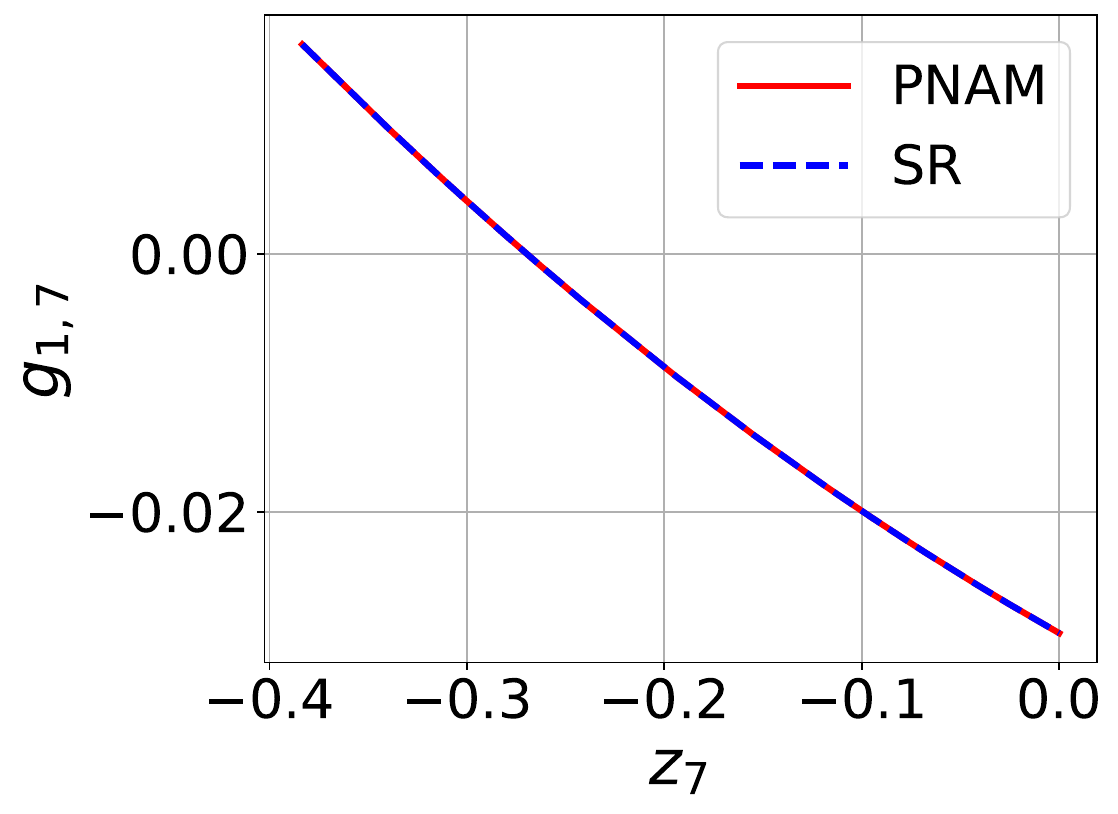}
\includegraphics[width=0.24\textwidth]{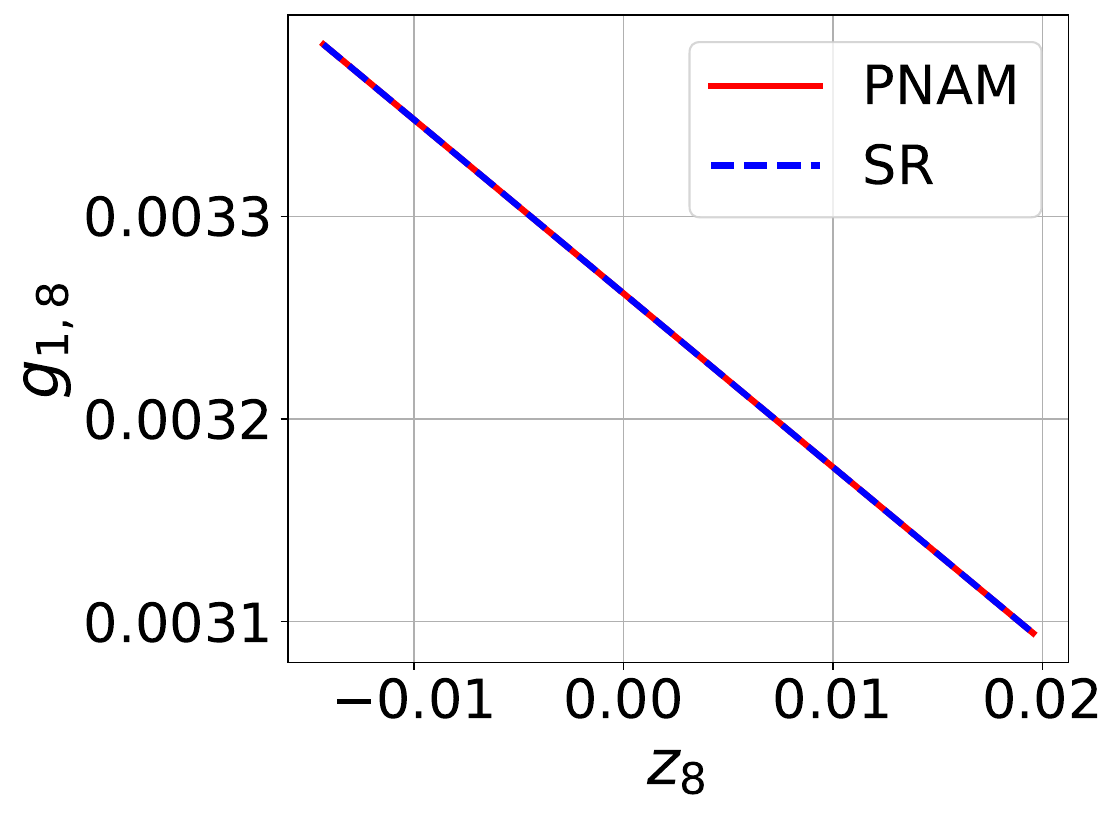} \\
\includegraphics[width=0.24\textwidth]{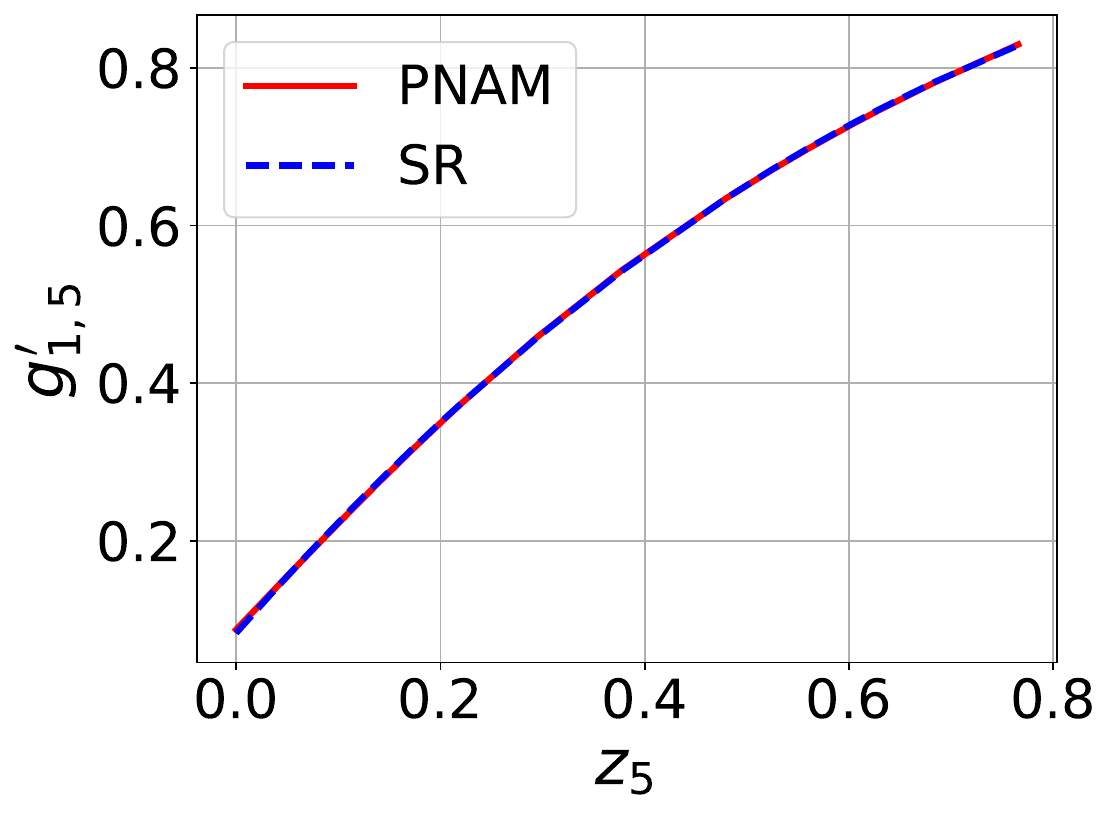}
\includegraphics[width=0.24\textwidth]{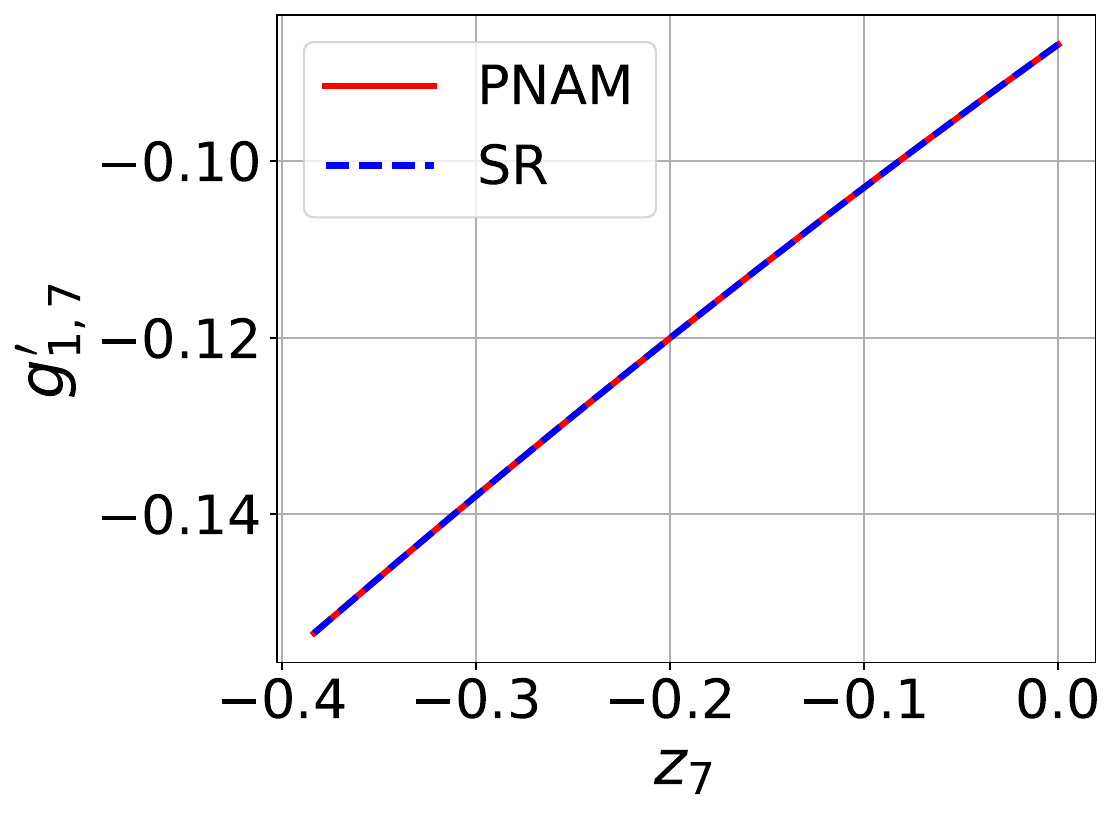}
\includegraphics[width=0.24\textwidth]{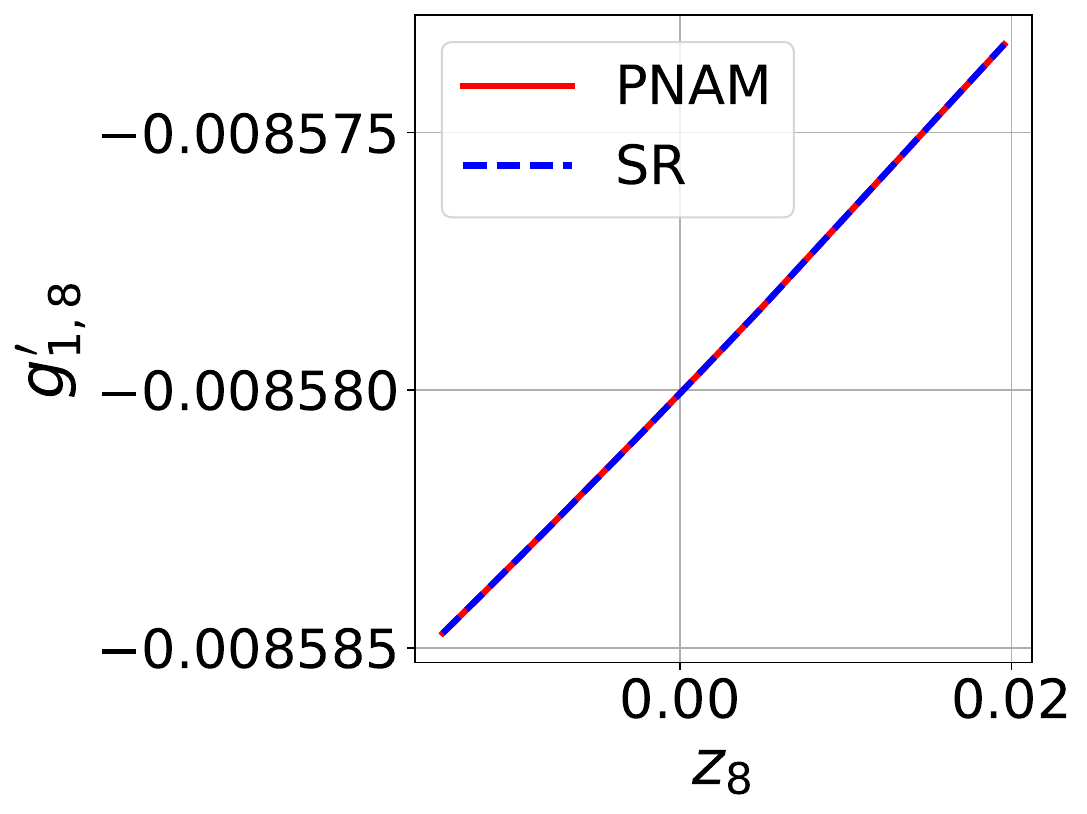}
\caption{Bases $\{g_{1j}\}$ and their derivatives. The expressions of $\{g_{1j}\}$ are presented in Table \ref{tab:phase_field_eqs}.}
\label{fig:phase_field_g1}
\end{figure}

\begin{figure}[!ht]
\centering
\includegraphics[width=0.24\textwidth]{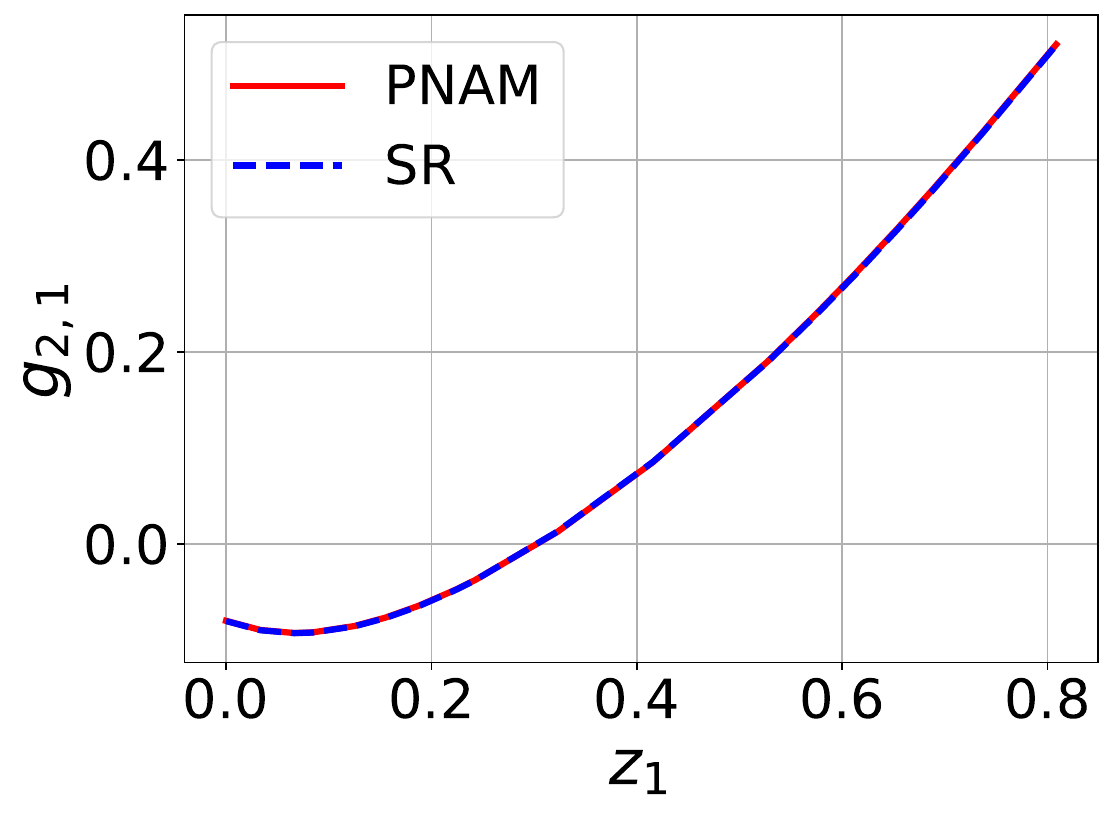}
\includegraphics[width=0.24\textwidth]{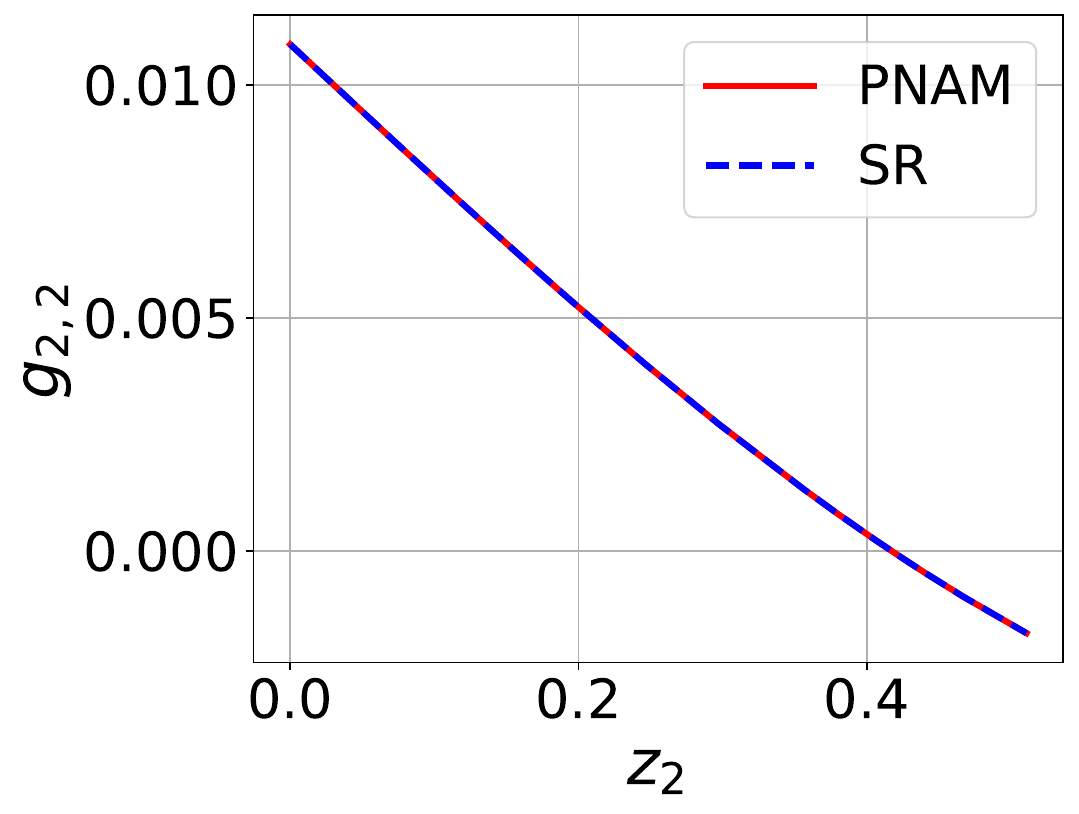}
\includegraphics[width=0.24\textwidth]{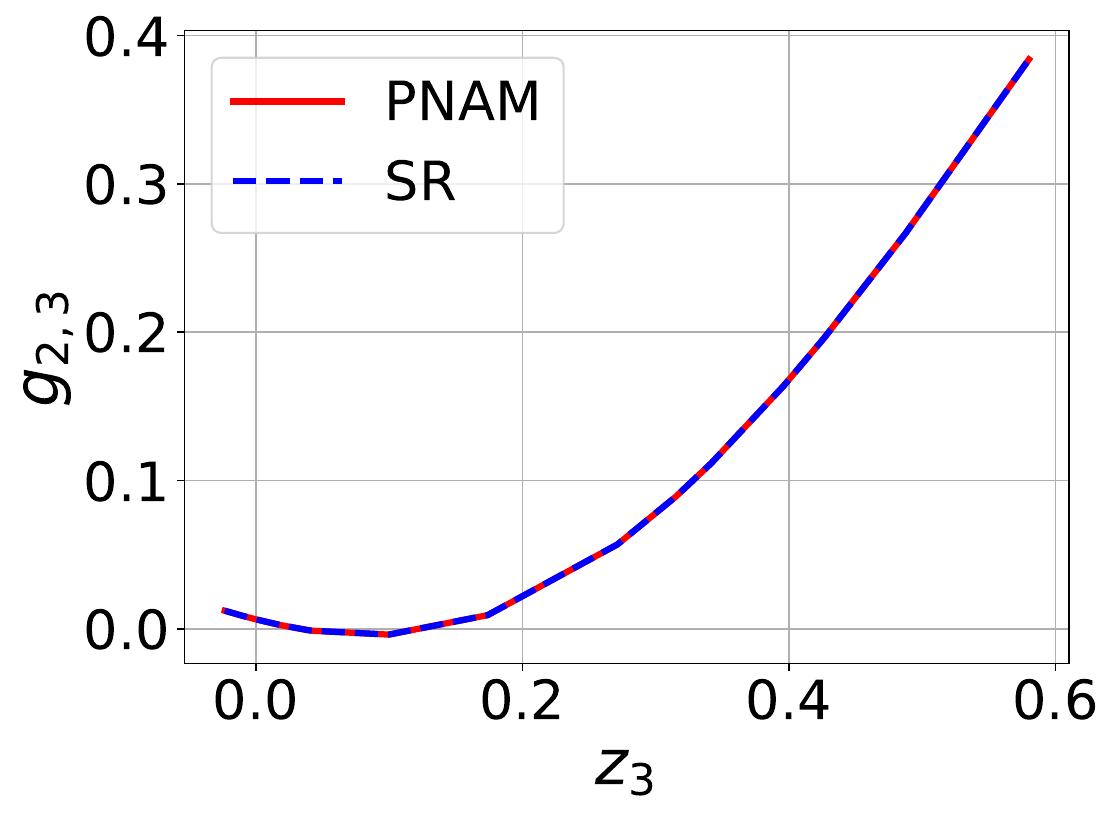} \\
\includegraphics[width=0.24\textwidth]{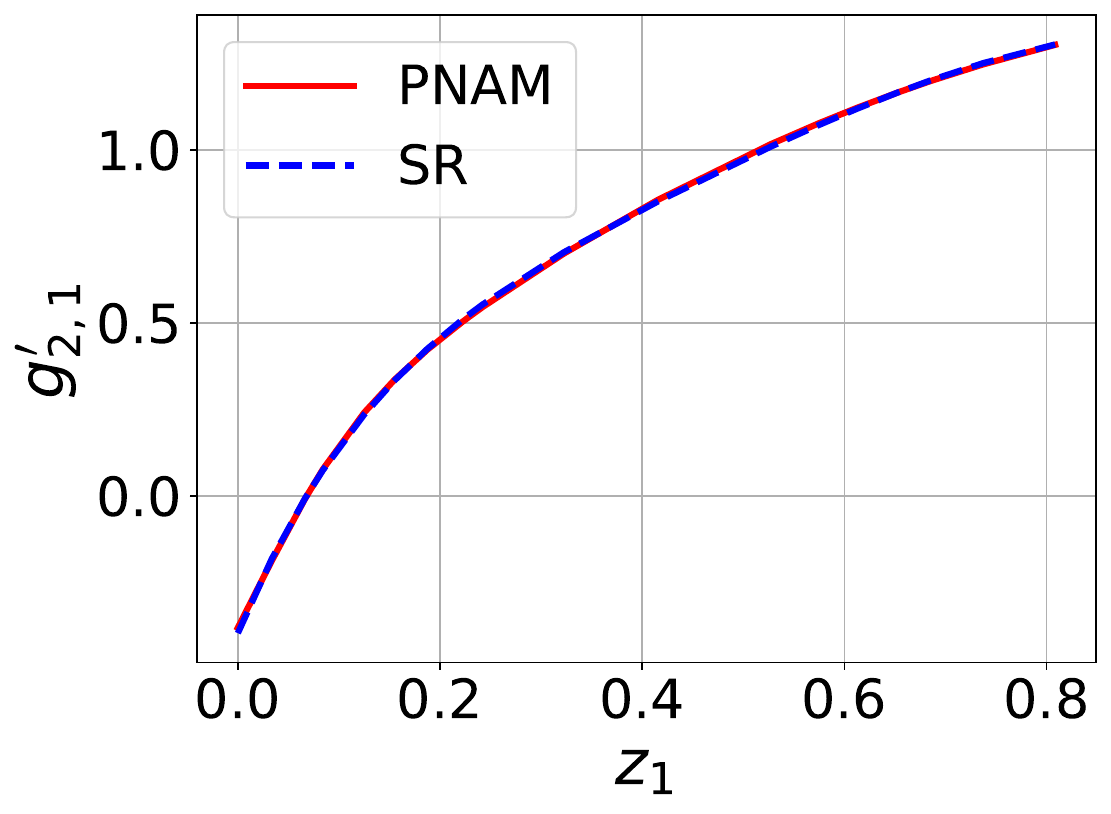}
\includegraphics[width=0.24\textwidth]{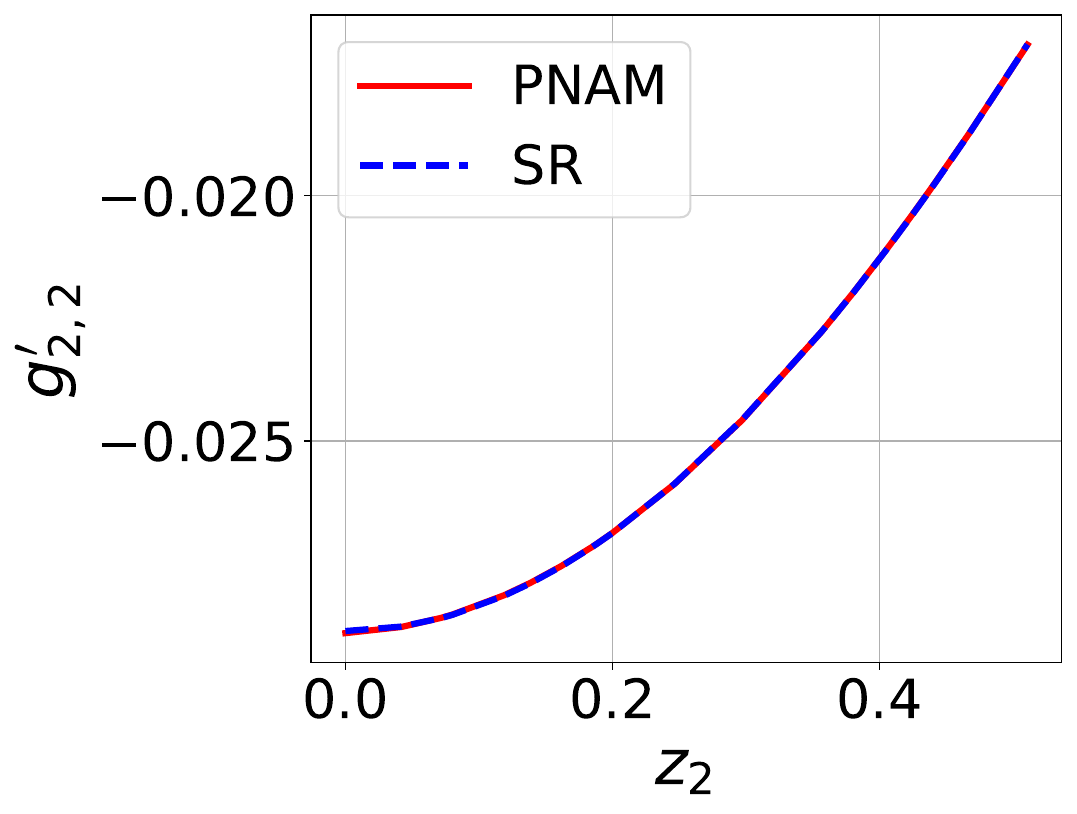}
\includegraphics[width=0.24\textwidth]{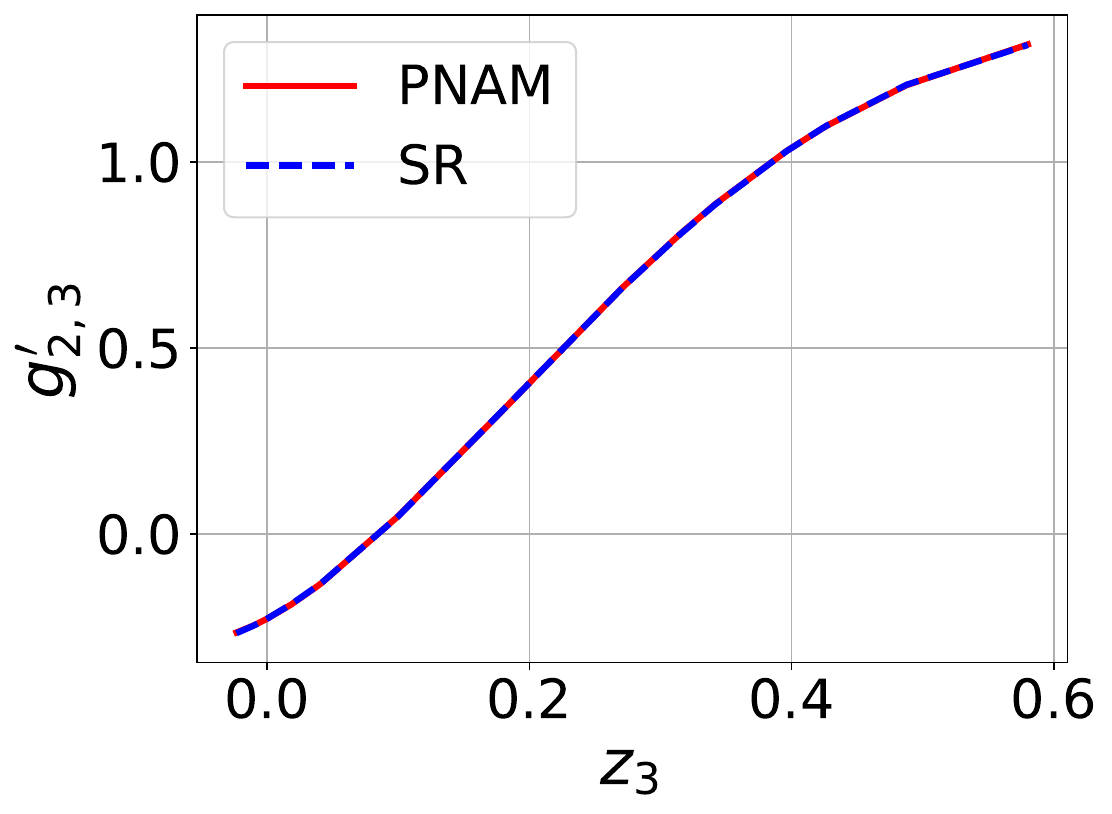} \\
\includegraphics[width=0.24\textwidth]{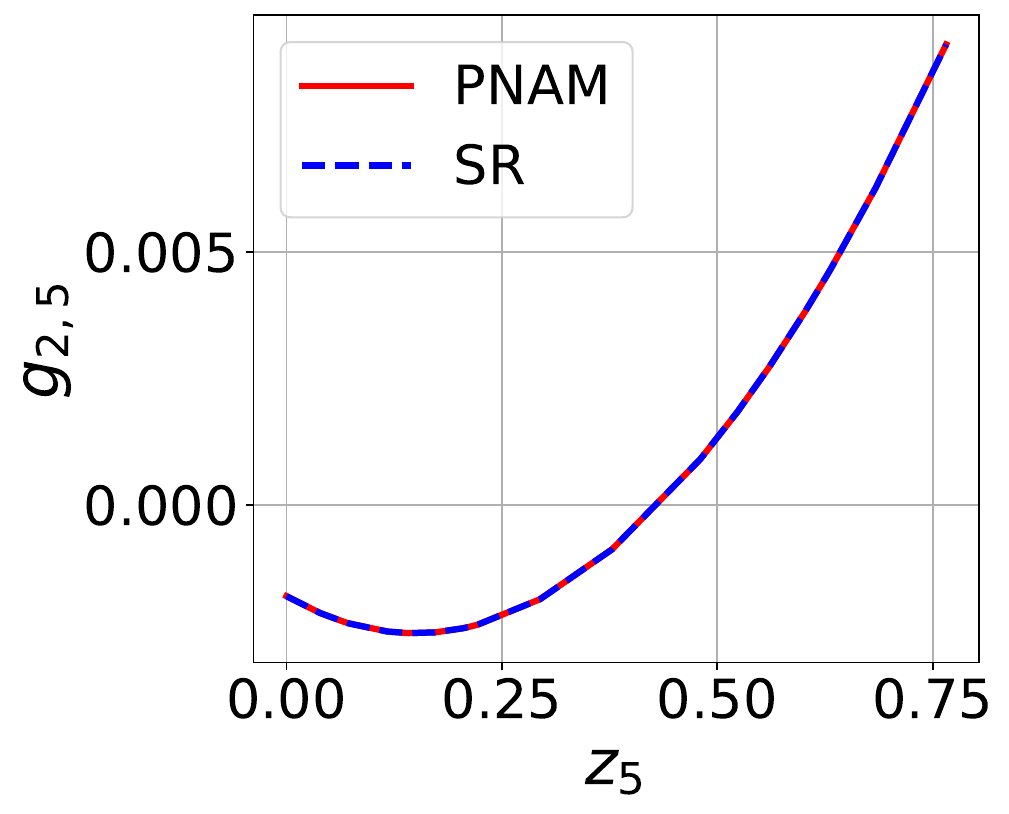}
\includegraphics[width=0.24\textwidth]{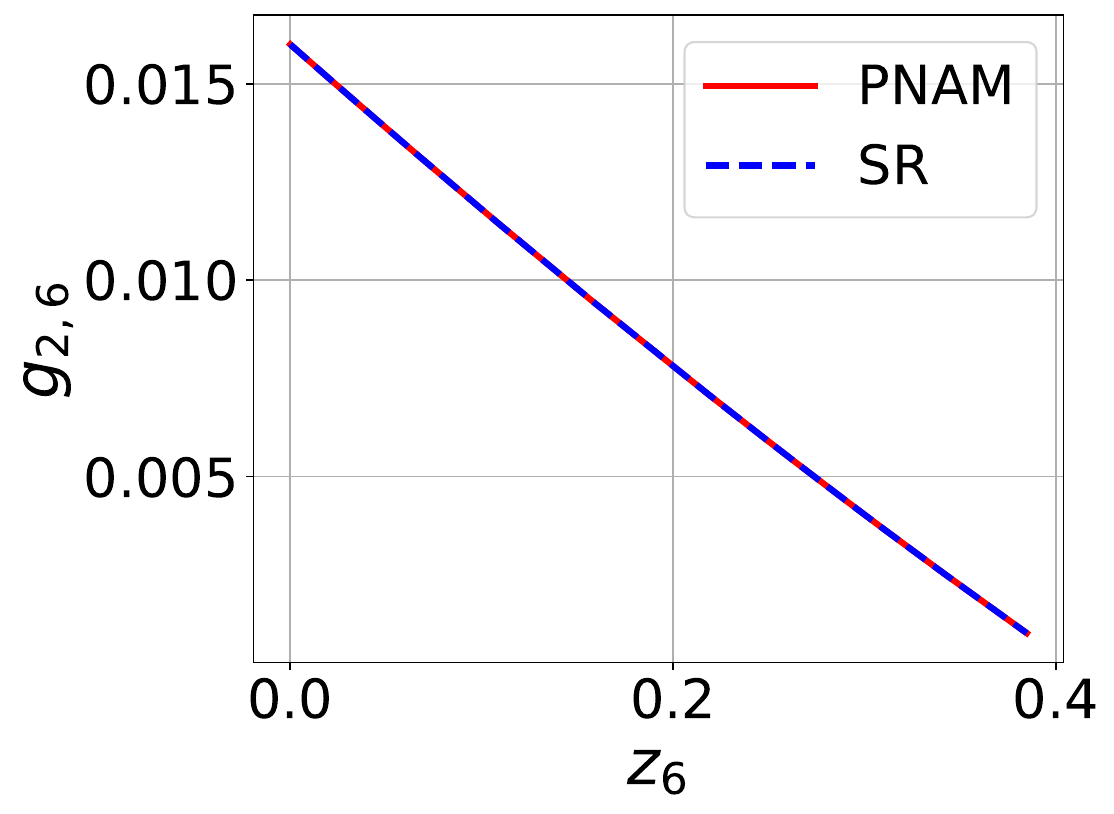}
\includegraphics[width=0.24\textwidth]{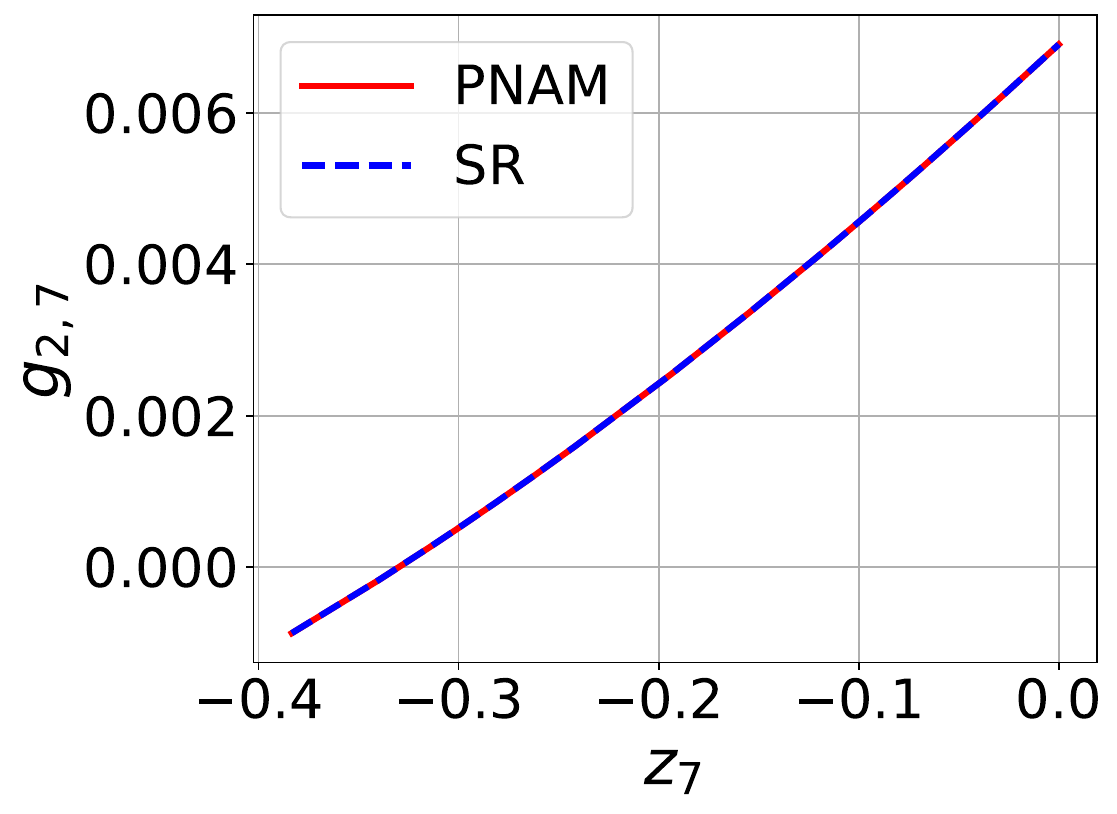} \\
\includegraphics[width=0.24\textwidth]{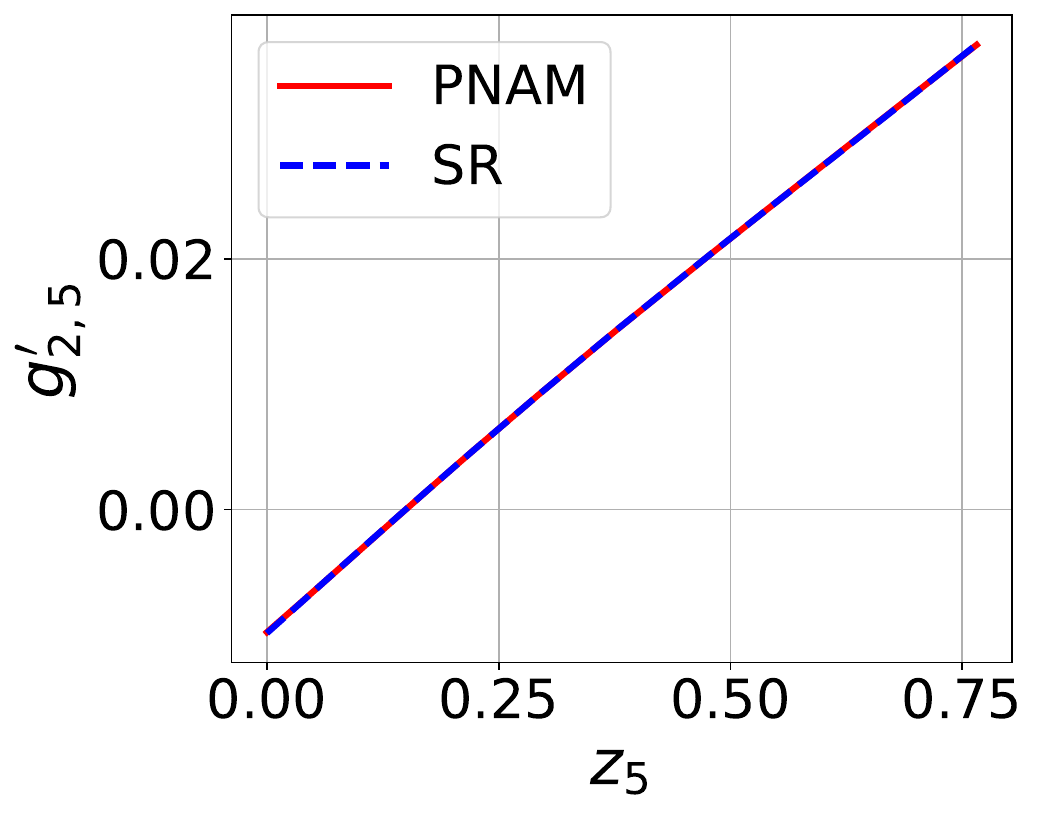}
\includegraphics[width=0.24\textwidth]{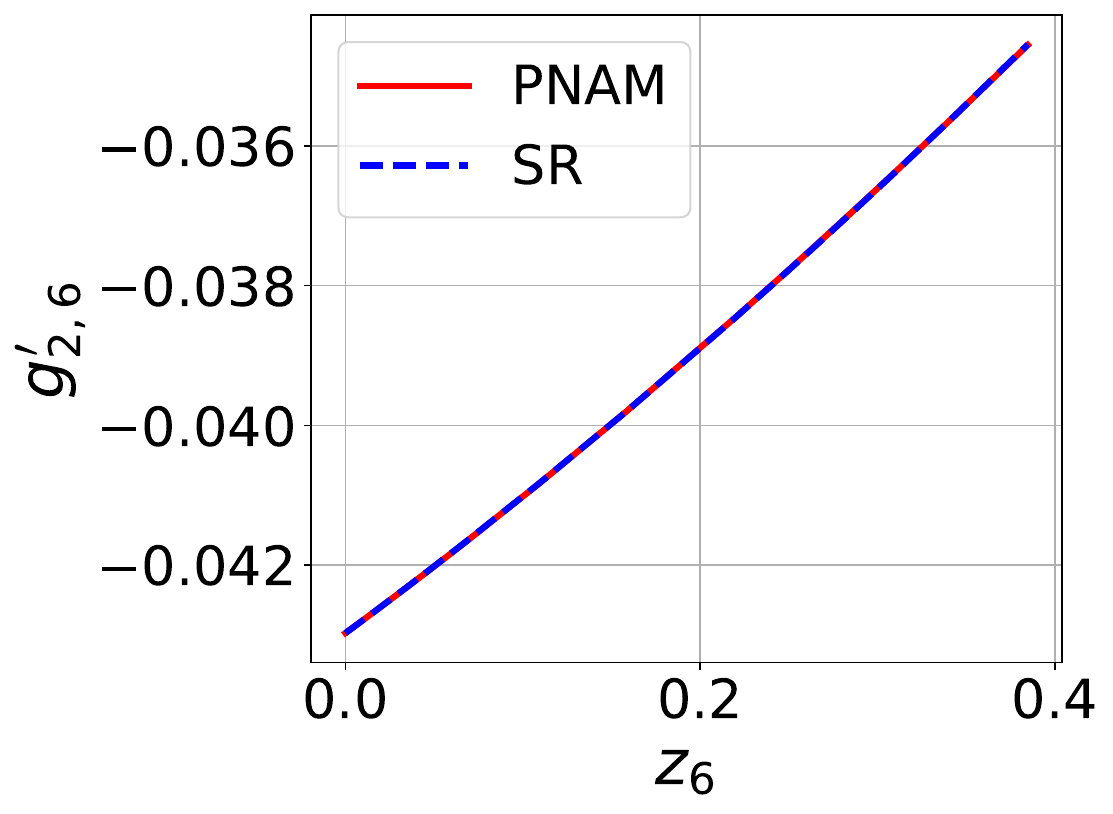}
\includegraphics[width=0.24\textwidth]{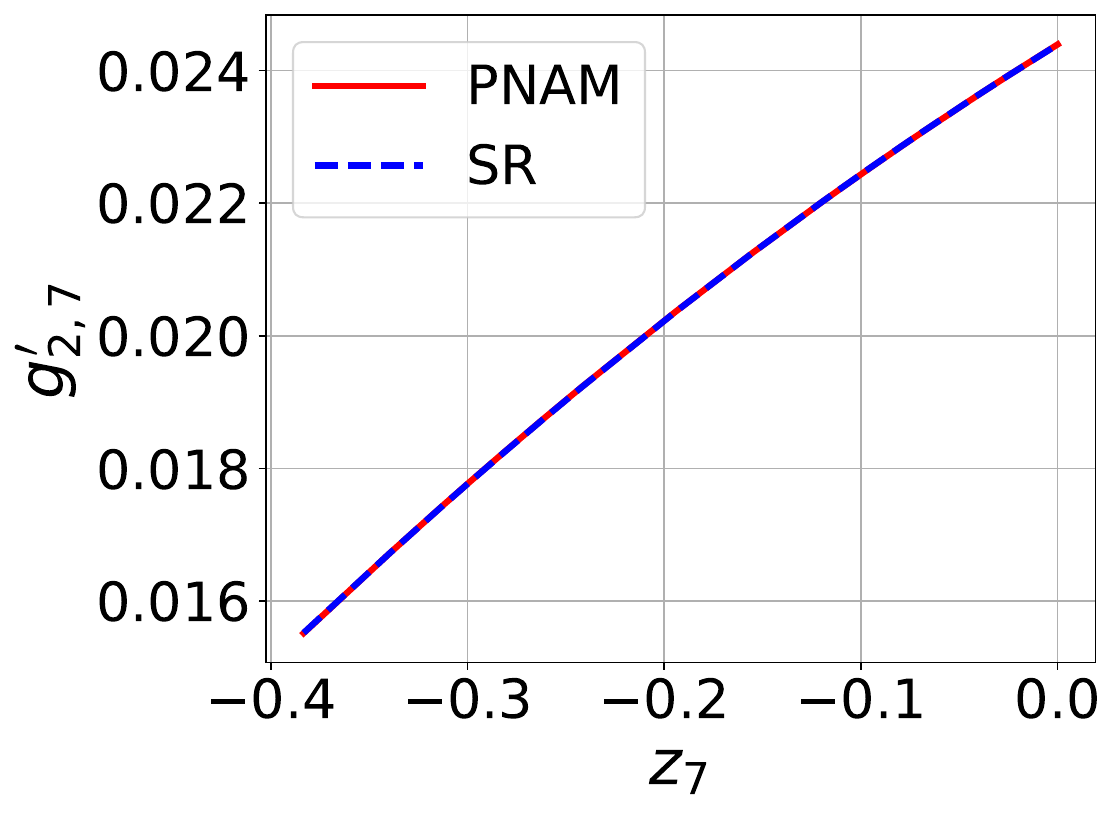}
\caption{Bases $\{g_{2j}\}$ and their derivatives. The expressions of $\{g_{2j}\}$ are presented in Table \ref{tab:phase_field_eqs}.}
\label{fig:phase_field_g2}
\end{figure}

\begin{figure}[!ht]
\centering
{(a)}{\includegraphics[width=0.3\textwidth]{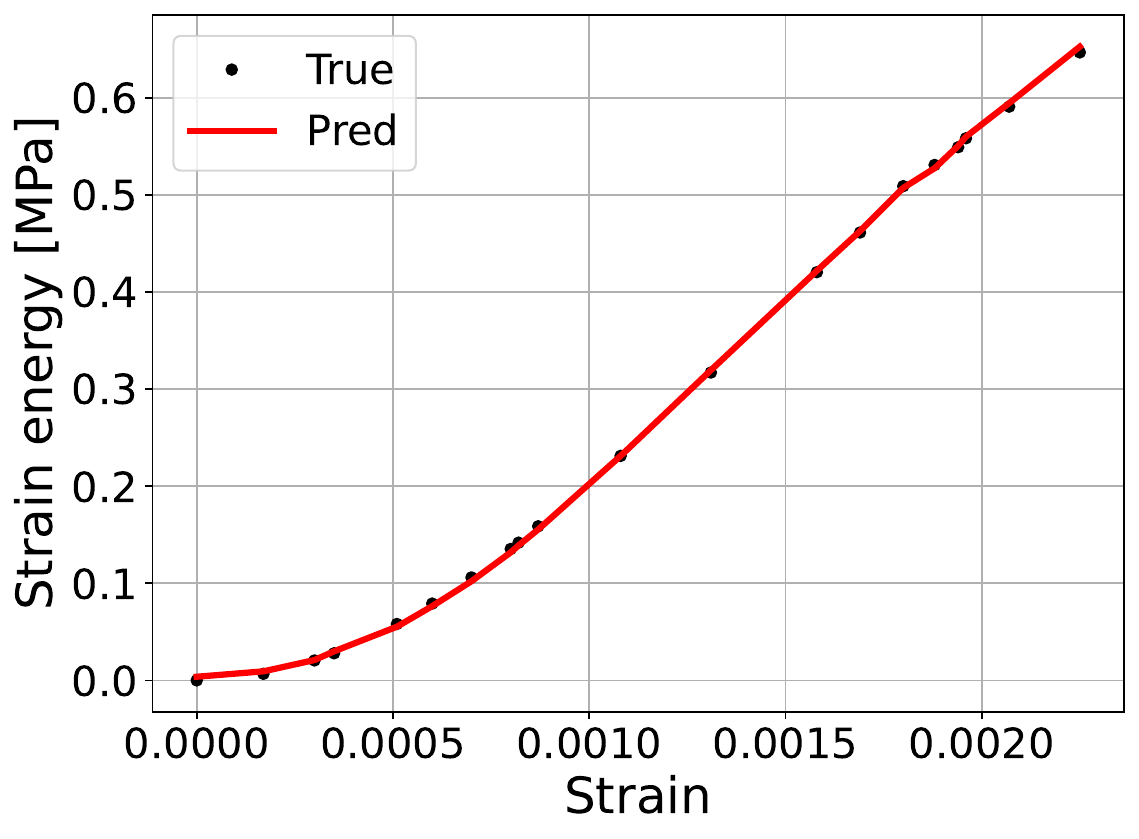}}
{(b)}{\includegraphics[width=0.3\textwidth]{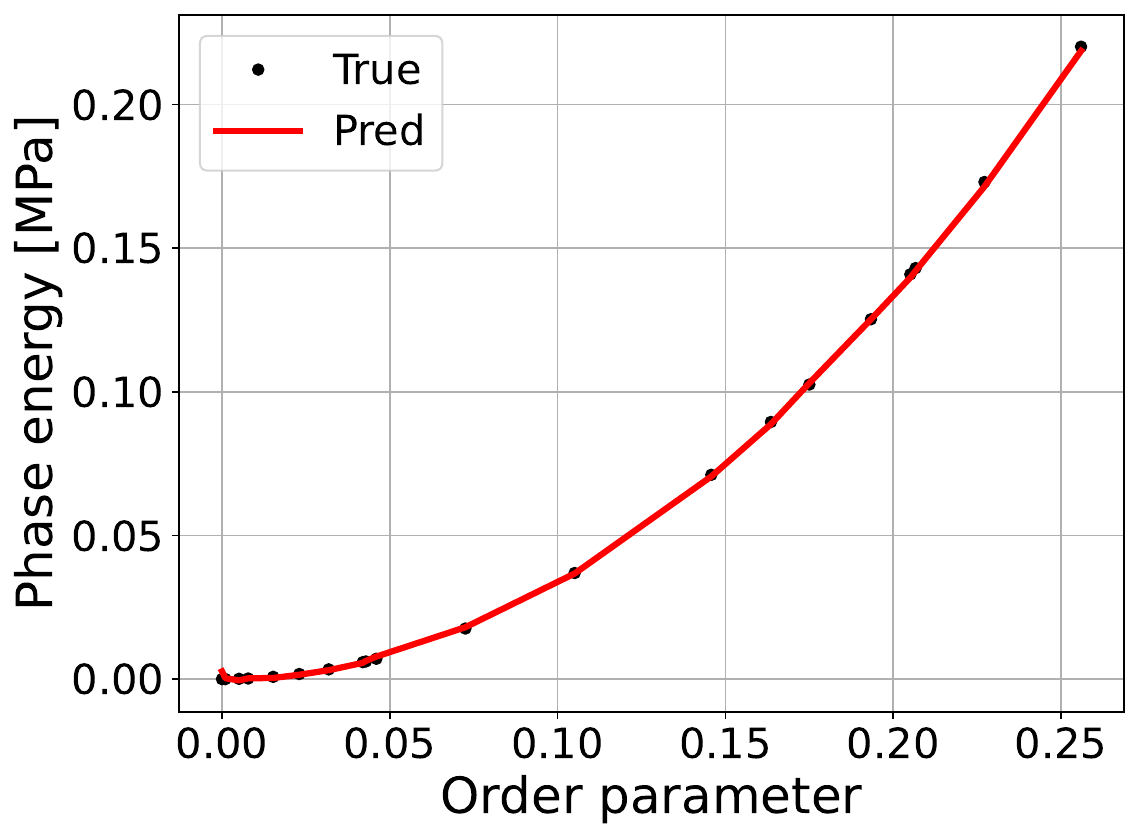}}
{(c)}{\includegraphics[width=0.3\textwidth]{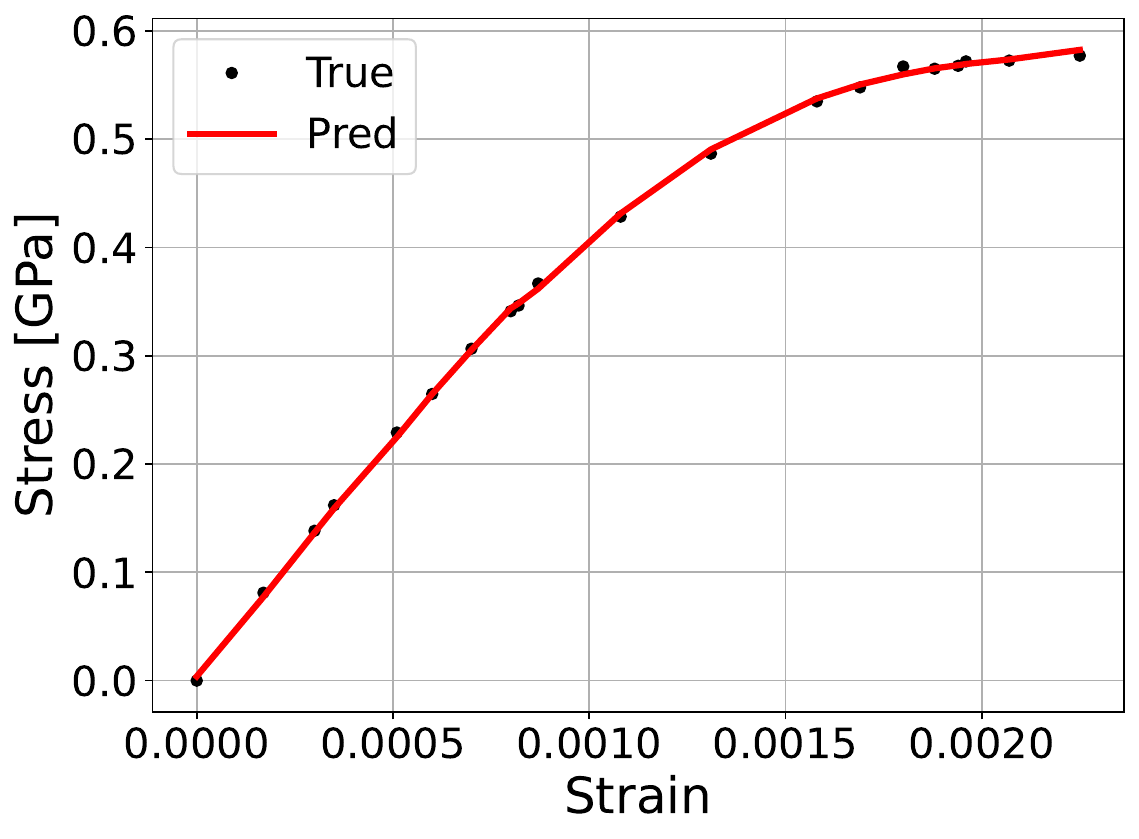}}
\caption{Test predictions for the linear combinations of the polynomials $\{g_{ij}\}$ in Table \ref{tab:phase_field_eqs}. (a) Linear combination of $\{g_{1j}\}$, (b) linear combination of $\{g_{2j}\}$, and (c) derivative of the sum of the linear combinations of $\{g_{1j}\}$ and $\{g_{2j}\}$ with respect to the strain.}
\label{fig:phase_field}
\end{figure}

All variables are scaled to have zero min and unit max.
Table \ref{tab:phase_field_compare} summarizes the performance of the MLP, the NAM, and four parameterizations of the PNAM, all utilizing three hidden layers of 256 neurons and trained for up to 5000 epochs with an early stopping patience of 50 epochs.
On average, the test MSE of PNAMs decreases as the projection dimension $M$ increases \citep{phan2025hydra}.
Nevertheless, a PNAM with $M = 8$ achieves the lowest validation MSE among all 60 networks underlying Table \ref{tab:phase_field_compare}, and we select this network for symbolic distillation.
We then set $w_6 = 1$ in Eq.~\eqref{eq:sr_loss} to ensure accurate stress predictions and convert this network with approximately $10^6$ parameters into a symbolic form with roughly 100 parameters.
As illustrated in Figs.~\ref{fig:phase_field_g1} and \ref{fig:phase_field_g2}, by employing SiLU as the activation function $A$ in Eq.~\eqref{eq:mlp}, all bases $\{g_{ij}\}$ can be approximated as single-variable polynomials, listed in Table \ref{tab:phase_field_eqs}.
Predictions of the strain energy, phase energy, and stress for the linear combinations of these polynomials are shown in Fig.~\ref{fig:phase_field}, achieving a test MSE of $2.25 \times 10^{-5}$.

\begin{table}[!ht]
\caption{Performance comparison of different neural network architectures for the two-output regression problem of predicting the strain energy and phase energy, from which the stress is derived. The mean and standard deviation of the combined test MSE for all three variables are computed over 10 runs.}
\label{tab:phase_field_compare}
\centering
\begin{tabular}{llll}
\toprule
Method & Architecture & Parameter count & Test MSE \\
\midrule
MLP & 4 layers: [3, 256, 256, 256, 2] & $2.6 \times 10^5$ & $(7.05 \pm 2.02) \times 10^{-5}$ \\
NAM & 4 layers: $3 \times [1, 256, 256, 256, 2]$ & $7.9 \times 10^5$ & $(1.12 \pm 0.62) \times 10^{-4}$ \\
PNAM & 4 layers: $3 \times [1, 256, 256, 256, 2]$ & $7.9 \times 10^5$ & $(3.31 \pm 2.11) \times 10^{-4}$ \\
PNAM & 4 layers: $8 \times [1, 256, 256, 256, 2]$ & $2.1 \times 10^6$ & $(1.34 \pm 0.91) \times 10^{-4}$ \\
PNAM & 4 layers: $16 \times [1, 256, 256, 256, 2]$ & $4.2 \times 10^6$ & $(8.01 \pm 3.76) \times 10^{-5}$ \\
PNAM & 4 layers: $32 \times [1, 256, 256, 256, 2]$ & $8.4 \times 10^6$ & $(6.47 \pm 2.71) \times 10^{-5}$ \\
\bottomrule
\end{tabular}
\end{table}

\begin{table}[!ht]
\caption{Mathematical expressions for the phase field dataset. Inputs $\chi_1$, $\chi_2$, and $\chi_3$ are the strain, order parameter, and magnitude of the material gradient of the order parameter, respectively. To convert the PNAM with $M = 8$ in Table \ref{tab:phase_field_compare} and $T_0 = \zeta_0 = 0.05$ (see Section \ref{subsec:sr}) into symbolic form, bases $\{g_{ij}\}$ in Eq.~\eqref{eq:sr_loss} use addition, subtraction, multiplication, and square as operators. Counting every constant in $\{g_{ij}\}$ and $\{z_j\}$ as one parameter, the symbolic form has 112 parameters and achieves a test MSE of $2.25 \times 10^{-5}$.}
\label{tab:phase_field_eqs}
\centering
\resizebox{\textwidth}{!}{
\begin{tabular}{llll}
\toprule
\multicolumn{2}{c}{Bases $\{g_{ij}\}$} & \multicolumn{2}{c}{Projected variables $\{z_j\}$} \\
\cmidrule(lr){1-2} \cmidrule(lr){3-4}
Basis & Expression & Variable & Expression \\
\midrule
\multirow{2}{*}{$g_{1, 1}$} & $-z_1^3 + 1.524 z_1^2 + z_1 + ((-z_1 + (z_1 - 0.028)^2 + 0.75)^2 \ -$ & \multirow{4}{*}{$z_1$} & \multirow{4}{*}{$203.741 \chi_1 + 1.87 \chi_2 - 3.048 \chi_3$} \\
& $0.102)^2 - 0.269$ & & \\
\multirow{2}{*}{$g_{2, 1}$} & $z_1^2 - (z_1 - 0.597) (z_1 - 0.291) (1.378 z_1 (0.786 z_1 - 1)^4 \ +$ & & \\
& $0.525 z_1 - 0.074) - 0.093$ & & \\
\hline
$g_{1, 2}$ & $z_2^3 (0.147 - 0.074 z_2) + 0.383 z_2^2 + 0.119 z_2 - 0.119$ & \multirow{2}{*}{$z_2$} & \multirow{2}{*}{$235.816 \chi_1 - 0.461 \chi_2 + 2.333 \chi_3$} \\
$g_{2, 2}$ & $-10.229 z_2 (0.003 - 0.002 z_2^2) + 0.011$ & & \\
\hline
$g_{1, 3}$ & $0.967 z_3 (z_3 - 1.108) (-3.255 z_3^3 + z_3 - 1.418) - 2 z_3 + 0.07$ & \multirow{2}{*}{$z_3$} & \multirow{2}{*}{$-114.828 \chi_1 + 1.284 \chi_2 + 12.089 \chi_3$} \\
$g_{2, 3}$ & $1.046 z_3 + 2.265 (z_3 - 0.403)^2 (0.65 z_3^3 - z_3^2 + 0.835)^2 - 0.25$ & & \\
\hline
$g_{1, 4}$ & $-z_4 (0.009 z_4 + 0.069) + 0.005$ & \multirow{2}{*}{$z_4$} & \multirow{2}{*}{$-185.754 \chi_1 + 1.866 \chi_2 + 2.19 \chi_3$} \\
$g_{2, 4}$ & $0$ & & \\
\hline
$g_{1, 5}$ & $0.118 z_5 (2 z_5 + 0.217) (0.033 z_5^2 (z_5 + 1) - z_5 + 3.239) - 0.116$ & \multirow{3}{*}{$z_5$} & \multirow{3}{*}{$190.698 \chi_1 + 0.282 \chi_2 + 6.271 \chi_3$} \\
\multirow{2}{*}{$g_{2, 5}$} & $-z_5 (-0.03 z_5 + 0.003 (0.772 z_5 (z_5 - 0.466)^2 - z_5 + 0.596)^2 \ +$ & & \\
& $0.009) - 0.002$ & & \\
\hline
$g_{1, 6}$ & $0$ & \multirow{2}{*}{$z_6$} & \multirow{2}{*}{$1.503 \chi_2$} \\
$g_{2, 6}$ & $z_6 (0.003 z_6^2 + 0.009 z_6 - 0.043) + 0.016$ & & \\
\hline
$g_{1, 7}$ & $z_7 (z_7^2 (z_7^{10} (z_7 + 0.917)^2 - 0.014) + 0.079 z_7 - 0.087) - 0.029$ & \multirow{2}{*}{$z_7$} & \multirow{2}{*}{$-119.681 \chi_1 + 0.531 \chi_2 - 5.929 \chi_3$} \\
$g_{2, 7}$ & $-0.004 z_7^3 + 0.009 z_7^2 + 0.024 z_7 + 0.007$ & & \\
\hline
$g_{1, 8}$ & $0.003 - 0.009 z_8$ & \multirow{2}{*}{$z_8$} & \multirow{2}{*}{$-27.984 \chi_1 - 0.769 \chi_2 + 6.628 \chi_3$} \\
$g_{2, 8}$ & $0$ & & \\
\bottomrule
\end{tabular}}
\end{table}

\begin{figure}[!ht]
\centering
\includegraphics[width=0.5\textwidth]{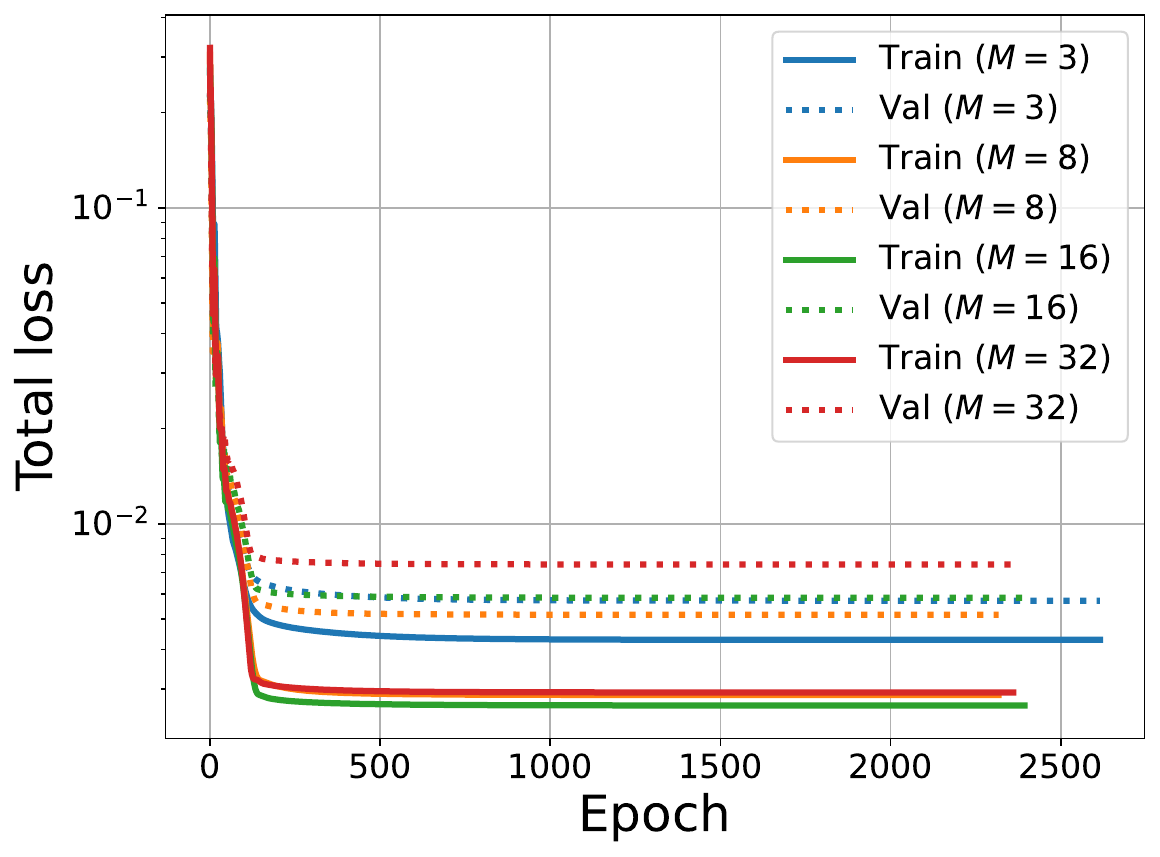}
\caption{Effects of the projection dimension $M$ on the convergence of PNAMs for the phase field dataset. Total loss corresponds to the loss function in Eq.~\eqref{eq:nn_loss}. Each basis MLP has three hidden layers of 256 neurons; all models are trained using regularization weights $w_1 = \dots = w_5 = 0.001$.}
\label{fig:phase_field_loss}
\end{figure}

Table \ref{tab:phase_field_time} reports training and inference times for the neural networks in Table \ref{tab:phase_field_compare}, alongside the inference time of the symbolic form in Table \ref{tab:phase_field_eqs}.
As expected, training and inference times increase with $M$.
However, converting the PNAM into a compact symbolic expression yields faster inference, consistent with the knot theory results in Table \ref{tab:knot_theory_regress_time}.
Finally, Fig.~\ref{fig:phase_field_loss} reveals that increasing $M$ improves performance but may lead to overfitting after saturation.

\begin{table}[!ht]
\caption{Training and inference times of the neural networks in Table \ref{tab:phase_field_compare}, along with the inference time of the combined expression (i.e., the sum of the linear combinations of $\{g_{1j}\}$ and $\{g_{2j}\}$) in Table \ref{tab:phase_field_eqs}. Each nonzero basis $g_{ij}$ is evolved for one minute. Training times are obtained from an NVIDIA A100-SXM4-40GB GPU, while inference times are obtained from an Apple M1 CPU with 8 GB of memory. The test MSE is repeated to aid comparison.}
\label{tab:phase_field_time}
\centering
\resizebox{\textwidth}{!}{
\begin{tabular}{llll}
\toprule
Method & Training time (s/epoch) & Inference time ($\upmu$s/sample) & Test MSE \\
\midrule
MLP & 1.1 & 27 & $(7.05 \pm 2.02) \times 10^{-5}$ \\
NAM & 3.0 & 61 & $(1.12 \pm 0.62) \times 10^{-4}$ \\
PNAM ($M = 3$) & 3.2 & 58 & $(3.31 \pm 2.11) \times 10^{-4}$ \\
PNAM ($M = 8$) & 6.8 & 150 & $(1.34 \pm 0.91) \times 10^{-4}$ \\
PNAM ($M = 16$) & 13 & 320 & $(8.01 \pm 3.76) \times 10^{-5}$ \\
PNAM ($M = 32$) & 26 & 610 & $(6.47 \pm 2.71) \times 10^{-5}$ \\
Expression (Table \ref{tab:phase_field_eqs}) & -- & 17 & $2.25 \times 10^{-5}$ \\
\bottomrule
\end{tabular}}
\end{table}

\end{document}